\pdfoutput=1
\documentclass[10pt,twocolumn]{article}

\usepackage[letterpaper,textwidth=6.75in,textheight=9.25in,columnsep=0.25in,centering]{geometry}

\usepackage[round]{natbib}

\usepackage[hyphens]{url}
\usepackage{graphicx}
\usepackage{caption}
\usepackage{longtable}
\usepackage{float}
\usepackage{amsmath}
\usepackage{amssymb}
\usepackage[hidelinks,hypertexnames=false,hyperfootnotes=false]{hyperref}
\hypersetup{pdftitle={The Signed Geometry of One-Shot Recourse: On-Path Validity and the Signed-Curvature Criterion},pdfauthor={Hazar Yueksel}}
\newcommand{\suppnum}[1]{\refstepcounter{equation}\theequation\label{#1}}

\makeatletter
\newcommand{\StartSupplement}{%
  \clearpage
  \onecolumn
  \setcounter{secnumdepth}{0}%
  \setcounter{section}{0}\setcounter{subsection}{0}\setcounter{subsubsection}{0}%
  \setcounter{paragraph}{0}\setcounter{subparagraph}{0}%
  \setcounter{equation}{0}%
  \setcounter{table}{0}\setcounter{figure}{0}\setcounter{footnote}{0}%
  \def\@currentlabel{}%
  \renewcommand{\theequation}{S1.\arabic{equation}}%
  \renewcommand{\thetable}{S\arabic{table}}%
  \renewcommand{\thefigure}{S\arabic{figure}}%
  \raggedbottom
  \setlength{\emergencystretch}{3em}%
  \renewcommand{\topfraction}{0.9}%
  \renewcommand{\bottomfraction}{0.85}%
  \renewcommand{\textfraction}{0.08}%
  \renewcommand{\floatpagefraction}{0.75}%
  \setcounter{topnumber}{4}%
  \setcounter{totalnumber}{6}}
\makeatother

\begin{document}

\twocolumn[{%
\centering
{\LARGE\bfseries The Signed Geometry of One-Shot Recourse:\\[2pt] On-Path Validity and the Signed-Curvature Criterion\par}
\vspace{1.2em}
{\large Hazar Yueksel\par}
\vspace{0.3em}
Google\\
\texttt{hazar@hazaryueksel.com}\par
\vspace{1.8em}
}]
{\renewcommand{\thefootnote}{}\footnotetext{The author is now an independent researcher.}%
\footnotetext{The code archive referred to in the text is not included in this version.}}

\begin{abstract}
Closed-form recourse moves a rejected user along the unit gradient \(\hat g\) of the classifier score \(f\) by the promised distance \(d_p=|f(x)|/\lVert\nabla f(x)\rVert\), at which the linearized score reaches zero. We ask when this one-shot step succeeds and what additional model queries change. To leading order the step ends on the favorable side exactly when the path curvature \(\kappa=\hat g^\top\nabla^2 f(x)\,\hat g\) is nonnegative. Across 80 shallow models, the fraction of rejected users whose step ends there and the fraction with \(\kappa\ge0\) correlate at \(r=0.985\), although on Fashion-MNIST the first falls below the second by 8.2 points on average. No rule that uses only the score value and gradient can be valid for every score with path curvature bounded by \(K\) without overshooting some by order \(Kd_p^2/\lVert\nabla f(x)\rVert\). When the curvature is also Lipschitz and the step is short, one evaluation of \(f\) at the promised point attains the minimax rate among deterministic one-query rules that know the curvature bound and its Lipschitz constant, and split-conformal calibration makes such a rule reach the first crossing or abstain with probability at least \(1-\delta\). Training with an asymmetric curvature penalty lets 99--100\% of paths cross within the promised step on undershoot-prone shallow data, at about \mbox{4--22} times the overshoot of symmetric penalties (Fashion-MNIST, COMPAS). Because \(\kappa\) and \(d_p\) depend on how the score is scaled, part of this gain can be a longer promised step, and at matched validity a smaller audit of briefly trained models finds no uniform advantage over tuned inflation. Where a per-user line search along the ray is affordable, it is exact to grid resolution and preferable.
\end{abstract}

\section{Introduction}
\label{sec:intro}

Algorithmic recourse tells a person who received an adverse decision how to change their features to reverse it \citep{wachter2017counterfactual,karimi2022survey}. The simplest recommendation moves a rejected point \(x\) along the unit gradient \(\hat g\) of the classifier score \(f\) by the promised distance \(d_p=|f(x)|/\lVert\nabla f(x)\rVert\), the length at which the linearized score reaches zero. We call this closed-form step \emph{alpha-1} (or \emph{one-shot}) \emph{recourse}, because it scales the promised distance by \(\alpha=1\) and involves no search. It costs one gradient per user. For a binary classifier it coincides with one DeepFool iteration \citep{moosavidezfooli2016deepfool}, and generators of the DeepFool type, which linearize and project, repeat it. When the score is curved along the path, the step can stop short and leave the user rejected (undershoot), or pass the boundary and ask for more change than needed (overshoot).

This paper explains when the closed-form step fails and how its failures shrink as a rule is allowed more model queries. With no query beyond the gradient, one can train with a curvature penalty or inflate the step by a factor tuned on validation data. One extra function value or Hessian-vector product per user, with a calibration set, allows a calibrated correction, and a line search along the ray finds the crossing to grid resolution. We ask what the cheaper rules can achieve when queries are limited. All rules are analyzed on the fixed path of the closed-form step, so for iterative generators, which re-linearize between steps, our results describe single steps only.

On the ray profile \(\phi(t)=f(x+t\hat g)\), alpha-1 is one Newton step from \(t=0\). A Newton step taken from below a root lands past it when the function is convex and short of it when the function is concave \citep{ostrowski1966solution,ortega2000iterative}, and the quadratic correction of Section~\ref{sec:sqr} is Euler's method \citep{traub1964iterative,melman1997geometry}. The recourse setting adds four things. (i)~Success is decided on the ray \(x+t\hat g\), and the nearest-boundary distance that robustness analysis usually studies does not decide it (Section~\ref{sec:diagnosing}). (ii)~The \emph{signed-curvature criterion} of Section~\ref{sec:signed} states that to leading order the alpha-1 step succeeds exactly when the path curvature \(\kappa=\hat g^\top\nabla^2f(x)\,\hat g\) is nonnegative. Theorem~5.2 bounds the crossing error from both sides, and Section~\ref{sec:results} measures how closely the population version holds on trained models. (iii)~Because an undershooting recommendation denies recourse, the loss is one-sided, and what a rule can guarantee under this loss depends on its query budget. A rule that sees only the score value and gradient cannot be both valid and economical over all scores with bounded path curvature (Theorem~6.1). When the path curvature is Lipschitz and the step is short, one extra function value at the promised point attains the minimax rate among deterministic one-query rules, with explicit constants (Theorem~6.3). (iv)~Split-conformal calibration gives any of these rules first-crossing coverage or abstention (Proposition~7.1). The experiments are diagnostics of the theory. They train with an asymmetric penalty on \(\kappa\), which raises alpha-1 ray-hit validity on undershoot-prone shallow data at a cost in overshoot (Sections~\ref{sec:penalties} and~\ref{sec:results}), run deep networks, and study a group gap in COMPAS and a deployment shift. Proofs are in supplements S1 and S6, and full tables in S4--S7.

\paragraph{Related work.} \emph{Root finding.} Besides the correspondences above, fixed inflation is DeepFool's overshoot factor \(1+\eta\) \citep{moosavidezfooli2016deepfool}. The forward-probe rule of Section~\ref{sec:sqr} uses the same information as one step of Ostrowski's two-step method (\(\phi\) and \(\phi'\) at \(0\), and \(\phi\) at the alpha-1 endpoint \(d_p\)) \citep{ostrowski1966solution}, a method whose fourth-order convergence \citet{kung1974optimal} conjectured to be optimal for three evaluations. Theorem~6.3 is a uniform, finite-scale counterpart with explicit constants and a one-sided loss, in the framework of optimal recovery and information-based complexity \citep{micchelli1977survey,traub1988ibc}. Decision-based attacks such as HopSkipJump \citep{chen2020hopskipjump} locate the boundary by bisection under a query budget, but they seek a small perturbation in any direction.

\emph{Recourse.} Optimization-based counterfactuals \citep{wachter2017counterfactual,mothilal2020dice,karimi2020model}, actionable and causal recourse \citep{ustun2019actionable,karimi2021algorithmic}, and conformal counterfactual generation \citep{altmeyer2024eccco} search for a recommendation at inference time, and recourse verification \citep{kothari2024prediction} tests whether any reachable action changes the decision. We instead fix the direction of a rule and study whether its prescribed step length succeeds. \citet{pawelczyk2022exploring} bound the distance between counterfactual explanations and adversarial examples, and \citet{fokkema2023attribution} prove that attribution-based explanations that provide recourse cannot be robust. Our impossibility result concerns first-order step rules on curved paths. Recourse has been hardened against or studied under model shift \citep{upadhyay2021robust,rawal2020recourse}, retraining \citep{black2022consistent,hamman2023robust}, dataset shift, where model curvature affects stability \citep{meyer2023minimizing}, execution noise \citep{pawelczyk2023probabilistic}, feature uncertainty \citep{dominguezolmedo2022adversarial} and model multiplicity \citep{turbal2025ellice}, as surveyed by \citet{jiang2024robustsurvey}. Work on fairness either audits or equalizes the cost of recourse across groups \citep{gupta2019equalizing,vonkugelgen2022fairness,sharma2020fairn}.

\emph{Curvature.} The curvature of the decision boundary bounds robustness to random noise \citep{fawzi2016robustness} and is small in most directions near data \citep{fawzi2018empirical}. Penalties on input curvature, the input Hessian or local nonlinearity \citep{moosavidezfooli2019robustness,qin2019adversarial,singla2020secondorder,peebles2020hessian,srinivas2022efficient} are unsigned, like our symmetric baselines, and \citet{dombrowski2019explanations} relate the fragility of gradient explanations to geometric properties of the network. Certified radii from curvature bounds \citep{singla2020secondorder} or from additive noise and randomized smoothing \citep{li2019certified,cohen2019certified} bound the distance to the boundary from below, whereas recourse needs a crossing within the recommended step along one direction, which depends on the sign of \(\kappa\).

\section{On-Path Recourse}
\label{sec:formal}

Let the deployed score \(f:\mathbb R^d\to\mathbb R\) be \(C^2\), with favorable set \(\{x:f(x)\ge0\}\), and fix a rejected point \(x\), so \(f(x)<0\). Write \(m=|f(x)|\), \(a=\lVert\nabla f(x)\rVert_2>0\), \(\hat g=\nabla f(x)/a\), \(d_p=m/a\), \(H=\nabla^2 f\), and \(\kappa=\hat g^\top H(x)\hat g\), the curvature of the score along the ray. The ray profile \(\phi(t)=f(x+t\hat g)\) satisfies \(\phi(0)=-m\), \(\phi'(0)=a\) and \(\phi''(0)=\kappa\). Three distances describe the geometry: the promised distance \(d_p\); the first-crossing distance \(d_{\text{ray}}=\inf\{t>0:\phi(t)\ge0\}\) along the ray (\(\inf\emptyset=+\infty\)); and the nearest-boundary distance \(d_\star=\inf\{\lVert x'-x\rVert_2:f(x')\ge0\}\), which satisfies \(d_\star\le d_{\text{ray}}\).

A recommendation of length \(t\) along \(\hat g\) is \emph{endpoint-valid} if \(f(x+t\hat g)\ge0\), and it \emph{hits} the boundary if \(d_{\text{ray}}\le t\); over rejected points, the frequencies of these events are the \emph{endpoint validity} and the \emph{ray-hit validity} of a rule. Endpoint validity implies a hit by continuity. The converse fails only if the path reaches the boundary before \(t\) and is unfavorable again at \(t\), an event whose probability we denote \(r_t\). The signed gap \(g=d_{\text{ray}}-d_p\) splits into undershoot \([g]_+=\max(g,0)\) and overshoot \([g]_-=\max(-g,0)\). Four baseline rules recur: \emph{alpha-1} recommends \(d_p\hat g\); \emph{fixed inflation} recommends \(\alpha d_p\hat g\) for a chosen \(\alpha>1\); \emph{validation-tuned inflation} selects one global \(\alpha\) on held-out data; and \emph{ray line search} evaluates \(f\) along the ray to find \(d_{\text{ray}}\) for each user.

Unless an action set is specified, these gradient rays are continuous relaxations in feature space and do not describe implementable or causal actions. For a convex cone of feasible directions, Lemma~5.1, Theorems~5.2 and~6.1 and Proposition~7.1 transfer to the steepest feasible direction up to the first binding constraint, if a feasible ascent direction exists (supplement S6).

Because \(d_p\) and \(\kappa\) are computed from the score, it matters how they behave under a \emph{score gauge}, a strictly increasing rescaling \(\psi\) of the score with \(\psi(0)=0\), which leaves every decision unchanged.

\noindent\textbf{Proposition 2.1 (score gauge and action map).}
\textit{Let \(h:\mathbb R^k\to\mathbb R^d\) be a \(C^2\) map from action coordinates \(z\) to features, with Jacobian \(J_h\) and second derivative \(D^2h\) (\(h=\mathrm{id}\) if no action map is given). Let \(f\) be \(C^2\) near \(x=h(z)\), put \(F=f\circ h\), and fix \(z\) with \(F(z)\ne0\), \(a=\|\nabla_zF(z)\|>0\) and \(v=\nabla_zF(z)/a\). Let \(\psi\) be \(C^2\) and strictly increasing with \(\psi(0)=0\) and \(\psi'(F(z))>0\). Then \(\psi\circ F\) has the same decisions, boundary and normalized ray as \(F\), and}
\[
\begin{aligned}
a_\psi&=\psi'(F(z))a,&
d_{p,\psi}&=\frac{|\psi(F(z))|}{\psi'(F(z))a},\\
\kappa_\psi&=\psi'(F(z))\kappa_{\rm eff}+\psi''(F(z))a^2,
\end{aligned}
\]
\textit{where \(\kappa_{\rm eff}=(J_hv)^\top\nabla^2f(x)(J_hv)+\langle\nabla f(x),D^2h[v,v]\rangle\), which equals \(\kappa\) when \(h=\mathrm{id}\). If moreover \(\psi'(s)>0\) for every \(s\ne0\) and \(d_{p,\psi}=d_p\) at every nonzero score, then \(\psi(s)=cs\) for some \(c>0\).}

A gauge changes \(m\), \(a\), \(d_p\) and \(\kappa\) but leaves \(d_{\text{ray}}\) fixed, so \(\kappa\) is a property of the deployed score that the decision boundary alone does not determine. Supplement S1 gives a family of gauges under which \(d_p\) takes every positive value and \(\kappa\) either sign. On each of COMPAS, German and Adult, a fixed grid of such gauges moves alpha-1 endpoint validity across the whole range from 0 to 1 (supplement S1). Every result below concerns the score as deployed.

\section{Nearest-Boundary Distance as a Diagnostic}
\label{sec:diagnosing}

We use \(d_\star\), which robustness analyses usually measure, only as a diagnostic, because \(d_\star\le d_{\text{ray}}\) means that a small \(d_\star\) cannot show that the recommended step succeeds. A common estimator of \(d_\star\) is also tied to the quantity it would audit, since an unrefined DeepFool iterate moves by \((|f|/\lVert\nabla f\rVert^2)\nabla f=d_p\hat g\), which is the alpha-1 step itself. We therefore estimate \(d_\star\) with a Carlini--Wagner-style objective \citep{carlini2017towards} that does not use \(d_p\) (supplement S2), and no validity or first-crossing quantity uses \(d_\star\).

\section{Curvature Bounds the Nearest-Boundary Gap}
\label{sec:frobenius}

Let \(\mathcal N\) be a ball about \(x\) that contains the alpha-1 endpoint \(y=x+d_p\hat g\) and a closest boundary point, on which \(f\) is \(C^2\) with \(\lVert\nabla f\rVert\ge\gamma>0\). Assume that the normalized gradient flow \(\dot\eta=-\operatorname{sign}(f(y))\nabla f(\eta)/\lVert\nabla f(\eta)\rVert^2\), \(\eta(0)=y\), has a solution in \(\mathcal N\) on \([0,|f(y)|]\).

\noindent\textbf{Theorem 4.1 (curvature bound).}
\textit{If \(\lVert H\rVert_2\le L\) on \(\mathcal N\), then \(|d_\star-d_p|\le Ld_p^2/(2\gamma)\).}

Theorem~4.1 is blind to the sign of the curvature. Curvature bounds of this kind on the distance to the boundary are known \citep{fawzi2016robustness,moosavidezfooli2019robustness}. Its hypothesis holds when \(\lVert H\rVert_F\le L\) on \(\mathcal N\); the Hutchinson penalties of Table~\ref{tab:main} only reduce an average of \(\lVert H\rVert_F^2\) over training points, but they do shrink the nearest-boundary gap. The flow assumption is strong, since it requires the boundary to be reachable from \(y\) within a region where the gradient stays away from zero.

\section{The Signed-Curvature Criterion}
\label{sec:signed}

Fix a rejected \(x\) with the notation of Section~\ref{sec:formal}, and let \(M\) be a Lipschitz constant of \(\phi''\) on \([0,d_p]\). Because \(ad_p=m\), the linear term cancels at the promised distance, and Taylor's theorem gives the \emph{endpoint identity}
\[
f(x+d_p\hat g)=\tfrac{\kappa}{2}d_p^2+R,\qquad |R|\le\tfrac{M}{6}d_p^3.
\]
The alpha-1 step is therefore endpoint-valid when \(\kappa>Md_p/3\) and invalid when \(\kappa<-Md_p/3\); only points in the band \(|\kappa|\le Md_p/3\) are undetermined. Over a population of rejected points, with \(\kappa\), \(M\) and \(d_p\) varying by point (\(M=\infty\) where \(\phi''\) is not Lipschitz), alpha-1 endpoint validity equals \(\mathbb P(\kappa\ge0)\) up to the probability of this band (supplement, Corollary~\ref{S-cor:s1-ambiguity}). We call this the \emph{signed-curvature criterion}. Because \(d_p\) is the Newton iterate from \(t=0\) for the root of \(\phi\), the criterion is the local form of the convexity condition for Newton's method recalled in Section~\ref{sec:intro}. Lemma~5.1 and Theorem~5.2 bound the size of the miss. Theorem~5.2 is a two-sided, finite-scale form of the classical error expansion of a Newton step, in which the new error is about \(\phi''/(2\phi')\) times the square of the old one.

\paragraph{Lemma 5.1 (quadratic model).}
\textit{Let \(q(t)=-m+at+\tfrac\kappa2t^2\) and \(s=2\kappa d_p/a>-1\). The smallest positive root \(d_{\text{ray}}^q\) of \(q\) satisfies}
\[
d_{\text{ray}}^{q}-d_p=-\frac{\kappa d_p^2}{2a}\rho(s),
\qquad
\rho(s)=\frac{4}{(1+\sqrt{1+s})^2}.
\]
\textit{If \(H\) is constant along the ray, then \(d_{\text{ray}}^q=d_{\text{ray}}\).}

\paragraph{Theorem 5.2 (two-sided on-path bound).}
\textit{Assume that \(\phi''\) is \(M\)-Lipschitz on \([0,2d_p]\) and that regime \((\star)\) holds: \(4|\kappa|d_p\le a\) and \(8Md_p^2\le a\). Then \(\phi\) has a first root \(d_{\text{ray}}\le2d_p\), and \(g=d_{\text{ray}}-d_p\) satisfies}
\[
\left|g+\frac{\kappa d_p^2}{2a}\rho(s)\right|\le\frac{8Md_p^3}{3a}.
\]

The hypothesis on \(\phi''\), which implies the one on \([0,d_p]\), holds when the Hessian is \(M\)-Lipschitz on the segment \(\{x+t\hat g:t\in[0,2d_p]\}\), and regime \((\star)\) says that the step is short relative to the curvature scales. Under \((\star)\), \(\phi\) is strictly increasing on \([0,2d_p]\) (supplement, Lemma~\ref{S-lem:s1-no-recross}), so \(r_t=0\) and endpoint and ray validity coincide for every recommendation up to \(2d_p\).

\begin{figure}[H]
\centering
\includegraphics[width=0.6\textwidth]{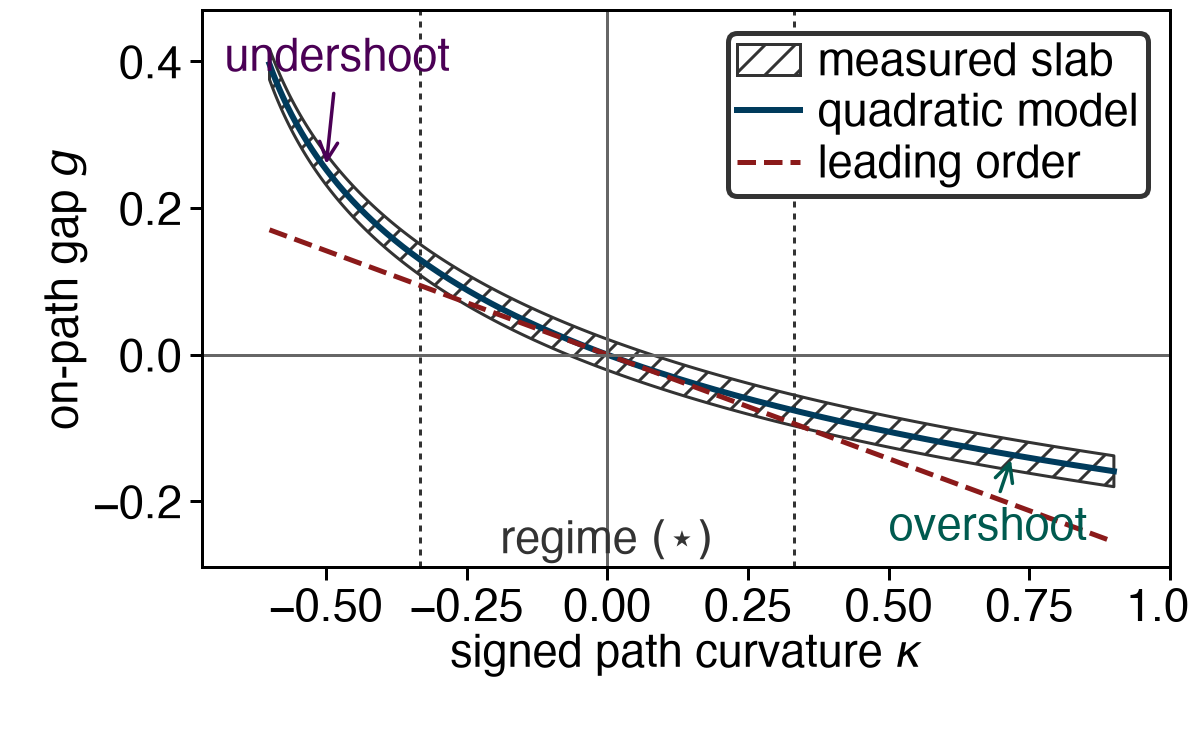}
\caption{Quadratic-model gap \(d_{\text{ray}}^q-d_p\) of Lemma 5.1 (equal to \(g\) when \(\phi''=\kappa\) on \([0,d_{\text{ray}}^q]\)), its leading-order approximation, and the \emph{measured} two-sided slab \(\pm8\hat Md_p^3/(3a)\) versus \(\kappa\), at the COMPAS-unregularized median \((\hat M,d_p,a)=(0.018,0.75,1.00)\) (\(n{=}800\) points). As \(\hat M\) is a finite-grid lower bound, the slab is an empirical diagnostic, not a certified remainder; per point it contains \(99\%\) of the points in regime \((\star)\) (certified enclosure in S3.1).}
\label{S-fig:supp-twosided}
\end{figure}

Since \(\rho\) is decreasing with \(\rho(0)=1\), the naive term \(-\kappa d_p^2/(2a)\) underestimates undershoot when \(\kappa<0\) and overestimates overshoot when \(\kappa>0\) (supplement, Figure~\ref{S-fig:supp-twosided}). Theorem~5.2 gives \([g]_+\ge|\kappa|d_p^2/(2a)-8Md_p^3/(3a)\) when \(\kappa<0\) and \([g]_+\le8Md_p^3/(3a)\) when \(\kappa\ge0\). Undershoot is thus of order \(|\kappa|d_p^2/a\) on concave paths once the remainder is small, and at most the remainder on convex ones. The asymmetric penalty of Section~\ref{sec:penalties} therefore pushes \(\kappa\) above a small target on the rejected side.

\section{The First-Order Barrier}
\label{sec:barrier}

Fixed and tuned inflation use only the score value and gradient at the user's point, so they cannot tell whether the path bends up or down. Fix \(f(x)=-m\), \(\nabla f(x)=a\hat g\) and \(d_p=m/a\), and let \(\mathcal F_K(x)\) be the set of \(C^2\) scores with this value and gradient whose ray profile satisfies \(|\phi''|\le K\) on \([0,2d_p]\). A \emph{first-order rule} recommends \(d_{\text{rec}}=r(f(x),\nabla f(x))\).

\noindent\textbf{Theorem 6.1 (first-order barrier).}
\textit{Let \(K>0\) with \(Kd_p<a/2\), and put \(u=2Kd_p/a\). The ray profiles \(\phi_\pm(t)=-m+at\pm\tfrac K2 t^2\) are those of scores in \(\mathcal F_K(x)\) and satisfy \(d_{\text{ray}}(\phi_+)<d_p<d_{\text{ray}}(\phi_-)\), with}
\[
\begin{aligned}
d_{\text{ray}}(\phi_-)-d_{\text{ray}}(\phi_+)
&=\tfrac aK\big(2-\sqrt{1+u}-\sqrt{1-u}\big)\\
&\in\big[\tfrac{Kd_p^2}{a},\,\tfrac{Kd_p^2}{a}+6(Kd_p/a)^3 d_p\big].
\end{aligned}
\]
\textit{Consequently, every first-order rule is either invalid on \(\phi_-\) or overshoots \(\phi_+\) by at least this gap.}

Two scores can share the value and gradient at \(x\) and bend in opposite directions along the ray, so a rule that reads only these must fail one user or overshoot for the other. The theorem is a worst case over \(\mathcal F_K(x)\) at a single point. Typical populations, rules that use more information and training that changes the distribution of \(\kappa\) fall outside it.

\noindent\textbf{Corollary 6.2 (one forward query).} \textit{Let \(\phi''\) be \(M\)-Lipschitz on \([0,d_p]\). The probe value \(f(x+d_p\hat g)\) gives \(\hat\kappa=2f(x+d_p\hat g)/d_p^2\) with \(|\hat\kappa-\kappa|\le Md_p/3\). Under the hypotheses of Theorem~5.2, the smallest positive root of the quadratic that matches \(\phi\) in value and slope at \(0\) and in value at \(d_p\) lies within \(\varepsilon_\star=(8/3+C_0/3)Md_p^3/a\) of \(d_{\text{ray}}\), where \(C_0<1.15\). Under the hypotheses of Theorem~6.3 below, the error is at most \(4M|\kappa|d_p^4/a^2+\tfrac43M^2d_p^5/a^2\).}

Adding the margin \(\varepsilon_\star\) gives deterministic first-crossing validity when \(M\) is known; Section~\ref{sec:sqr} calibrates a margin from data instead.

\begin{samepage}
\noindent\textbf{Theorem 6.3 (one-query minimax rate).} \textit{Fix \(a,d_p,K_0,M>0\) with \(K_0d_p/a\le1/8\) and \(Md_p^2/a\le1/8\), and let \(\mathcal G\) be the set of ray profiles \(\phi\in C^{2,1}([0,\infty))\) with \(\phi(0)=-ad_p\), \(\phi'(0)=a\), \(|\phi''(0)|\le K_0\) and \(\operatorname{Lip}(\phi'')\le M\). Consider deterministic rules that know \((a,d_p,K_0,M)\), evaluate \(\phi\) once at a point \(t_1\ge0\) chosen from this information alone, and output \(\widehat r\). Every \(\phi\in\mathcal G\) has exactly one root in \([0,2d_p]\), which is \(d_{\text{ray}}(\phi)\). The minimax error \(\sup_{\mathcal G}|\widehat r-d_{\text{ray}}|\), and the minimax overshoot \(\sup_{\mathcal G}(\widehat r-d_{\text{ray}})\) over rules with \(\widehat r\ge d_{\text{ray}}\) on all of \(\mathcal G\), are both \(\Theta(MK_0d_p^4/a^2+M^2d_p^5/a^2)\). The probe \(t_1=d_p\) with an explicit margin attains this rate.}
\end{samepage}

Probing at the promised distance is thus rate-optimal among deterministic one-query rules, both for estimating the crossing and for guaranteeing it. The lower and upper constants for estimation, \(1/2592\) and \(4\), differ by more than \(10^4\), so only the rate is settled, and randomized queries are outside the theorem. With \(\kappa\) revealed and no function value, the best worst-case error is of order \(Md_p^3/a\) (supplement, Corollary~\ref{S-cor:s1-lattice}; Theorem~\ref{S-thm:s1-batch} treats \(q\ge3\) queries), which is consistent with the advantage of the forward probe over the Hessian-vector product in Section~\ref{sec:sqr}.

\begin{table*}[t]
\centering
\small
\caption{Symmetric regularizers: nearest-boundary gap and on-path validity (10 seeds; mean \(\pm\) population standard deviation). Ray metrics use all rejected test points; \(d_\star\) is a C\&W-style solver estimate on a seeded random subset of at most 128 rejected points per seed. Validity is ray-hit validity, \(d_{\text{ray}}\le d_p+10^{-6}\). The gap columns are seed means; Theorem~4.1 bounds the first pointwise, and Theorem~5.2 the second after its leading term is removed. Spectral Norm validity is bimodal across seeds (Section~\ref{sec:results}).}
\label{tab:main}
\begin{tabular}{llrrrr}
\hline
Dataset & Method & Bal. Acc. (\%) & \(|d_\star-d_p|\) (nearest) & \(|d_{\text{ray}}-d_p|\) (on-path) & Validity (\%) \\
\hline
COMPAS & Unregularized & \(66.83\pm0.58\) & \(0.0543\pm0.0111\) & \(0.0697\pm0.0123\) & \(73.58\pm3.34\) \\
COMPAS & Spectral Norm & \(65.41\pm0.52\) & \(0.0199\pm0.0087\) & \(0.0206\pm0.0085\) & \(10.90\pm29.72\) \\
COMPAS & 1-Lipschitz GP & \(66.68\pm0.56\) & \(0.0114\pm0.0049\) & \(0.0141\pm0.0046\) & \(31.65\pm15.33\) \\
COMPAS & Global Hutchinson & \(66.59\pm0.32\) & \(0.0123\pm0.0031\) & \(0.0105\pm0.0019\) & \(90.94\pm4.32\) \\
COMPAS & MW-Hutchinson & \(66.60\pm0.43\) & \(0.0134\pm0.0030\) & \(0.0151\pm0.0019\) & \(84.12\pm2.95\) \\
German & Unregularized & \(73.38\pm0.79\) & \(0.0249\pm0.0059\) & \(0.0220\pm0.0057\) & \(100.00\pm0.00\) \\
German & Spectral Norm & \(70.13\pm1.68\) & \(0.0072\pm0.0050\) & \(0.0094\pm0.0063\) & \(26.72\pm35.56\) \\
German & 1-Lipschitz GP & \(73.13\pm1.00\) & \(0.0097\pm0.0041\) & \(0.0075\pm0.0037\) & \(96.65\pm6.02\) \\
German & Global Hutchinson & \(73.74\pm0.76\) & \(0.0160\pm0.0019\) & \(0.0141\pm0.0018\) & \(100.00\pm0.00\) \\
German & MW-Hutchinson & \(73.74\pm0.86\) & \(0.0184\pm0.0026\) & \(0.0162\pm0.0025\) & \(100.00\pm0.00\) \\
Adult & Unregularized & \(81.93\pm0.26\) & \(0.0695\pm0.0133\) & \(0.1030\pm0.0110\) & \(0.26\pm0.13\) \\
Adult & Spectral Norm & \(79.69\pm0.52\) & \(0.2823\pm0.1079\) & \(0.2189\pm0.0815\) & \(99.80\pm0.15\) \\
Adult & 1-Lipschitz GP & \(81.25\pm0.15\) & \(0.4648\pm0.0929\) & \(0.1831\pm0.0166\) & \(90.50\pm2.59\) \\
Adult & Global Hutchinson & \(82.03\pm0.15\) & \(0.0142\pm0.0019\) & \(0.0140\pm0.0012\) & \(43.63\pm12.13\) \\
Adult & MW-Hutchinson & \(82.04\pm0.24\) & \(0.0174\pm0.0042\) & \(0.0184\pm0.0021\) & \(26.32\pm7.26\) \\
\hline
\end{tabular}
\end{table*}

\section{Signed-Quadratic Recourse and Conformal Calibration}
\label{sec:sqr}

Theorem~5.2 suggests a correction. Under its hypotheses the root of the quadratic model, \(d_{\text{quad}}=d_{\text{ray}}^q=2d_p/(1+\sqrt{1+2\kappa d_p/a})\), is within \(8Md_p^3/(3a)\) of \(d_{\text{ray}}\), while the alpha-1 error is of order \(|\kappa|d_p^2/a\) once this remainder is small against \(|\kappa|d_p^2/(2a)\). This is one step of Euler's method, which takes the root of the second-order Taylor polynomial \citep{traub1964iterative,melman1997geometry}. \emph{Signed-quadratic recourse} recommends \(d_{\text{quad}}\) from one gradient and one Hessian-vector product (HVP) with \(\hat g\) held fixed. It lengthens the step on concave paths and shortens it on convex ones. When \(1+2\kappa d_p/a\le0\) it returns \(2d_p\), the limit of the formula (supplement S3.2). The \emph{forward-probe} variant replaces \(\kappa\) by \(\hat\kappa\) from Corollary~6.2. It costs one extra evaluation of \(f\), needs no HVP, and can be computed when \(f\) is not twice differentiable. Bare signed-quadratic recourse, a point estimate of \(d_{\text{ray}}\), has mean ray-hit validity of only 28.9\% in our suite (Table~\ref{tab:deploy}), so it needs a margin.

Split conformal prediction \citep{lei2018distribution} calibrates the margin. Fix a base rule \(d_{\text{base}}\) before calibration. On a held-out set of \(n\) rejected points, form residuals \(r_i=d_{\text{ray},i}-d_{\text{base},i}\), with \(r_i=+\infty\) when the ray search finds no crossing, and let \(\hat q_{1-\delta}\) be their \(\lceil(1-\delta)(n+1)\rceil\)-th smallest value (\(+\infty\) if this index exceeds \(n\)). If \(\hat q_{1-\delta}\) is finite, the rule recommends \(d_{\text{conf}}=\max\{0,d_{\text{base}}+\hat q_{1-\delta}\}\); otherwise it abstains.

\noindent\textbf{Proposition 7.1 (coverage or abstention).} \textit{Let \(\delta\in(0,1)\), suppose that the \(n\) calibration residuals and the test residual are exchangeable, and that \(d_{\text{base}}\) is fixed independently of the calibration data. Then \(\Pr[\text{abstain or }d_{\text{ray}}\le d_{\text{conf}}]\ge1-\delta\). If the \(n+1\) pairs of residual and protected attribute \(A\) are exchangeable and \(A\) takes finitely many values, per-group quantiles give the same bound given \(A=c\) (supplement S1).}

Proposition~7.1 is split conformal prediction with a one-sided residual, its per-group version is Mondrian conformal prediction \citep{vovk2005algorithmic}, and the proof is the usual rank argument. The probability that the rule issues a finite step that stops short of the first crossing is at most \(\delta\) marginally over calibration and test draws, so the coverage realized on one model's test points can fall below \(1-\delta\). A finite recommendation requires \(n\ge\lceil1/\delta\rceil-1\), which is necessary but not sufficient, since unbracketed calibration rays enter as \(+\infty\). In our experiments \(d_{\text{ray}}\) is the output of the ray search (a uniform grid of spacing \(A/159\) on \([0,A]\) with \(A\approx8d_p\), then bisection; supplement S3.2), so coverage refers to that estimate, and a crossing that the grid steps over is missed. At test points that satisfy the hypotheses of Theorem~5.2 and have \(d_{\text{conf}}\le2d_p\), coverage also gives endpoint validity, and elsewhere we check it empirically (supplement S2).

Every base rule inherits Proposition~7.1, whether it is alpha-1 (\(d_{\text{base}}=d_p\)), the signed quadratic or the probe quadratic, so an extra query changes the base rule and, through it, the overshoot at a given nominal coverage. We report \emph{conformal-quadratic} and \emph{conformal-probe}. Validation-tuned inflation is close to a conformal rule with residual \(d_{\text{ray}}/d_p\), but our implementation omits the finite-sample correction and caps \(\alpha\) at 3.

\paragraph{Empirical check.} At \(\delta=0.05\) on the 80 models of Section~\ref{sec:results}, conformal-probe has lower mean overshoot than conformal-quadratic in all 16 dataset\(\times\)method cells at similar pooled validity (\(96.0\%\) versus \(95.5\%\); Table~\ref{tab:deploy}), so at a common nominal target it is the one-query rule we recommend. Conformal-quadratic has lower mean overshoot than validation-tuned inflation in 15 of the 16 cells (each the median of five seeds), the exception being unregularized Adult; in 7 cells alpha-1 already meets the \(0.95\) target and tuned inflation reduces to \(\alpha=1\), and conformal-quadratic wins 8 of the other 9. The two rules share a nominal target, yet pooled over the suite conformal-quadratic reaches \(95.5\%\) ray-hit validity and tuned inflation \(98.2\%\). Part of the overshoot saving therefore comes from validity closer to the target, and the two rules are not compared at equal realized validity.

\begin{table}[t]
\centering
\footnotesize
\setlength{\tabcolsep}{3pt}
\caption{Rules by the model queries they need per user beyond the gradient, mean over \(n\) models of the five-seed suite (4 datasets \(\times\) 4 training methods \(\times\) 5 seeds) at target \(0.95\): ray-hit validity and mean overshoot. Tuned inflation needs a validation set and the conformal rules a calibration set. Rows differ in realized validity; the last row, alpha-1 on the 20 asymmetric-penalty models among the 80, is a training-time intervention and not comparable with the others.}
\label{tab:deploy}
\begin{tabular}{llrrr}
\hline
Queries & Rule & \(n\) & Val.\,\% & Oversh. \\
\hline
none & alpha-1 & 80 & 76.6 & 0.041 \\
none & tuned inflation & 80 & 98.2 & 0.061 \\
1 HVP & signed-quadratic & 80 & 28.9 & 0.005 \\
1 HVP & conformal-quadratic & 80 & 95.5 & 0.025 \\
1 forward & conformal-probe & 80 & 96.0 & 0.0016 \\
189 forward & ray line search \(=d_{\text{ray}}\) & 80 & 100.0 & 0.000 \\
\hline
training & asymmetric, alpha-1 & 20 & 99.6 & 0.103 \\
\hline
\end{tabular}
\end{table}

\section{Penalties}
\label{sec:penalties}

We use Softplus activations so that input Hessians carry information; on piecewise-affine ReLU networks the input Hessian vanishes almost everywhere and HVP penalties are vacuous. The margin-weighted (MW) Hutchinson penalty estimates \(\lVert H\rVert_F^2\) with one Rademacher vector \(v\) (one input HVP) per example and step \citep{hutchinson1989stochastic}, where forming \(H\) would take \(d\) HVPs (784 on F-MNIST), and down-weights points far from the boundary:
\[
\mathcal L=\mathcal L_{\text{BCE}}+\lambda\mathbb E_x\!\left[
e^{-|f|/\tau}\mathbb E_v\lVert\nabla_x(\nabla_x f^\top v)\rVert_2^2
\right],
\]
with \(\mathbb E_x\) over training points, \(\mathbb E_v\) over Rademacher \(v\) and \(\tau=\mathrm{std}(\text{logits})\); Global Hutchinson uses weight~1.

\noindent\textbf{Remark 8.1 (fixed-ray leading statistic).} If two scores have the same value and gradient at \(x\) (hence the same \(d_p\) and \(\hat g\)), the same \(\kappa\), and \(C^{2,1}\) ray profiles whose second derivatives share a Lipschitz constant \(M\), then in regime \((\star)\) Theorem~5.2 and the triangle inequality place their first crossings within \(16Md_p^3/(3a)\) of each other.

On a fixed ray, validity and the on-path gap therefore depend on the Hessian at \(x\) only through \(\kappa\), up to the remainder. A Frobenius or Hutchinson penalty acts on all of \(H\), including curvature off the ray, and it is even under \(H\mapsto-H\), so it cannot prefer \(\kappa>0\) to \(\kappa<0\) and pushes \(\kappa\) toward zero from both sides. The \emph{asymmetric penalty} acts on \(\kappa\) alone. It computes \(\kappa\) with one input HVP per example (\(\hat g\) held fixed; training cost in supplement S3) and penalizes \([\delta-\kappa]_+^2\), weighted by \(e^{-[f]_+/\tau}\) toward the rejected side. Its curvature target \(\delta\ge0\), distinct from the miscoverage level of Section~\ref{sec:sqr}, sets the balance between undershoot and overshoot. An ablation in supplement S4 separates the ingredients. A matched sign-blind \(\kappa^2\) penalty is unstable and harmful on COMPAS, while the sign twin \((\delta-\kappa)^2\), which lacks the hinge, matches the asymmetric penalty except on Adult, where the hinge matters (\(93\!\to\!99.4\%\)).

A penalty that raises \(\kappa\) can also act through the scale of the score. By Proposition~2.1, a convex gauge \(\psi\) raises the curvature relative to the slope, \(\kappa_\psi/a_\psi=\kappa/a+a\,\psi''(f(x))/\psi'(f(x))\ge\kappa/a\), so it can make a negative \(\kappa\) nonnegative but never the reverse, and it lengthens \(d_p\) at every rejected point (supplement S1). A penalty that raises \(\kappa\) can therefore act in part like a learned, point-dependent inflation of the step, and alpha-1 validity under the penalty does not separate a change in the geometry of the path from a change of scale. The overshoot grows with any such inflation, so a validity gain at \(\alpha=1\) is informative only together with its overshoot.

\section{Experimental Setup}
\label{sec:setup}

\textbf{Data.} COMPAS (ProPublica two-year recidivism \citep{angwin2016machinebias}; \(n=5278\), race in \{African-American, Caucasian\}, \(d=11\)); German Credit (Statlog/UCI \citep{dua2017uci}, \(n=1000\), \(d=48\)); Adult (UCI \citep{dua2017uci}; full public 14-feature schema, subsampled to 20{,}000 rows, or to a nested 8{,}000-row cohort for the suites marked 8k in supplement S3.2; \(d=100\) after one-hot encoding); and Fashion-MNIST (F-MNIST) \citep{xiao2017fashion} (Pullover versus Coat, \(n=8000\), \(d=784\)). Protected attributes are COMPAS race, German sex and Adult gender.

\textbf{Models.} We use an MLP with two Softplus layers \([128,64]\) (\([256,128]\) for F-MNIST), Adam at \(10^{-3}\), batch size 256, and \texttt{BCEWithLogitsLoss} with positive-class weight equal to the training split's negative-to-positive ratio (the F-MNIST models of the penalty comparison use unweighted BCE). Training runs 50 epochs, except F-MNIST (15) and the 8{,}000-row Adult cohort (30). Table~\ref{tab:main} compares Unregularized, Spectral Norm \citep{miyato2018spectral}, 1-Lipschitz Gradient Penalty (GP; \citealp{gulrajani2017improved}), Global Hutchinson and MW-Hutchinson, with \((\lambda_{\rm GP},\lambda_{\rm Global},\lambda_{\rm MW})=(0.5,0.2,0.2)\) on COMPAS, \((0.05,0.05,0.05)\) on German, and \((0.5,0.05,0.05)\) on Adult. The \texttt{StandardScaler} is fit on the training split only. Results come from separately trained suites, listed with their models, seeds and caps in supplement S3.2: the ten-seed main comparison (Table~\ref{tab:main}); a five-seed suite of 80 models for the criterion, the recourse rules and Table~\ref{tab:deploy}; the Adult sweep and a three-seed COMPAS penalty suite; F-MNIST (five seeds) and CIFAR-10 (ten); 20 resplits; and three- and five-seed COMPAS case-study suites.

\textbf{Metrics.} We measure endpoint validity directly for the criterion and the score-gauge audit. Every other validity we report is ray-hit validity at the recommended length \(t\), evaluated as \(d_{\text{ray}}\le t+10^{-6}\). The two events coincide when \(\phi\) is monotone on \([0,t]\), and elsewhere we measure their difference \(r_t\). On the 80 models of the five-seed suite they agree on \(99.55\%\) of audited test points, and every disagreement is a tolerance case at the boundary with no re-crossing (supplement S2). Undershoot and overshoot are computed from \(d_{\text{ray}}-d_p\).

\section{Results}
\label{sec:results}

\paragraph{Signed-curvature criterion.} Across the 80 models of the five-seed suite, alpha-1 endpoint validity \(V\) and \(\mathbb P(\kappa\ge0)\), computed on the same held-out rejected points, correlate at Pearson \(r=0.985\). The suite covers four datasets (COMPAS, German, Adult, F-MNIST), four training methods (unregularized, MW-Hutchinson, Global Hutchinson, asymmetric) and five seeds. A bootstrap over its 16 dataset\(\times\)method cells gives the 95\% interval \([0.958,0.997]\), although it ignores dependence between cells of one dataset, and the minimum leave-one-dataset-out \(r\) is \(0.966\). A correlation across models can hide level offsets. Over the 20 models of each dataset, the mean of \(|V-\mathbb P(\kappa\ge0)|\) is 1.0, 0.2, 3.1 and 8.2 points on COMPAS, German, Adult and F-MNIST, and its maximum is 3.6, 2.4, 8.5 and 20.5 points. On F-MNIST the criterion is systematically optimistic: \(V\) lies below \(\mathbb P(\kappa\ge0)\) by 8.2 points on average and by 17.5 points for the unregularized models. Point by point, the sign of \(\kappa\) agrees with endpoint success on 94.7\% of the 24{,}189 held-out points, and on 82.5\% for unregularized F-MNIST. Endpoint success is exactly the event \(\hat\kappa\ge0\), with \(\hat\kappa=2f(x+d_p\hat g)/d_p^2\) a weighted average of \(\phi''\) over the step (supplement, Remark~\ref{S-prop:s1-probe-exact}), so a disagreement is a point where \(\phi''\) drifts enough along the step to change the sign of this average. Section~\ref{sec:signed} bounds each model's offset by the probability of the band \(|\kappa|\le Md_p/3\). Under a conditional floating-point enclosure of \(M\), this bound holds on a separate drift-screen sample of unregularized COMPAS, German and Adult models (supplement S1). F-MNIST has no enclosure of \(M\), so the bound cannot be checked there.

\paragraph{Main comparison.} Curvature regularization reduces the nearest-boundary gap \(|d_\star-d_p|\) (Table~\ref{tab:main}), as Theorem~4.1 suggests, but a small gap does not bring high ray-hit validity; unregularized Adult reaches \(0.3\%\) and the Hutchinson variants 26--44\%. Spectral Norm and the gradient penalty reach high Adult validity (\(99.8\%, 90.5\%\)), but their nearest-boundary gaps grow to \(0.28\)--\(0.46\) (\(0.014\)--\(0.017\) for the Hutchinson variants), and they fail on COMPAS. No method is best on every dataset, and margin weighting does not beat its unweighted ablation in Table~\ref{tab:main} (Global Hutchinson is better on COMPAS and Adult). The largest standard deviations come from seeds that split into a low cluster and a few high values (every ray found a crossing). Spectral Norm's validity is \(0.3\)--\(3.6\%\) on nine COMPAS seeds and \(100\%\) on the tenth, and \(0\)--\(2.5\%\) on six German seeds and \(25\)--\(96\%\) on the other four.

\paragraph{Asymmetric penalty.} The asymmetric penalty trades overshoot for validity at \(\alpha=1\). German has no undershoot and stays near 100\%. The primary evidence is an audit over \(20\) stratified resplits with per-split records and hyperparameters chosen on validation data, in which the asymmetric penalty beats the validation-selected best symmetric penalty on every COMPAS, Adult and F-MNIST resplit and on \(13/20\) German resplits, at similar mean accuracy; this audit did not record overshoot. The resplits overlap, so they measure robustness to the split and are not independent replications. Three further suites, an Adult penalty sweep, F-MNIST (five seeds) and a separately retrained three-seed COMPAS suite, were stored only as printed summaries (supplement S10). On Adult, the symmetric penalties at \(\lambda\in\{0.05,1.0\}\) reach at most 70.9\% (MW-Hutchinson) and 43.4\% (Global) ray-hit validity, both at \(\lambda=1.0\), and the asymmetric penalty at \(\delta=0.05,0.1,0.2\) reaches 99.4\%, 100.0\% and 100.0\%, with balanced accuracy 81.4--81.5\% and overshoot 0.038--0.092. On F-MNIST it reaches \(99.9\%\) against 74.7\% and 68.8\% for the symmetric penalties, while mean overshoot rises from \(\sim\!0.05\)--\(0.07\) to \(0.22\), about four times as much. On COMPAS (unregularized \(75.2\%\)) it reaches \(99.5\%\) at \(\delta=0.1\) (\(96.3\%\) at \(0.05\)) against \(86.5\%\) for MW-Hutchinson, the only symmetric arm, at flat accuracy and mean overshoot \(0.111\) against \(0.005\). All of these comparisons are at \(\alpha=1\), where the gain can include the lengthening of \(d_p\) described in Section~\ref{sec:penalties}. A smaller matched-validity audit of briefly trained models finds no uniform advantage over validation-tuned inflation (supplement S5), and on Adult the five-seed symmetric penalties with tuned inflation nearly match it at a fifth of the overshoot.

\paragraph{Deep networks.} On CIFAR-10 \citep{krizhevsky2009learning} automobile versus truck (ten seeds on two GPU types; setup in supplement S4), curvature-aware training raises alpha-1 ray-hit validity from \(81.0\%\) for the unregularized models to \(95.4\%\) with Global Hutchinson and \(91.6\%\) with the asymmetric penalty (overshoot \(4.644\) and \(1.003\) vs.\ \(1.102\), accuracy \(93.1\) and \(92.0\%\) vs.\ \(94.5\%\)). One asymmetric run, at curvature target \(\delta=0\), collapsed in training; in a separate ten-seed run at \(\delta=0.05\) no seed collapsed and Global still led on the mean, within the seed spread (supplement, Table~\ref{S-tab:supp-deep-probe-negative}). A training penalty built on the forward probe instead of the HVP ended as a chance-level classifier in every full-length run, with Softplus and with GroupNorm--ReLU networks. Because GroupNorm is not piecewise affine, these runs did not test the ReLU case in which HVP penalties vanish (supplement S4).

\paragraph{Case study: a group gap on COMPAS.} On the unregularized models of the three-seed COMPAS per-point suite, alpha-1 ray-hit validity is \(67\%\) for African-American defendants (the disadvantaged group) and \(89\%\) for Caucasian defendants, and \(\mathbb P(\kappa\ge0)\) is 0.66 versus 0.87. Because the same test people recur across seeds, we report no significance tests. Reweighting the disadvantaged group's validities on \(\{\kappa\ge0\}\) and \(\{\kappa<0\}\) to the advantaged group's \(\mathbb P(\kappa\ge0)\) closes \(87.9\%\) of the gap, which follows from the criterion and is not separate evidence. The disadvantaged group also has larger \(d_p^2/a\) (median \(0.75\) versus \(0.20\)), so curvature is not isolated as the cause. Reweighting on the joint distribution of \(\kappa\) and \(d_p^2/a\) closes \(74\%\) of the gap, and \(\kappa\) accounts for \(29\)--\(36\%\) when it is added after the scale. In a separate five-seed audit with validation-selected, retrained models, a group-blind asymmetric penalty reduces the gap from 20.7 to 1.6 points at flat accuracy but raises the disadvantaged group's overshoot from \(0.009\) to \(0.080\). A group-conditional variant narrowed it further on none of the five seeds, which does not show the two variants to be equivalent (supplements S4 and S6). Two further datasets show no validity gap of 10 points or more, so the COMPAS gap is a single case and supports no general fairness claim.

\paragraph{Case study: deployment shift.} Validation-tuned inflation calibrated on a source slice can go stale when the rejected test points are split at the median of one feature or by protected group (supplement S4). On the COMPAS split by prior-offense count it loses 34 points of ray-hit validity on the target (\(100\!\to\!66\%\)), whose mass on \(\kappa\ge0\) is 35 points lower. Because the scale \((d_p,a)\) shifts as well, this match does not isolate curvature, and on F-MNIST, validity drops \(6.3\) points although the target's curvature marginal moves toward convexity. At a common scale, supplement Theorem~\ref{S-thm:s1-shift} gives the loss up to a boundary layer. Asymmetric alpha-1 training stays above \(98\%\) on every split, usually at higher overshoot.

\section{Discussion}
\label{sec:discussion}

The audits support the on-path mechanism, in which the gap follows \(g\approx-\kappa d_p^2\rho/(2a)\), with its sign set by the path curvature and its size by \(d_p^2\). No single rule in Table~\ref{tab:deploy} is best on validity, overshoot and query budget together, and inflation recovers validity by spending overshoot. Among rules that meet the target without per-user search, conformal-probe has the smallest overshoot, but it needs a calibration set. Ray line search is valid by construction and costs 189 forward evaluations per user in our implementation, and where affordable it is the right choice. Two attempts to improve the trade-off at training time failed (supplement S8).

\paragraph{Limitations.}\label{sec:limitations} The theory is local. Theorem~5.2 needs regime \((\star)\), which \(94\%\), \(100\%\) and \(68\%\) of the audited rejected COMPAS, German and Adult points pass under a floating-point enclosure of \(M\) conditional on library accuracy and on the stored gradient and curvature values (supplement S3.1). The criterion is not established for deep networks. In the ResNet-18 CelebA depth audit \citep{he2016deep,liu2015deep}, where max-pooling makes the score only piecewise smooth, a finite-grid screen leaves 5--12\% of points per attribute of the unregularized models unrejected (supplement S3.1); pooled over unregularized and asymmetric models, \(\mathbb P(d_{\text{ray}}\le d_p)\) and \(\mathbb P(\kappa\ge0)\) differ by 35.2 points, and 81.3\% of points lie in the band \(|\kappa|\le\hat Md_p/3\) computed with a lower bound \(\hat M\) on \(M\) (supplement S1). We do not attribute the CIFAR-10 gains of Section~\ref{sec:results} to the sign of \(\kappa\), which is untested there. The CIFAR-10 audit (smoothed activations) and the main comparison use one fixed, untuned \(\lambda\) per method and dataset, and the high-dimensional \(d_\star\) values rely on the C\&W solver. Tree ensembles have no gradient ray, and a Gaussian-smoothed surrogate's \(\mathbb P(\kappa\ge0)\) did not track a gradient-boosted ensemble's ray-hit validity (supplement S8), so tree ensembles remain open.

\section{Conclusion}
\label{sec:conclusion}

A user who follows the closed-form gradient step succeeds, to leading order, exactly when the deployed score bends upward along the recommended path. A rule that sees only the score value and gradient must trade failed recourse against wasted effort; one extra evaluation at the promised point reaches the best rate available to deterministic one-query rules that know the curvature bound and its Lipschitz constant, and split-conformal calibration of any base rule yields one that abstains or reaches the first crossing with marginal probability at least \(1-\delta\). Converses for randomized or noisy queries and paths constrained by actionability remain open.

\subsection*{AI use statement}

In this work, we used generative AI tools to generate and refactor research code and, in preparing this submission, to re-run experiments from the committed code; to assist in drafting proofs and, in revising the paper, to formulate and correct mathematical statements (hypotheses, constants and definitions), to supply short derivations, which we checked, and to draft the interpretation of results; and, as review panels during development, to give feedback on the research methodology and the experimental design. Some of the panels' suggestions were adopted, including the pass criteria of several prespecified audits and some additional checks. Tasks that involve human participants or qualitative data, such as formulating survey questions, transcribing recordings or thematic data analysis, are not applicable to this work. Additionally, we used generative AI tools to draft and edit text; as review panels on drafts of the manuscript and the code; and, in preparing this submission, for editing, for consistency and correctness checking, and for checking every bibliography entry against Crossref, OpenAlex, arXiv, DBLP, OpenReview, zbMATH, Open Library or publisher records. We have reviewed all AI-assisted work. Code generated or refactored with AI was verified by tests, by re-running the experiments, and by the repository's checks of the printed numbers against the committed results. We take responsibility for the final content of this work, including text, claims, or artifacts produced with the aid of generative AI.

\bibliography{references}

\section*{Checklist}

\begin{enumerate}

  \item For all models and algorithms presented, check if you include:
  \begin{enumerate}
    \item A clear description of the mathematical setting, assumptions, algorithm, and/or model. [Yes] Section~\ref{sec:formal} defines the setting and every distance and event; Sections~\ref{sec:frobenius}--\ref{sec:penalties} state each result with its assumptions and define the recourse rules and penalties; Section~\ref{sec:setup} specifies data and models.
    \item An analysis of the properties and complexity (time, space, sample size) of any algorithm. [Yes] Costs are counted in model queries: signed-quadratic recourse needs one Hessian-vector product, its probe variant one forward evaluation, and ray line search a per-user search of 189 forward evaluations in our implementation (Section~\ref{sec:sqr}); the penalties use one Hessian-vector product per example and step (Section~\ref{sec:penalties}). Section~\ref{sec:barrier} gives minimax rates per query budget and Section~\ref{sec:sqr} a necessary calibration sample size. The paper makes no wall-clock comparison; the README of the code archive lists measured CPU runtimes, and supplement S3 points to the stored per-batch HVP timings of the CIFAR-10 networks.
    \item (Optional) Anonymized source code, with specification of all dependencies, including external libraries. [Yes] The code archive in the supplementary material contains the anonymized code, a requirements file with pinned library versions, and setup instructions.
  \end{enumerate}

  \item For any theoretical claim, check if you include:
  \begin{enumerate}
    \item Statements of the full set of assumptions of all theoretical results. [Yes] Each result stated in Sections~\ref{sec:formal}--\ref{sec:penalties} includes its hypotheses; the results stated only in the supplement (S1 and S6), including Theorem~\ref{S-thm:s1-shift}, include theirs.
    \item Complete proofs of all theoretical results. [Yes] Supplement S1 proves every stated result, except the constrained-action proposition and corollary, which S6 states and proves.
    \item Clear explanations of any assumptions. [Yes] Each theorem is followed by a plain-language reading, and regime \((\star)\) and the query model are explained where they are introduced (Sections~\ref{sec:signed} and~\ref{sec:barrier}).
  \end{enumerate}

  \item For all figures and tables that present empirical results, check if you include:
  \begin{enumerate}
    \item The code, data, and instructions needed to reproduce the main experimental results (either in the supplemental material or as a URL). [Yes] The code archive in the supplementary material contains the code, the prepared COMPAS, German and Adult data, a script that rebuilds the Fashion-MNIST arrays from the official files, the result files behind every table, the commands, and a check that recomputes the headline numbers from those files. The CelebA depth audit (supplement S3.1) ships its scripts and per-point result files; a rerun can agree with them only in distribution, because the local CelebA copy used for them lacked a small fraction of the aligned images. The historical ResNet-18/CelebA scale suite recorded in supplement S4 cannot be regenerated from the retained code: its producer imported a variant of the curvature helper that was not retained, and its checkpoints and point-level ledgers were not retained either. It is not used as evidence for any claim (supplement S4).
    \item All the training details (e.g., data splits, hyperparameters, how they were chosen). [Yes] Section~\ref{sec:setup} and supplement S3.2 give splits, architectures, optimizers, epochs, penalty weights and seeds. The main comparison uses one fixed, untuned penalty weight per method and dataset (Section~\ref{sec:limitations}); the resplit audit selects hyperparameters on validation data only (supplement S4).
    \item A clear definition of the specific measure or statistics and error bars (e.g., with respect to the random seed after running experiments multiple times). [Yes] Section~\ref{sec:setup} defines the validity, undershoot and overshoot measures; Table~\ref{tab:main} and the supplement tables with \(\pm\) entries report seed means with population or sample standard deviations, as each caption or supplement S3.2 states, Table~\ref{tab:deploy} and the other tables report no seed dispersion, and the correlation behind the signed-curvature criterion has a cluster-bootstrap interval (Section~\ref{sec:results}).
    \item A description of the computing infrastructure used. (e.g., type of GPUs, internal cluster, or cloud provider). [Yes] The tabular and Fashion-MNIST experiments need only a CPU; the checks and CPU reruns listed in the README of the code archive were run on one Apple M1 CPU (8 cores, 8 GB RAM). The CIFAR-10 networks and the CelebA networks (the depth audit and the historical scale suite) were trained on CUDA GPUs, the CIFAR-10 seeds on two GPU types (supplement S4); the CIFAR-10 result files record the device only as CUDA, not the GPU model, and the CelebA result files do not record it.
  \end{enumerate}

  \item If you are using existing assets (e.g., code, data, models) or curating/releasing new assets, check if you include:
  \begin{enumerate}
    \item Citations of the creator If your work uses existing assets. [Yes] COMPAS, German Credit, Adult, and Fashion-MNIST are cited in Section~\ref{sec:setup} and CIFAR-10 in Section~\ref{sec:results}; CelebA and the ResNet-18 architecture, which the depth audit of the regime screen uses, are cited in the Limitations paragraph of Section~\ref{sec:limitations} and in supplement S3.1; ACS\-Income and the Taiwan credit-default data are cited in supplement S4; PyTorch and scikit-learn are cited in supplement S3.2.
    \item The license information of the assets, if applicable. [Yes] Supplement S11 states the license or terms of use of each dataset, or that its source states none.
    \item New assets either in the supplemental material or as a URL, if applicable. [Yes] The code and result files are in the code archive of the supplementary material.
    \item Information about consent from data providers/curators. [Not Applicable] We use only publicly released benchmark datasets and collected no data.
    \item Discussion of sensible content if applicable, e.g., personally identifiable information or offensive content. [Yes] Supplement S11 discusses the contested COMPAS and CelebA labels and states that the released COMPAS copy excludes names, dates of birth and case numbers.
  \end{enumerate}

  \item If you used crowdsourcing or conducted research with human subjects, check if you include:
  \begin{enumerate}
    \item The full text of instructions given to participants and screenshots. [Not Applicable] No crowdsourcing or human-subject research.
    \item Descriptions of potential participant risks, with links to Institutional Review Board (IRB) approvals if applicable. [Not Applicable] No human subjects.
    \item The estimated hourly wage paid to participants and the total amount spent on participant compensation. [Not Applicable] No participants.
  \end{enumerate}

\end{enumerate}

\StartSupplement

{\centering
{\LARGE\bfseries Supplementary Material\par}
\vspace{0.6em}
{\Large\bfseries The Signed Geometry of One-Shot Recourse:\\[2pt] On-Path Validity and the Signed-Curvature Criterion\par}
\vspace{1.8em}}

\section*{Overview}

This supplement contains the proofs, the results stated only here (among them Corollary~\ref{S-cor:s1-lattice} and Theorems~\ref{S-thm:s1-batch} and~\ref{S-thm:s1-shift}, which the main paper summarizes), further experiments with their full tables, and a reproducibility section. S1 gives the proofs and derivations, S2 the diagnostics of the distance estimators, S3 the experimental setup, S4 to S7 the full empirical results, S8 the negative results and S9 an additional figure. S10 maps every result to the code and result files behind it. It also explains what we mean when we say that pass criteria were fixed before a run, and why reviewers cannot verify that timing, which is recorded only in the history of a private repository. S11 covers data and ethics. Results that also appear in the main paper carry their main-paper numbers (for example, Theorem 5.2). Results stated only here are either numbered in the sequence S1.1, S1.2, \dots, which S1 shares with its displayed equations, or unnumbered. Tables and figures are numbered S1, S2, \dots; a reference to a table of the main paper says so (for example, main-paper Table~1).

\section{S1. Proofs and Derivations}

\subsection{Notation}

Throughout, the deployed score \(f:\mathbb R^d\to\mathbb R\) is \(C^2\) (standing assumption). For rejected \(x\), let \(f(x)<0\), \(a=\|\nabla f(x)\|_2>0\), \(m=|f(x)|\), \(d_p=m/a\), \(\hat g=\nabla f(x)/\|\nabla f(x)\|_2\), and \(\phi(\alpha)=f(x+\alpha\hat g)\). The first-crossing distance is \(d_{\text{ray}}=\inf\{\alpha>0:\phi(\alpha)\ge0\}\), with \(\inf\emptyset=+\infty\). Because \(\phi\) is continuous and \(\phi(0)<0\), the infimum is attained when it is finite, \(d_{\text{ray}}>0\), and \(d_{\text{ray}}\le t\) holds exactly when \(\max_{0\le\alpha\le t}\phi(\alpha)\ge0\). Endpoint success at recommendation \(t\) is \(\{\phi(t)\ge0\}\); the ray-hit event is \(\{d_{\text{ray}}\le t\}\), and their difference has mass \(r_t\). First-crossing undershoot and overshoot use \(d_{\text{ray}}-d_p\); the overshoot of a recommendation \(t\) is \([t-d_{\text{ray}}]_+\), which is \(0\) when \(d_{\text{ray}}>t\), including \(d_{\text{ray}}=+\infty\). The global nearest-boundary distance \(d_\star\) is a diagnostic quantity only.

\paragraph{Profiles.} A ray profile is a function on \([0,\infty)\), and its derivatives at \(0\) are one-sided. \(C^{2,1}([0,\infty))\) is the set of twice continuously differentiable functions on \([0,\infty)\) whose second derivative is Lipschitz on \([0,\infty)\), and \(\operatorname{Lip}(\phi'')\) denotes the smallest Lipschitz constant of \(\phi''\). The ray profile of a \(C^2\) score is \(C^2\) on \([0,\infty)\). A hypothesis that \(\phi''\) is \(M\)-Lipschitz on a named interval refers to that interval; \(\operatorname{Lip}(\phi'')\le M\), or such a bound stated along the ray with no interval, refers to \([0,\infty)\).

\paragraph{Populations.} A \emph{population} of rejected points is a Borel probability measure \(\mathcal P\) on the open set \(\{x:f(x)<0,\ \nabla f(x)\ne0\}\) of the fixed score \(f\). Probabilities \(\mathbb P\) and expectations \(\mathbb E\) are taken under \(\mathcal P\), and source and target populations \(S\) and \(T\) are two such measures for the same score. The quantities \(m\), \(a\), \(d_p\), \(\hat g\) and \(\kappa\) are continuous functions of \(x\), and \(\phi_x(t)\) and \(\phi_x''(t)\) are continuous in \((x,t)\). Hence, for every Borel recommendation \(t(x)\ge0\), the endpoint event \(\{\phi_x(t(x))\ge0\}\) and the ray-hit event \(\{d_{\text{ray}}(x)\le t(x)\}=\{\max_{0\le s\le t(x)}\phi_x(s)\ge0\}\) are Borel sets. Their probabilities are the endpoint validity and the ray-hit validity of the rule. In Remark~\ref{S-prop:s1-probe-exact}, Proposition~\ref{S-prop:s1-drift-screen}, Corollary~\ref{S-cor:s1-ambiguity} and their proofs, ``valid'' without qualification means endpoint validity. The per-point constant \(M(x)\in[0,\infty]\) is the smallest Lipschitz constant of \(\phi_x''\) on \([0,d_p(x)]\), and \(M(x)=+\infty\) if \(\phi_x''\) is not Lipschitz there; such a point lies in the band \(|\kappa|\le Md_p/3\) used below. \(M(x)\) is a Borel function of \(x\), because it is the supremum of the countably many continuous functions \(|\phi_x''(ud_p)-\phi_x''(u'd_p)|/(|u-u'|d_p)\) with \(u\ne u'\) rational in \([0,1]\), and any Borel upper bound on it may replace it below. The curvature CDF of a population is \(F_{\mathcal P}(b)=\mathbb P(\kappa\le b)\); a result that needs it to be continuous says so.

\subsection{Proof of Proposition 2.1: Score Gauge and Action Map}

Let \(h:\mathbb R^k\to\mathbb R^d\) be \(C^2\), let \(f\) be \(C^2\) near \(h(z)\), \(F=f\circ h\), and fix \(z\) with
\(F(z)\ne0\), \(a=\|\nabla_zF(z)\|>0\), and \(v=\nabla_zF/a\). The second-order
chain rule along \(v\) gives
\[
\kappa_{\rm eff}
=v^\top\nabla_z^2F\,v
=(J_hv)^\top\nabla^2f(h(z))(J_hv)
 +\left\langle\nabla f(h(z)),D^2h(z)[v,v]\right\rangle .
\]
For \(h=\mathrm{id}\) this is \(\kappa=\hat g^\top\nabla^2f\,\hat g\). This separates boundary bending in feature space from curvature of the feasible action map.

Now let \(\psi:\mathbb R\to\mathbb R\) be strictly increasing and \(C^2\), with
\(\psi(0)=0\) and \(\psi'(F(z))>0\). Strict monotonicity and the fixed zero preserve the
classifier and decision boundary. Moreover
\[
\nabla_z(\psi\circ F)=\psi'(F)\nabla_zF ,
\]
so the normalized action direction remains \(v\) and
\[
a_\psi=\psi'(F)a,\qquad
d_{p,\psi}=\frac{|\psi(F)|}{\psi'(F)a}.
\]
The assumption \(\psi'(F)>0\) is necessary: strict monotonicity alone permits an isolated zero
derivative, at which the normalized gradient ray is undefined. A second differentiation yields
\[
\nabla_z^2(\psi\circ F)
=\psi'(F)\nabla_z^2F+\psi''(F)\nabla_zF\nabla_zF^\top ,
\]
and therefore
\[
\kappa_\psi
=v^\top\nabla_z^2(\psi\circ F)v
=\psi'(F)\kappa_{\rm eff}+\psi''(F)a^2 .
\]

For the uniqueness statement, suppose that for every nonzero score \(s\) we have \(\psi'(s)>0\) (so that \(d_{p,\psi}\) is defined) and \(d_{p,\psi}=d_p\) for
every \(a>0\). Because \(\psi\) preserves the sign of \(s\),
\[
\frac{|\psi(s)|}{\psi'(s)}=|s|
\quad\Longleftrightarrow\quad
\psi(s)=s\psi'(s).
\]
On each open half-line the solutions are \(\psi(s)=c_-s\) and \(\psi(s)=c_+s\).
Differentiability at zero forces \(c_-=c_+=c\), and strict monotonicity forces \(c>0\).
Thus only a positive affine score map fixing zero preserves the promised length globally.

For a convex gauge, \(\psi''\ge0\), so
\[
\frac{\kappa_\psi}{a_\psi}=\frac{\kappa_{\rm eff}}{a}+\frac{\psi''(F)}{\psi'(F)}\,a\ \ge\ \frac{\kappa_{\rm eff}}{a}.
\]
The curvature relative to the slope therefore rises, and \(\kappa_\psi\ge0\) whenever \(\kappa_{\rm eff}\ge0\); a convex gauge can make a negative curvature nonnegative but never the reverse. The curvature itself need not rise, since the formula gives only \(\kappa_\psi\ge\psi'(F)\kappa_{\rm eff}\); for example, \(\psi(s)=e^s-1\) at \(F=-5\), \(a=1\) and \(\kappa_{\rm eff}=10\) gives \(\kappa_\psi=11e^{-5}<\kappa_{\rm eff}=10\). A convex \(\psi\) also lies above its tangent at \(F\), so \(0=\psi(0)\ge\psi(F)-\psi'(F)F\); at a rejected point (\(F<0\)) this gives \(|\psi(F)|\ge\psi'(F)|F|\), hence \(d_{p,\psi}\ge d_p\). A convex gauge thus raises the curvature relative to the slope at every point and lengthens the promised step at every rejected point, while the ray and \(d_{\text{ray}}\) stay fixed; this is the scale effect that main-paper Section~8 distinguishes from a change in the geometry of the path.

Finally, for any \(\tau>0\) define the globally increasing family
\[
\psi_c(s)=
\begin{cases}
\dfrac{\tau}{c}\bigl(e^{cs/\tau}-1\bigr),&c\ne0,\\[4pt]
s,&c=0.
\end{cases}
\]
It satisfies \(\psi_c'(s)=e^{cs/\tau}>0\), \(\psi_c(0)=0\), and
\(\psi_c''(s)=(c/\tau)e^{cs/\tau}\). For a fixed nonzero score, put
\(u=cs/\tau\). Then
\[
\frac{d_{p,\psi_c}}{d_p}=\frac{1-e^{-u}}{u},
\]
with its continuous value \(1\) at \(u=0\). As \(c\) ranges over \(\mathbb R\), this ratio ranges
continuously over all positive promised lengths. Also
\[
\kappa_{\psi_c}=e^{u}\left(\kappa_{\rm eff}+\frac{c}{\tau}a^2\right),
\]
so the same unchanged classifier and normalized action ray can be assigned either curvature sign.
This proves every assertion and establishes that \(d_p\) and \(\kappa\) belong to the deployed
score representation, not to the decision boundary alone. \hfill\(\square\)

\paragraph{Score-gauge audit.}
We applied the family \(\psi_c\) to trained models on COMPAS, German and Adult, with all four
training arms, five seeds, every rejected test point and \(c\in\{-1,-0.5,0,0.5,1\}\); \(\tau\) is
fit on training logits only, and the pass criteria were fixed before the run. On every dataset,
alpha-1 endpoint validity ranges from zero to one across the gauge grid, so the gauge changes the
outcome materially. No method ordering and no protected-group direction reverses, so this finding
rests on the swing in endpoint validity alone. A separate row-by-row check of the identities of
Proposition~2.1 requires every row to be finite and fails. The finite rows match the
promised-length and curvature formulas within \(1.3\times10^{-11}\) and the normalized ray within
\(1.3\times10^{-15}\), with no label, endpoint-sign or bracket mismatch, but 83 Adult rows at
\(c=1\) give non-finite transformed endpoint scores once the gauge assigns an extreme promised
length. We attribute these rows to floating-point overflow, not to a failure of the chain rule, and
we keep the criterion as it was stated. The experiment supports only the limitation that \(d_p\)
and \(\kappa\) depend on the score representation; it does not show that the criterion is robust to
the gauge or that a learned calibrator would help.

\subsection{Proof of Theorem 4.1}

\noindent\textit{Outline.} As the linear term cancels and \(\lVert H\rVert_2\le L\) (which holds in particular when \(\lVert H\rVert_F\le L\)), Taylor's theorem gives \(|f(y)|\le Ld_p^2/2\). The flow moves \(f\) to zero at unit rate with speed at most \(1/\gamma\), so \(d_\star\le d_p+|f(y)|/\gamma\le d_p+Ld_p^2/(2\gamma)\). Conversely, Taylor's theorem about \(x\), evaluated at a closest boundary point, with Cauchy--Schwarz gives \(a(d_p-d_\star)\le Ld_\star^2/2\), so \(d_p-d_\star\le Ld_p^2/(2a)\le Ld_p^2/(2\gamma)\) if \(d_\star<d_p\).

Let \(m=|f(x)|\), \(a=\|\nabla f(x)\|\), \(d_p=m/a\), \(\hat g=\nabla f(x)/a\), and \(y=x+d_p\hat g\). Throughout, \(\mathcal N\) denotes the recourse neighborhood of the theorem: a ball about \(x\) on which \(f\) is \(C^2\) with Hessian \(H\), \(\|H\|_2\le L\), and \(\|\nabla f\|\ge\gamma>0\), and which contains the one-shot point \(y\) and a closest boundary point. Being a ball about \(x\), it contains the segments from \(x\) to both points, which is all the two Taylor expansions below use. As in the theorem, we assume that the normalized gradient flow below, started at \(\eta(0)=y\), has a solution on \([0,|f(y)|]\) that stays in \(\mathcal N\). Taylor expansion along the one-shot step gives
\[
\begin{aligned}
f(y)
  &= f(x)+\nabla f(x)^\top(d_p\hat g)+R_y
  = R_y,\\
|R_y|&\le \frac{L}{2}d_p^2,
\end{aligned}
\]
because \(\|H\|_2\le L\) throughout the neighborhood; only this operator-norm bound is used, and \(\|H\|_F\le L\) implies it. To turn this residual into a distance bound, follow the normalized gradient-flow path
\[
\frac{d\eta}{dt}
  = -\operatorname{sign}(f(y))\frac{\nabla f(\eta)}{\|\nabla f(\eta)\|^2}.
\]
Along this path, \(d|f(\eta(t))|/dt=-1\) until the boundary is reached, and the spatial path speed is at most \(1/\gamma\). By assumption the solution stays in \(\mathcal N\), where \(\|\nabla f\|\ge\gamma\), on \([0,|f(y)|]\); hence \(f(\eta(|f(y)|))=0\), and the path from \(y\) to this boundary point has length at most \(|f(y)|/\gamma\). Therefore a boundary point lies within \(|f(y)|/\gamma\le Ld_p^2/(2\gamma)\) of \(y\), and
\[
d_\star\le d_p+\frac{L d_p^2}{2\gamma}.
\]

For the lower bound, let \(z\) be a closest boundary point and set \(r=\|z-x\|=d_\star\). Taylor expansion at \(x\) and Cauchy--Schwarz give
\[
0=f(z)\le -m+a r+\frac{L}{2}r^2.
\]
Since \(m=a d_p\), this implies
\[
a(d_p-r)\le \frac{L}{2}r^2.
\]
If \(r\ge d_p\), this already gives \(d_p-r\le0\). If \(r<d_p\), then \(r^2\le d_p^2\), hence
\[
d_p-r\le \frac{L}{2a}d_p^2\le \frac{L}{2\gamma}d_p^2.
\]
Together with the upper bound, this yields
\[
|d_\star-d_p|
  \le\frac{L}{2\gamma}d_p^2.
\]
This result concerns \(d_\star\), the global nearest-boundary diagnostic. It does not imply ray validity, undershoot, or overshoot for \(d_{\text{ray}}\).

\subsection{Proof of Theorem 5.2}

\noindent\textit{Outline.} With \(q\) as in Lemma 5.1, Taylor's theorem with integral remainder and the Lipschitz bound on \(\phi''\) give \(|\phi(\alpha)-q(\alpha)|\le M\alpha^3/6\), and \((\star)\) gives \(\phi(2d_p)\ge0\), so the first crossing lies in \([0,2d_p]\). There the remainder is at most \(4Md_p^3/3\) while \(q'\ge a-2|\kappa|d_p\ge a/2\), so the mean-value theorem turns the residual \(|q(d_{\text{ray}})|\) into the slab \(8Md_p^3/(3a)\) around the root \(d_{\text{ray}}^q=2d_p/(1+\sqrt{1+s})\) of \(q\), which gives the displayed leading term. The refined slab repeats the argument after bootstrapping the interval length (below).

Work on the ray segment \(S=\{x+\alpha\hat g:\alpha\in[0,2d_p]\}\), and assume directly that the directional-curvature profile \(\phi''\) is \(M\)-Lipschitz there; an \(M\)-Lipschitz Hessian is a stronger sufficient condition. Under \((\star)\) we show below that the first on-path crossing exists and lies in \([0,2d_p]\), so \(d_{\text{ray}}\) is well defined as that first root. Let \(q(\alpha)=-m+a\alpha+\tfrac\kappa2\alpha^2\). By Taylor expansion with integral remainder and \(|\phi''(t)-\kappa|\le Mt\) on \(S\),
\[
\begin{aligned}
\phi(\alpha)
  &= q(\alpha)+\int_0^\alpha(\alpha-t)(\phi''(t)-\kappa)\,dt,\\
|\phi(\alpha)-q(\alpha)|
  &\le M\alpha^3/6.
\end{aligned}
\]
Thus \(\overline\phi=q+M\alpha^3/6\) and \(\underline\phi=q-M\alpha^3/6\) sandwich \(\phi\), and their first roots sandwich \(d_{\text{ray}}\).

Under \((\star)\), \(\phi(2d_p)\ge m-2(|\kappa|+2Md_p)d_p^2\ge0\), so \(d_{\text{ray}}\le2d_p\). Solving \(q=0\) and rationalizing gives its smallest positive root
\[
d_{\text{ray}}^q=\frac{2d_p}{1+\sqrt{1+s}},\qquad
s=\frac{2\kappa d_p}{a},
\]
hence
\[
d_{\text{ray}}^q-d_p=-\frac{\kappa d_p^2}{2a}\rho(s),
\qquad
\rho(s)=\frac{4}{(1+\sqrt{1+s})^2}.
\]
On \([0,2d_p]\), \(|\phi-q|\le4Md_p^3/3\). Since \(\phi(d_{\text{ray}})=0\), \(|q(d_{\text{ray}})|\le4Md_p^3/3\). By the mean value theorem with \(q'(\xi)=a+\kappa\xi\ge a-2|\kappa|d_p\ge a/2\),
\[
|d_{\text{ray}}-d_{\text{ray}}^q|\le \frac{8Md_p^3}{3a}.
\]
The curvature-aware refinement \(\mathrm{slab}_1\) follows from \(|q(d_{\text{ray}})|\le M U^3/6\) and \(q'(\xi)\ge a-|\kappa|U\) for \(0\le\xi\le U\), where \(U\ge\max(d_{\text{ray}},d_{\text{ray}}^q)\). The first slab bootstraps \(U\) from \(U_0=2d_p\) to
\[
\mathrm{slab}_1=\frac{MU_1^3}{6(a-|\kappa|U_1)},\quad
U_1=\min\!\Big(d_{\text{ray}}^q+\tfrac{4M}{3}\tfrac{d_p^3}{a-2|\kappa|d_p},\,2d_p\Big),
\]
and \(|d_{\text{ray}}-d_{\text{ray}}^q|\le\mathrm{slab}_1\). With \(d_p\) and \(a\) fixed, \(\mathrm{slab}_1/(Md_p^3/(6a))=(U_1/d_p)^3\,a/(a-|\kappa|U_1)\) tends to \(1\) as \((\kappa,M)\to(0,0)\) along any path with \(M>0\), since \(d_{\text{ray}}^q\to d_p\) and hence \(U_1\to d_p\). The leading constant \(1/6\) is attained in this limit: for \(\kappa=0\) and \(\phi(t)=-m+at\pm Mt^3/6\) (so \(\phi''=\pm Mt\)), \(d_{\text{ray}}^q=d_p\) and \(\delta=d_{\text{ray}}-d_p\) solves \(a\delta=\mp M(d_p+\delta)^3/6\), so \(|d_{\text{ray}}-d_{\text{ray}}^q|/(Md_p^3/(6a))=(1+\delta/d_p)^3\to1\) as \(M\to0\).

\medskip
\noindent\textbf{Lemma \suppnum{S-lem:s1-no-recross} (no re-crossing in regime \((\star)\)).} \textit{Under the assumptions
of Theorem 5.2, \(\phi\) is strictly increasing on \([0,2d_p]\); hence its first crossing is its only
crossing there, and \(d_{\text{ray}}\le d_p \iff f(x+d_p\hat g)\ge0\): first-crossing validity and
endpoint success coincide.}

\smallskip
\noindent\textit{Proof.} By \(|\phi''(t)-\kappa|\le Mt\) and \(\phi'(0)=a\), for \(\alpha\in[0,2d_p]\):
\(\phi'(\alpha)\ge a-|\kappa|\alpha-\tfrac{M}{2}\alpha^2\ge a-2|\kappa|d_p-2Md_p^2\ge
a-\tfrac{a}{2}-\tfrac{a}{4}=\tfrac{a}{4}>0\), using \((\star)\) twice. So \(\phi\) is strictly
increasing on \([0,2d_p]\), and since \(\phi(0)=-m<0\le\phi(2d_p)\) (shown above), \(d_{\text{ray}}\)
is its unique zero there. If \(\phi(d_p)\ge0\) the intermediate value theorem gives
\(d_{\text{ray}}\le d_p\); conversely \(d_{\text{ray}}\le d_p\) and monotonicity give
\(\phi(d_p)\ge\phi(d_{\text{ray}})=0\). The constant is tight: \(\phi''=\kappa-M\alpha\) with
\(4|\kappa|d_p=a=8Md_p^2\) attains \(\phi'(2d_p)=a/4\). \hfill\(\square\)

\smallskip
\smallskip
\noindent\textit{Remark (longer segments).} For \(c\ge1\), assume that \(\phi''\) is
\(M\)-Lipschitz on \([0,c\,d_p]\) and that \((\star_c)\) holds:
\(2c|\kappa|d_p\le a\) and \(2c^2Md_p^2\le a\). The same computation gives
\(\phi'(\alpha)\ge a-c|\kappa|d_p-\tfrac{M}{2}c^2d_p^2\ge a/4>0\) on \([0,c\,d_p]\), so
first-crossing and endpoint success coincide for \emph{any} recommended distance
\(d_{\text{rec}}\le c\,d_p\); \(c=2\) recovers \((\star)\). Under \((\star)\) itself, with
\(\phi''\) \(M\)-Lipschitz on \([0,c\,d_p]\), the worst-case lower bound
\(a(1-c/4-c^2/16)\) on \(\phi'\) there remains positive for
\(c<2\sqrt5-2\approx2.47\) and reaches zero at \(c=2\sqrt5-2\).

\smallskip
\noindent Consequently, in regime \((\star)\) any rule whose recommended distance satisfies
\(d_{\text{rec}}\le2d_p\) certifies \emph{endpoint} success \(f(x+d_{\text{rec}}\hat g)\ge0\)
whenever it certifies the crossing: this covers alpha-1 (\(d_{\text{rec}}=d_p\)) and the
zero-search repair (\(\alpha\in[1,2]\)). The finite conformal distance
\(d_{\text{conf}}=\max\{0,d_{\mathrm{quad}}+\hat q_{1-\delta}\}\) is data-dependent, and no theorem
bounds it by \(2d_p\). Proposition 7.1 certifies crossing coverage-or-abstention unconditionally,
and the certificate extends to the endpoint on the subpopulation with \(d_{\text{conf}}\le c\,d_p\)
at points where \(\phi''\) is \(M\)-Lipschitz on \([0,c\,d_p]\) and \((\star_c)\) holds. The endpoint audit of S2 measures how often the length condition holds:
\(98.1\%\) of shallow-suite test points satisfy \(d_{\text{conf}}\le2d_p\), and the exceptions are
near-boundary points with tiny \(d_p\).
Endpoint success implies a crossing unconditionally, by the intermediate value theorem. The
converse, that a crossing before the recommendation implies endpoint success, requires that the
path not recross. Outside the regime the two events can differ in principle; the same audit
finds \(99.55\%\) raw agreement across the shallow suites and no true re-crossing.

\smallskip
\noindent\textit{Segment-safe fallback.} A theorem-safe endpoint variant is
\[
\widetilde d_{\mathrm{conf}}(x):=\max\{0,d_{\mathrm{conf}}(x)\},
\qquad
d_{\mathrm{seg}}(x)=
\begin{cases}
\widetilde d_{\mathrm{conf}}(x),&\widetilde d_{\mathrm{conf}}(x)\le2d_p(x),\\
d_p(x),&\widetilde d_{\mathrm{conf}}(x)>2d_p(x).
\end{cases}
\]
Thus \(0\le d_{\mathrm{seg}}\le2d_p\) by construction, and Lemma~\ref{S-lem:s1-no-recross} makes first-crossing and endpoint success equivalent for this rule at every point in regime \((\star)\). Clipping at zero can only increase the recommendation and therefore preserves the original coverage-or-abstention property before the fallback is applied. The second branch \emph{abstains from the conformal correction} and returns alpha-1, so it does not inherit the \(1-\delta\) coverage-or-abstention guarantee of Proposition 7.1. By the endpoint audit of S2, the pooled fallback rate is \(1.9\%\) (the complement of \(98.1\%\)); a script that recomputes it and checks the branch invariant is listed in S10. No result above bounds \(d_{\mathrm{conf}}/d_p\) for the unmodified rule by a data-independent constant \(c\): when \(\hat q_{1-\delta}>0\), every test point with \(d_p<\hat q_{1-\delta}/c\) has \(d_{\mathrm{conf}}\ge d_{\mathrm{quad}}+\hat q_{1-\delta}>c\,d_p\), because \(d_{\mathrm{quad}}\ge0\). A useful bound under an additional lower bound on \(d_p\), or with a scale-adaptive calibration margin, is open.

\subsection{Derivation of the Leading-Order Fixed-Alpha Observation}

At the promised distance, the linear term cancels; with \(M\) a Lipschitz constant of \(\phi''\) on \([0,d_p]\),
\[
f(x+d_p\hat g)=\frac{\kappa}{2}d_p^2+R,\qquad |R|\le \frac{M}{6}d_p^3.
\]
The alpha-1 recommendation is endpoint-valid exactly when this expression is nonnegative (by
Lemma~\ref{S-lem:s1-no-recross}, in regime \((\star)\) the endpoint event coincides with the
first-crossing event \(d_{\text{ray}}\le d_p\)). Hence
\[
\kappa\ge -\frac{2R}{d_p^2},\qquad
\left|\frac{2R}{d_p^2}\right|\le \frac{M d_p}{3}.
\]
Thus the alpha-1 endpoint event (equivalently, in regime \((\star)\), the crossing event) differs from \(\{\kappa\ge0\}\) only inside the pointwise boundary layer \(|\kappa|\le Md_p/3\). For a population (Notation) with per-point constant \(M=M(x)\): if \(Md_p\le\varepsilon\) at almost every point and the law of \(\kappa\) has a density at most \(C\) on \([-\varepsilon/3,\varepsilon/3]\), then Corollary~\ref{S-cor:s1-ambiguity} gives \(|\mathbb P(\phi(d_p)\ge0)-\mathbb P(\kappa\ge0)|\le\mathbb P(|\kappa|\le\varepsilon/3)\le2C\varepsilon/3\). Without such a uniform bound on \(Md_p\), the pointwise inclusion is the statement.

Now hold the one-shot rule fixed and consider points with a common scale \(h=d_p^2/a\). Under the hypotheses of Theorem 5.2,
\[
\Big|d_p-d_{\text{ray}}-\frac{\kappa h}{2}\rho(s)\Big|\le\frac{8Md_p^3}{3a},
\qquad s=\frac{2\kappa d_p}{a},
\]
so, up to this remainder and the factor \(\rho(s)\), the alpha-1 overshoot \([d_p-d_{\text{ray}}]_+\) is \(h[\kappa]_+/2\). We call \(\mathbb P(\kappa\ge0)\) and \(\mathbb E[h[\kappa]_+/2]\) the \emph{leading-order} alpha-1 validity and expected overshoot. Let \((\kappa_\theta)\) be a family of curvature laws that is stochastically nondecreasing in \(\theta\), that is, \(\mathbb P(\kappa_\theta>c)\) is nondecreasing in \(\theta\) for every real \(c\) (for example \(\kappa_\theta=\kappa_0+\theta\)). Both \(\mathbf 1\{\kappa\ge0\}\) and \([\kappa]_+\) are nondecreasing functions of \(\kappa\), so both leading-order quantities are nondecreasing in \(\theta\), the second wherever it is finite.

The observation is local, leading-order and specific to alpha-1. It holds the \(d_p^2/a\) scale fixed and assumes a monotone rightward shift of the curvature, and it says nothing about multiplicative or validation-tuned inflation, ray line search, changes of steepness, recourse rules that do not follow the gradient, model families that change \(d_p\) and \(a\) together with \(\kappa\), or arbitrary training objectives.

\subsection{Proof of Theorem 6.1 (First-Order Barrier)}

\noindent\textit{Outline.} \(\phi_\pm\) are the ray profiles of two scores that share the jet \((-m,a\hat g)\), so a first-order rule emits one number for both. Solving \(\phi_\pm=0\) gives crossings that straddle \(d_p\) and differ by at least \(Kd_p^2/a\) and by at most \(6(Kd_p/a)^3d_p\) more. Validity on \(\phi_-\) forces \(d_{\text{rec}}\ge d_{\text{ray}}(\phi_-)\), hence overshoot at least the gap on \(\phi_+\).

Fix \(x\) with \(f(x)=-m\), \(\nabla f(x)=a\hat g\), \(d_p=m/a\), \(K>0\), and \(Kd_p<a/2\). The scores \(f_\pm(x+v)=-m+a\langle\hat g,v\rangle\pm\tfrac K2\langle\hat g,v\rangle^2\) are polynomials, hence \(C^2\), have value \(-m\) and gradient \(a\hat g\) at \(x\), and have the ray profiles \(\phi_\pm(t)=-m+at\pm\tfrac K2 t^2\), with \(\phi_\pm''\equiv\pm K\). Hence \(|\phi_\pm''|\le K\) on \([0,2d_p]\) and \(f_\pm\in\mathcal F_K(x)\). Solving \(\phi_\pm=0\) for the first positive root,
\[
\begin{aligned}
d_{\text{ray}}(\phi_+)&=\frac aK\Big(\sqrt{1+\tfrac{2Kd_p}{a}}-1\Big)\quad(\text{overshoot, }<d_p),\\
d_{\text{ray}}(\phi_-)&=\frac aK\Big(1-\sqrt{1-\tfrac{2Kd_p}{a}}\Big)\quad(\text{undershoot, }>d_p),
\end{aligned}
\]
both real and in \((0,2d_p]\) since \(Kd_p<a/2\) and \(m=ad_p\). With \(u=2Kd_p/a\in(0,1)\), strict concavity gives \(\sqrt{1+u}<1+u/2\) and \(\sqrt{1-u}<1-u/2\), so \(d_{\text{ray}}(\phi_+)<d_p<d_{\text{ray}}(\phi_-)\), and the gap is \(d_{\text{ray}}(\phi_-)-d_{\text{ray}}(\phi_+)=\tfrac aK(2-\sqrt{1+u}-\sqrt{1-u})\). For the explicit bound, the binomial series \(\sqrt{1+w}=\sum_{j\ge0}\binom{1/2}{j}w^j\) converges absolutely on \([-1,1]\), and \(\binom{1/2}{2k}<0\) for every \(k\ge1\). Hence
\[
G(u):=2-\sqrt{1+u}-\sqrt{1-u}-\frac{u^2}{4}=\sum_{k\ge2}2\Big|\binom{1/2}{2k}\Big|\,u^{2k}
\]
(the \(k=1\) term is \(u^2/4\)) has nonnegative coefficients, so \(0\le G(u)\le G(1)\,u^4=(7/4-\sqrt2)\,u^4\) for \(u\in[0,1)\). Since \(\tfrac aK\cdot\tfrac{u^2}{4}=\tfrac{Kd_p^2}{a}\) and \(\tfrac aK\,u^4=16(Kd_p/a)^3d_p\),
\[
0\le d_{\text{ray}}(\phi_-)-d_{\text{ray}}(\phi_+)-\frac{Kd_p^2}{a}=\frac aK\,G(u)\le(28-16\sqrt2)\Big(\frac{Kd_p}{a}\Big)^3d_p<6\Big(\frac{Kd_p}{a}\Big)^3d_p .
\]
A first-order rule sees the identical jet for both scores and emits one \(d_{\text{rec}}\). If \(d_{\text{rec}}<d_{\text{ray}}(\phi_-)\), it is invalid on \(\phi_-\); otherwise \(d_{\text{rec}}\ge d_{\text{ray}}(\phi_-)=d_{\text{ray}}(\phi_+)+\text{gap}\), so the overshoot on \(\phi_+\) is at least the gap. \hfill\(\square\)

\subsection{Proof of Corollary 6.2 (One-Query Root-Estimation Upper Bound)}

\noindent\textbf{Query model.} A rule knows the first-order jet \((f(x),\nabla f(x))\) and may spend additional queries on the ray, either a forward evaluation \(z\mapsto f(z)\) or a Hessian-vector product \(v\mapsto H(x)v\). We distinguish these oracle types rather than assert an implementation-independent wall-clock ordering: in our implementation the probe is one additional forward pass, whereas the HVP invokes derivative passes whose exact cost depends on software and hardware.

\noindent\textbf{The probe.} Let \(\phi(t)=f(x+t\hat g)\), so \(\phi(0)=-m\), \(\phi'(0)=a\), \(\phi''(0)=\kappa\), and, with \(\phi''\) \(M\)-Lipschitz on \([0,d_p]\), \(\phi(t)=-m+at+\tfrac\kappa2 t^2+R(t)\) with \(|R(t)|\le\tfrac M6 t^3\) for \(t\in[0,d_p]\). Evaluate once at \(t=d_p\): since \(a d_p=m\), the linear terms cancel and
\[
y:=\phi(d_p)=\tfrac\kappa2 d_p^2+R(d_p),\qquad
\hat\kappa:=\frac{2y}{d_p^2}=\kappa+\frac{2R(d_p)}{d_p^2},
\]
so \(|\hat\kappa-\kappa|=2|R(d_p)|/d_p^2\le \tfrac13 Md_p\).

\noindent\textbf{Probe-quadratic error.} Write \(d_{\mathrm{quad}}(u)=2d_p/(1+\sqrt{1+2ud_p/a})\). Regime \((\star)\) gives \(|2\kappa d_p/a|\le\tfrac12\) and \(|2(\hat\kappa-\kappa)d_p/a|\le\tfrac1{12}\), so the segment between \(\kappa\) and \(\hat\kappa\) has \(1+2ud_p/a\ge5/12\). Therefore
\[
\left|\frac{\partial d_{\mathrm{quad}}(u)}{\partial u}\right|
\le C_0\frac{d_p^2}{a},\qquad
C_0:=\frac{2}{\sqrt{5/12}\,(1+\sqrt{5/12})^2}<1.15.
\]
Hence \(|d_{\mathrm{quad}}(\hat\kappa)-d_{\mathrm{quad}}(\kappa)|\le (C_0/3)Md_p^3/a\), and by the triangle inequality with Theorem 5.2's explicit slab,
\[
|d_{\mathrm{quad}}(\hat\kappa)-d_{\text{ray}}|
\le \left(\frac83+\frac{C_0}{3}\right)\frac{Md_p^3}{a}
=:\varepsilon_\star,
\]
the same order as the exact-\(\kappa\) (HVP) rule, at one forward evaluation. On the exact quadratic pair \(\phi_\pm\) of Theorem 6.1, \(R\equiv0\), so \(\hat\kappa=\pm K\) exactly and the probe distinguishes that hard pair with one query.

\noindent\textbf{Validity requires a margin.} The bound does not imply \(d_{\mathrm{quad}}(\hat\kappa)\ge d_{\text{ray}}\): the bare rule can undershoot within \(\varepsilon_\star\) (its measured alpha-1-scale validity is well below target, e.g.\ \(72\%\) five-seed mean on the unregularized COMPAS model). Under the stated regime, \(d_{\mathrm{quad}}(\hat\kappa)+\varepsilon_\star\) is therefore deterministically first-crossing-valid when \(M\) is known. Alternatively, split conformal calibrates the residual \(d_{\text{ray}}-d_{\mathrm{quad}}(\hat\kappa)\) for probabilistic first-crossing coverage-or-abstention without \(M\) or an HVP (Proposition 7.1). Numerical checks of the exact recovery on quadratic profiles, of the \(Md_p/3\) bound and of the equivalence on trained models are listed in S10. \hfill\(\square\)

\paragraph{Collocation sharpening.} The triangle bound above does not use that the probe quadratic \(q(t)=-m+at+\hat\kappa t^2/2\) matches \(\phi\) in value and derivative at \(0\) and in value at \(d_p\), whereas the exact-\(\kappa\) quadratic \(q_H(t)=-m+at+\kappa t^2/2\) matches only the second-order jet at \(0\). This cancellation needs only a Lipschitz \(\phi''\), not a continuous third derivative. With
\[
R(t):=\int_0^t(t-u)\{\phi''(u)-\kappa\}\,du,
\qquad
\phi(t)-q(t)=R(t)-\frac{t^2}{d_p^2}R(d_p),
\]
and \(\phi''\) \(M\)-Lipschitz on \([0,2d_p]\), the difference \(\phi-q\) vanishes at \(d_p\) and its derivative is at most \(8Md_p^2/3\) in absolute value on \([0,2d_p]\) (proof of Theorem 6.3 below), so \(|\phi(t)-q(t)|\le\tfrac83Md_p^2|t-d_p|\) there, whereas \(\phi-q_H=R\) is bounded only by \(Mt^3/6\). The uniform consequence is \eqref{S-eq:s1-probe-upper} below: for every profile in the class of Theorem 6.3 (so \(|\kappa|\le K_0\), \(K_0d_p/a\le1/8\), \(Md_p^2/a\le1/8\)),
\[
|d_{\mathrm{quad}}(\hat\kappa)-d_{\mathrm{ray}}|\le\frac{4M|\kappa|d_p^4}{a^2}+\frac{4M^2d_p^5}{3a^2},
\]
against the bound \(8Md_p^3/(3a)\) of Theorem 5.2 for the exact-\(\kappa\) quadratic, whose hypotheses this class satisfies; it is the bound quoted in Corollary 6.2 of the main text. The ratio of the two bounds is \(\tfrac32|\kappa|d_p/a+\tfrac12Md_p^2/a\). It compares upper bounds, not actual errors, since the exact-\(\kappa\) error may vanish for a particular profile; the worst-case rates for each oracle are Theorem 6.3 and Corollary~\ref{S-cor:s1-lattice} below. A symbolic computation on cubic profiles agrees with the cancellation, as do numerical checks on non-polynomial and kinked \(C^{2,1}\) profiles (S10). The measured conformal residuals are consistent with this mechanism, but the coverage-or-abstention guarantee of the conformal rules comes from calibration alone.

\subsection{The Drift Screen: a Per-Point Validity Certificate Without Regime \((\star)\)}
\label{S-sec:supp-drift-screen}

Theorem 5.2 and the signed-curvature criterion are \emph{leading-order} statements that assume regime
\((\star)\) (\(4|\kappa|d_p\le a\) and \(8Md_p^2\le a\)). The finite-grid screen of S3.1 does not
reject this regime for \(94\)/\(100\)/\(68\%\) of COMPAS/German/Adult points, but for only
\(5\)--\(12\%\) of unregularized points at depth. The observation below is a per-point statement
under strictly weaker hypotheses (no regime condition and no assumption that a crossing exists),
and it accounts quantitatively for \emph{where} the criterion degrades.

Write \(\phi(t)=f(x+t\hat g)\) for a rejected point, so \(\phi(0)=-m<0\),
\(\phi'(0)=a\), and \(d_p=m/a\). Because \(-m+ad_p=0\) \emph{exactly}, Taylor's theorem
with integral remainder collapses to
\begin{equation}\label{S-eq:probe-identity}
\phi(d_p)=\int_0^{d_p}(d_p-t)\,\phi''(t)\,dt,
\qquad\text{hence}\qquad
\hat\kappa:=\frac{2\phi(d_p)}{d_p^2}
=\frac{2}{d_p^2}\int_0^{d_p}(d_p-t)\,\phi''(t)\,dt ,
\end{equation}
so \(\hat\kappa\) is exactly the \((d_p-t)\)-weighted mean of the directional second derivative
over the traversed segment (the weight integrates to \(d_p^2/2\)).

\noindent\textbf{Remark \suppnum{S-prop:s1-probe-exact} (probe exactness).} \textit{The endpoint-validity indicator
satisfies \(\mathbf 1\{f(x+d_p\hat g)\ge0\}=\mathbf 1\{\hat\kappa\ge0\}\) identically, with no
hypotheses.} Immediate from \eqref{S-eq:probe-identity} and \(d_p^2>0\). The forward probe of
Corollary 6.2 therefore does not \emph{estimate} the governing scalar; it \emph{is} the alpha-1
endpoint functional, which \(\kappa\) reads only up to drift.

\noindent\textbf{Proposition \suppnum{S-prop:s1-drift-screen} (drift screen).} \textit{If the directional second derivative
\(\phi''\) is \(M\)-Lipschitz on \([0,d_p]\), then \(|\hat\kappa-\kappa|\le Md_p/3\); consequently, on the
\emph{unambiguous} set \(\{|\kappa|>Md_p/3\)\} the endpoint-validity indicator equals
\(\mathbf 1\{\kappa\ge0\}\) pointwise.}

\noindent\textit{Proof.} Subtracting \(\kappa=\phi''(0)\) inside \eqref{S-eq:probe-identity},
\(\hat\kappa-\kappa=\tfrac{2}{d_p^2}\int_0^{d_p}(d_p-t)(\phi''(t)-\phi''(0))\,dt\), and
\(|\phi''(t)-\phi''(0)|\le Mt\) gives
\(|\hat\kappa-\kappa|\le\tfrac{2M}{d_p^2}\int_0^{d_p}(d_p-t)t\,dt=\tfrac{2M}{d_p^2}\cdot\tfrac{d_p^3}{6}=\tfrac{Md_p}{3}\),
the constant already used in Corollary 6.2. If \(|\kappa|>Md_p/3\) then \(\hat\kappa\) cannot
reach the opposite side of \(0\), so \(\operatorname{sign}\hat\kappa=\operatorname{sign}\kappa\);
Remark~\ref{S-prop:s1-probe-exact} converts this to the validity indicator. \(\square\)

\noindent\textbf{Corollary \suppnum{S-cor:s1-ambiguity} (ambiguity envelope).} \textit{For any population of rejected
points, with the per-point constant \(M=M(x)\) of the Notation subsection, alpha-1 endpoint validity satisfies
\(|\mathbb P(\phi(d_p)\ge0)-\mathbb P(\kappa\ge0)|\le\mathbb P(|\kappa|\le Md_p/3)\),
the \emph{ambiguity mass}.} The two indicators are Borel functions of \(x\) and agree off the ambiguous set
(Proposition~\ref{S-prop:s1-drift-screen} at each point with \(M(x)<\infty\)), so their
expectations differ by at most its mass.

\noindent\textbf{Corollary \suppnum{S-cor:s1-ray-envelope} (ray-validity envelope).} \textit{Remark~\ref{S-prop:s1-probe-exact}, Proposition~\ref{S-prop:s1-drift-screen} and
Corollary~\ref{S-cor:s1-ambiguity} are statements about \emph{endpoint} validity \(\{\phi(d_p)\ge0\}\). For a
population of rejected points, with \(d_{\textnormal{ray}}=\inf\{t>0:\phi(t)\ge0\}\) (\(\inf\emptyset=+\infty\)) and \(M=M(x)\) as in the Notation subsection,
write \(r:=\mathbb P\bigl(d_{\textnormal{ray}}\le d_p,\ \phi(d_p)<0\bigr)\) for the
\emph{early-hit/endpoint-failure} mass. Then}
\[
\bigl|\mathbb P(d_{\textnormal{ray}}\le d_p)-\mathbb P(\kappa\ge0)-r\bigr|
\;\le\;\mathbb P\!\left(|\kappa|\le Md_p/3\right),
\]
\textit{and hence}
\(\bigl|\mathbb P(d_{\textnormal{ray}}\le d_p)-\mathbb P(\kappa\ge0)\bigr|\le
\mathbb P(|\kappa|\le Md_p/3)+r\).
\noindent\textit{Proof.} Write \(A=\{d_{\text{ray}}\le d_p\}\), \(B=\{\phi(d_p)\ge0\}\); both are Borel (Notation). Since
\(\phi\) is continuous with \(\phi(0)=-m<0\), endpoint success gives a \(t\in(0,d_p]\) with
\(\phi(t)\ge0\) by the intermediate value theorem (if \(\phi(d_p)=0\) then \(t=d_p\)), so
\(B\subseteq A\) and \(\mathbb P(A)=\mathbb P(B)+r\) with \(r=\mathbb P(A\setminus B)\) exactly.
Corollary~\ref{S-cor:s1-ambiguity} bounds \(|\mathbb P(B)-\mathbb P(\kappa\ge0)|\) by the ambiguity mass; substituting
\(\mathbb P(B)=\mathbb P(A)-r\) gives the centered form, and the triangle inequality gives the
absolute one. \(\square\)

\noindent The set \(A\setminus B\) contains both genuine sign re-crossings and isolated tangential
contacts (\(\phi(t_0)=0\) with \(\phi(d_p)<0\)). The \emph{centered} form is the informative
one: the discrepancy between ray-hit and endpoint validity is exactly \(r\), up to the ambiguity
mass. The absolute form is useful only when \(r\) is controlled separately. Outside regime
\((\star)\) nothing here bounds \(r\), and the right-hand side can exceed one.

\noindent The gap between endpoint and ray-hit validity is thus an \emph{explicit term}. It provably
vanishes under the assumptions of Theorem 5.2 in regime \((\star)\): Lemma~\ref{S-lem:s1-no-recross}
gives \(\phi'\ge a/4>0\) on \([0,2d_p]\), so \(\phi\) is strictly increasing and \(r=0\) for
every recommendation \(\le2d_p\). Outside that regime \(r\) can only be measured. We recomputed
\(\phi(d_p)\) and the crossing in float64 for the unregularized COMPAS, German and Adult models
of the drift-screen sample (seeds \(0,1\); at most \(400\) rejected points per seed) and scanned
\((0,d_p]\) on a uniform \(200\)-point grid. The scan \emph{detected} \(0/1787\)
early-hit/endpoint-failure cases. A finite scan gives a detection result and no upper bound, because
it would miss a positive excursion narrower than the grid or a tangential zero between grid points,
so this zero count is not certified. This sample also differs from, and
is smaller than, the \(24{,}189\) points of the endpoint audit in S2, which separately finds no
true re-crossing. The paper's own ray solver also gives \(0/1787\) for the strict event. Its
\emph{tolerant} validity flag (\(d_{\text{ray}}\le d_p+10^{-6}\)) additionally marks \(3\) Adult
points whose true first crossing lies \(2\times10^{-8}\)--\(5\times10^{-7}\) \emph{past} \(d_p\);
these are tolerance cases, not early hits. Corollary~\ref{S-cor:s1-ray-envelope} is stated for the
strict event. The endpoint identity itself is the algebra of \eqref{S-eq:probe-identity} and needs no
measurement.

\paragraph{What the screen adds.} With the screen, the signed-curvature criterion is more than an
in-regime leading-order approximation, because its error is \emph{bounded by a single measurable
statistic} that needs no smallness condition. Corollary~\ref{S-cor:s1-ambiguity} also predicts where
the criterion fails: the ambiguity mass grows with \(Md_p\), and \(Md_p\) is large at depth.

\paragraph{Measured ambiguity mass.} Replacing \(M\) by an \emph{upper} bound can only shrink the
unambiguous set, so the screen remains sound with the outward-rounded interval enclosures
\(M_{\rm cert}\) of S3.1. These exist for the tabular models only. At depth only the finite-grid
\emph{lower} bound \(\hat M\) is available, so the deep row of the table below is a diagnostic and
not a certificate.

\begin{center}\small
\begin{tabular}{lrrrrr}
\hline
Suite & Ambig.\ mass & Cert.-unambig. & Viol. & Flip\(\mid\)amb. & Flip\(\mid\)unamb. \\
\hline
COMPAS & \(4.0\%\) & \(96.0\%\) & \(1/768\) & \(46.9\%\) & \(0.13\%\) \\
German & \(0.0\%\) & \(100.0\%\) & \(0/187\) & --- & \(0.00\%\) \\
Adult & \(6.5\%\) & \(93.5\%\) & \(0/748\) & \(17.3\%\) & \(0.00\%\) \\
CelebA (deep, diag.) & \(81.3\%\) & --- & --- & \(44.5\%\) & \(5.7\%\) \\
\hline
\end{tabular}
\end{center}

\noindent On the three shallow suites the envelope of Corollary~\ref{S-cor:s1-ambiguity} is not
contradicted (measured \(|{\rm validity}-\mathbb P(\kappa\ge0)|\) against ambiguity mass:
\(1.25\le4.00\), \(0.00\le0.00\), \(1.12\le6.50\)). The deep row is a diagnostic only: its mass
uses the lower bound \(\hat M\), and outside regime \((\star)\) nothing bounds the term \(r\) by
which ray-hit and endpoint validity can differ (Corollary~\ref{S-cor:s1-ray-envelope}), so
\(35.2\le81.3\) does not test Corollary~\ref{S-cor:s1-ambiguity}. The one violation in the table (COMPAS,
\(1/768\)) comes from float32 evaluation and disappears in float64. The violating point has the
second-smallest \(d_p\) in the COMPAS suite (\(d_p=2.8\times10^{-4}\), \(\kappa=+0.126\)). It lies
about \(3\times10^{-4}\) from the boundary, where the \emph{float32} model cannot evaluate the
near-zero endpoint, and hence the first crossing, reliably; in float32 the crossing falls
\(9.1\times10^{-8}\) past \(d_p\). Recomputing \(f\), \(\nabla f\), \(d_p\) and the endpoint of the
same trained model entirely in \emph{float64} removes the artifact. The endpoint becomes
\(+4.8\times10^{-9}\) (valid, as \(\kappa>0\) predicts), \(\hat\kappa=\kappa\) to four figures, and
\emph{both} extreme near-boundary points are resolved, which leaves \textbf{0/768} violations on the
certified-unambiguous set under the float64 endpoint test. That test uses the exact endpoint object
of Remark~\ref{S-prop:s1-probe-exact} rather than the ray. The float32 effect is confined to the two
points about \(3\times10^{-4}\) from the boundary and does not change the suite-level
common-cohort conclusion.

\paragraph{Ambiguity mass at depth.} Ambiguity mass rises from \(0\)--\(6.5\%\) on the shallow
suites to \(81.3\%\) on ResNet-18/CelebA (pooled over the unregularized and Asym models of the
depth audit in S3.1). The reason is that \(\hat M\) (median \(138\) on the unregularized models)
lies two to four orders of magnitude above the tabular medians \(0.018\)--\(0.61\), while \(d_p\)
does not fall correspondingly. The mass also ranks where the criterion fails. At depth the
sign-disagreement rate is \(44.5\%\) on ambiguous points against \(5.7\%\) on unambiguous ones, and
on shallow COMPAS it is \(46.9\%\) against \(0.13\%\), so the same statistic separates agreement
from disagreement by two to three orders of magnitude on shallow data and by nearly one order of
magnitude at depth. This pattern is consistent with the shallow-regime mechanism persisting at
depth with a larger ambiguous set, rather than with a different mechanism. At depth, however, the
screen is only a \emph{diagnostic}. Only the lower bound \(\hat M\) is available there, so the deep
unambiguous set is optimistic, which is why a \(5.7\%\) residual remains on it, and the association
does not prove why the criterion degrades at depth.

\subsection{A Matching Deterministic One-Forward-Query Converse}

The preceding collocation rate is tight in the precise oracle used by Corollary 6.2. The
result below separates that oracle from one that reveals curvature before the query, and it
also separates symmetric root estimation from the validity--overshoot consequence.

\smallskip
\noindent\textbf{Theorem 6.3 (deterministic one-forward-query minimax rate).}
\textit{Fix \(a,d,K_0,M>0\) with \(K_0d/a\le1/8\) and \(Md^2/a\le1/8\), and let
\(\mathcal G(a,d,K_0,M)\) contain the profiles \(\phi\in C^{2,1}([0,\infty))\) satisfying}
\[
 \phi(0)=-ad,\qquad \phi'(0)=a,\qquad
 |\phi''(0)|\le K_0,\qquad \operatorname{Lip}(\phi'')\le M.
\]
\textit{A deterministic one-forward-query rule is given the first-order jet and the class
bounds, chooses a finite \(t_1\ge0\) before observing curvature, reads \(\phi(t_1)\), and outputs
a root estimate \(\widehat r\). Every profile in the class has exactly one root in
\([0,2d]\), which is its first positive root, and}
\begin{equation}
 \frac1{2592}\left(\frac{MK_0d^4}{a^2}+\frac{M^2d^5}{a^2}\right)
 \le \inf_{\mathcal A}\sup_{\phi\in\mathcal G}
 |\widehat r_{\mathcal A}(\phi)-d_{\rm ray}(\phi)|
 \le 4\left(\frac{MK_0d^4}{a^2}+\frac{M^2d^5}{a^2}\right).
 \label{S-eq:s1-onequery-minimax}
\end{equation}
\textit{If the output must instead be first-crossing-valid uniformly over \(\mathcal G\), its
worst-case overshoot is at least the same expression with constant \(1/1296\); the probe rule
with the explicit margin below has worst-case overshoot at most eight times the parenthesized
rate. Uniform first-crossing validity means \(\widehat r(\phi)\ge d_{\rm ray}(\phi)\)
for every \(\phi\), and overshoot means \(\widehat r(\phi)-d_{\rm ray}(\phi)\).
Thus both deterministic root estimation and uniformly valid recommendation have rate
\(\Theta(MK_0d^4/a^2+M^2d^5/a^2)\).}

\smallskip
\noindent\textit{Class geometry and the upper bound.}
Taylor's integral remainder gives, on \([0,2d]\),
\[
 \phi'(t)\ge a-K_0t-\tfrac12Mt^2\ge a/2.
\]
The signs at \(3d/4\) and \(5d/4\), using
\(|\phi(t)+ad-at-\phi''(0)t^2/2|\le Mt^3/6\), are respectively negative and positive.
Hence every profile has one first root \(r\in(3d/4,5d/4)\).

Write \(\kappa=\phi''(0)\),
\(R(t)=\phi(t)+ad-at-\kappa t^2/2\), and probe at \(d\):
\[
 \widehat\kappa=\frac{2\phi(d)}{d^2}=\kappa+\frac{2R(d)}{d^2},
 \qquad |\widehat\kappa-\kappa|\le Md/3.
\]
Let \(q_p(t)=-ad+at+\widehat\kappa t^2/2\). The bounds above imply
\(q_p'(t)\ge2a/3\) on \([0,2d]\), while \(q_p(0)<0\) and \(q_p(2d)\ge2ad/3\);
thus its first positive root \(r_p\) lies in \((0,2d)\). The
collocation error
\[
 e(t):=\phi(t)-q_p(t)=R(t)-\frac{t^2}{d^2}R(d)
\]
satisfies \(e(d)=0\) and \(|e'(t)|\le8Md^2/3\) on \([0,2d]\). Since
\[
 |r-d|\le \frac{|\kappa|d^2}{a}+\frac{Md^3}{3a},
 \qquad q_p'(t)\ge2a/3,
\]
the mean-value theorem yields the pointwise bound
\begin{equation}
 |r_p-r|\le \frac{4M|\kappa|d^4}{a^2}
             +\frac{4M^2d^5}{3a^2}
 \le 4\left(\frac{MK_0d^4}{a^2}+\frac{M^2d^5}{a^2}\right).
 \label{S-eq:s1-probe-upper}
\end{equation}
Moreover, with
\[
 E(\widehat\kappa)=\frac{4M|\widehat\kappa|d^4}{a^2}
                    +\frac{8M^2d^5}{3a^2},
\]
the curvature-estimation bound implies \(E(\widehat\kappa)\ge|r_p-r|\). Therefore
\(r_p+E(\widehat\kappa)\) is uniformly first-crossing-valid. Combining \eqref{S-eq:s1-probe-upper},
\(|\widehat\kappa|\le K_0+Md/3\), and the added margin gives the stated overshoot upper bound.

\smallskip
\noindent\textit{An arbitrary-query indistinguishable pair.}
Set \(\theta=2^{-1/3}\), \(c=1-\theta\), and, for the rule's chosen \(t_1\), define
\begin{equation}
 h_{t_1}(t)=\frac{M}{12}\left[(t_+)^3-2(t-ct_1)_+^3\right].
 \label{S-eq:s1-kink}
\end{equation}
This function has zero 2-jet at zero and
\(h_{t_1}(t_1)=Mt_1^3(1-2\theta^3)/12=0\). Also
\[
 h_{t_1}''(t)=\frac M2\{t_+-2(t-ct_1)_+\},
 \qquad \operatorname{Lip}(h_{t_1}'')=M/2.
\]
The kink is in the third derivative; \(h_{t_1}\) itself is \(C^2\). We will use the elementary
bound, valid for every \(r>0\),
\begin{equation}
 |h_{t_1}(r)|\ge\frac{M}{24}r^2\min\{|r-t_1|,r\}.
 \label{S-eq:s1-kink-lower}
\end{equation}
For completeness, put \(z=t_1/r\). If \(cz\ge1\), the second hinge in \eqref{S-eq:s1-kink} vanishes at
\(r\), so \eqref{S-eq:s1-kink-lower} is immediate. Otherwise
\(12h_{t_1}(r)/(Mr^3)=F(z):=1-2(1-cz)^3\). The function \(F\) is increasing and has its only
zero at \(z=1\). For \(z\in[0,2]\), putting \(u=1-cz\) and using
\(1-c=\theta\) gives the exact factorization
\[
 \frac{|F(z)|}{|z-1|}=2c(\theta^2+\theta u+u^2)
 \ge2c(7\theta^2-5\theta+1)>\frac12.
\]
On \([2,1/c]\), monotonicity gives \(F(z)\ge F(2)>1/2\). This proves \eqref{S-eq:s1-kink-lower}.

Consider the two center profiles
\[
 b_\sigma(t)=-ad+at+\sigma\left(\frac {K_0}2t^2+\frac M{12}(t_+)^3\right),
 \qquad \sigma\in\{-1,+1\}.
\]
Their roots \(\rho_+<d<\rho_-\) lie in \((3d/4,5d/4)\). At \(d\), their values are
opposites with magnitude \(K_0d^2/2+Md^3/12\); their derivatives are at most \(3a/2\).
Consequently
\begin{equation}
 D:=\rho_--\rho_+\ge\frac{2}{3a}\left(K_0d^2+\frac{Md^3}{6}\right).
 \label{S-eq:s1-center-gap}
\end{equation}
At least one center root, denoted \(\rho_\sigma\), obeys
\(|\rho_\sigma-t_1|\ge D/2\). Since both center roots lie in \((3d/4,5d/4)\), also
\(\rho_\sigma\ge D/2\). Now define the two profiles
\begin{equation}
 \phi_{\sigma,\pm}=b_\sigma\pm h_{t_1}.
 \label{S-eq:s1-witnesses}
\end{equation}
They share their full 2-jet at zero, including \(\kappa=\sigma K_0\), and they share the
queried value because \(h_{t_1}(t_1)=0\). The center consumes half of the Lipschitz budget and the
kink consumes the other half: the a.e. slopes of \(\phi_{\sigma,\pm}''\) belong to
\(\{-M,0,M\}\), so both profiles are in \(\mathcal G\). They are strictly increasing on
\([0,2d]\). At the center root their values are \(\pm h_{t_1}(\rho_\sigma)\), hence their roots
straddle \(\rho_\sigma\). Equations \eqref{S-eq:s1-kink-lower}--\eqref{S-eq:s1-center-gap} and the derivative upper bound give
\begin{equation}
\begin{aligned}
 |h_{t_1}(\rho_\sigma)|&\ge \frac{Md^2D}{192},\\
 |d_{\rm ray}(\phi_{\sigma,+})-d_{\rm ray}(\phi_{\sigma,-})|
 &\ge\frac{4|h_{t_1}(\rho_\sigma)|}{3a}\\
 &\ge \frac{MK_0d^4}{216a^2}+\frac{M^2d^5}{1296a^2}
 \ge\frac1{1296}\left(\frac{MK_0d^4}{a^2}+\frac{M^2d^5}{a^2}\right).
\end{aligned}
 \label{S-eq:s1-witness-separation}
\end{equation}
A deterministic estimator sees the same transcript on the pair, so one absolute error is at
least half their root gap. A rule that is first-crossing-valid on both must clear the later
root, and therefore overshoots the earlier one by the full gap. This proves the theorem.

\paragraph{Same segment-capped class.}
To compare directly with Theorem 6.1, let \(\mathcal H_K\) be the \(C^{2,1}\) profiles with
the shared first-order jet, \(\operatorname{Lip}(\phi'')\le M\), and
\(\sup_{0\le t\le2d}|\phi''(t)|\le K\), and assume \(Kd/a\le1/8\) and
\(Md\le K/4\). The probe upper bound \eqref{S-eq:s1-probe-upper} is at most
\((13/3)MKd^4/a^2\); its valid-margin overshoot is at most
\((28/3)MKd^4/a^2\). For the converse, instantiate \eqref{S-eq:s1-witnesses} with \(K_0=K/2\).
Because each profile of the pair has curvature Lipschitz constant at most \(M\), on \([0,2d]\)
\[
 |\phi_{\sigma,\pm}''(t)|\le K/2+2Md\le K,
\]
so the pair lies in \(\mathcal H_K\), and \eqref{S-eq:s1-witness-separation} gives estimator error at least
\(MKd^4/(864a^2)\) and valid overshoot at least \(MKd^4/(432a^2)\).
Thus deterministic one-query estimation and valid overshoot are both
\(\Theta(MKd^4/a^2)\) on this common segment-capped class. The quadratic pair of
Theorem 6.1 also lies in \(\mathcal H_K\) (its second derivatives are constant), so zero queries
retain the valid-overshoot lower bound \(Kd^2/a\) of Theorem 6.1.

\paragraph{Pointwise sharpness of the \(d_p\) probe.}
The at-origin cap \(K_0\) in Theorem 6.3 can be replaced by the actual nonzero curvature for
the paper's fixed probe. Fix \(\kappa\), take \(t_1=d\), let \(\tau=cd\), and choose
\(\sigma=\operatorname{sign}(\kappa)\) (take \(\sigma=1\) if \(\kappa=0\)). With
\(B(t)=-ad+at+\kappa t^2/2\), the profiles
\begin{equation}
 \phi_E(t)=B(t)+\frac{\sigma M}{6}\{(t_+)^3-(t-\tau)_+^3\},\qquad
 \phi_L(t)=B(t)+\frac{\sigma M}{6}(t-\tau)_+^3
 \label{S-eq:s1-fixed-probe-pair}
\end{equation}
share the 2-jet and \(\phi_E(d)=\phi_L(d)=\kappa d^2/2+\sigma Md^3/12\).
They are \(C^{2,1}(M)\). Put \(\gamma=2^{-2/3}-1/2\). Expanding their simple roots at
\(d=0\), or equivalently expanding around the root of their average profile, gives
\begin{equation}
 \lim_{d\to0}\frac{|d_{\rm ray}(\phi_E)-d_{\rm ray}(\phi_L)|}
 {M|\kappa|d^4/a^2}=\frac\gamma2\quad(\kappa\ne0),\qquad
 \lim_{d\to0}\frac{|d_{\rm ray}(\phi_E)-d_{\rm ray}(\phi_L)|}
 {M^2d^5/a^2}=\frac\gamma{12}\quad(\kappa=0).
 \label{S-eq:s1-fixed-probe-limits}
\end{equation}
Indeed, the profile difference vanishes at \(d\), its derivative there is
\(-\sigma\gamma Md^2\). For fixed \(\kappa\ne0\), the average root obeys
\(\rho-d=-\kappa d^2/(2a)+O(d^3)\); at \(\kappa=0\), it obeys
\(\rho-d=-\sigma Md^3/(12a)+O(d^5)\). Division by the limiting root slope \(a\) gives
\eqref{S-eq:s1-fixed-probe-limits}. Thus neither term in the collocation upper rate is an artifact of the proof.

A rate-sharp lattice fixes the exchange rate (Corollary~\ref{S-cor:s1-lattice} below). With \(W=K_0d_p^2/a+Md_p^3/a\) the zero-query rate, one forward value improves it by a factor of order \(Md_p^2/a\), from \(\Theta(W)\) to \(\Theta(Md_p^2W/a)\), under both losses, while revealing exact \(\kappa\) beforehand drops the \(K_0\) term; randomized and noisy converses stay open. At larger budgets, choosing each query after seeing the earlier values matters more than adding queries: \(q\ge3\) locations chosen in advance give minimax \(\Theta((M/a)(W/q)^3)\) (Theorem~\ref{S-thm:s1-batch} in the next subsection), whereas adaptively chosen locations make the error decay faster than any power of \(q\) for fixed \((a,d,K_0,M)\) (``The separation'' in the next subsection).

\medskip
\noindent\textbf{Corollary \suppnum{S-cor:s1-lattice} (rate-sharp curvature--collocation oracle lattice).}
Retain the class \(\mathcal G(a,d,K_0,M)\) and smallness conditions of Theorem 6.3, and put
\[
 W:=\frac{K_0d^2}{a}+\frac{Md^3}{a},\qquad
 U:=\frac{Md^3}{a},\qquad \lambda:=\frac{Md^2}{a}.
\]
A deterministic exact oracle either hides \(\kappa=\phi''(0)\), or supplies that scalar
before any forward-query location is selected. With zero or one exact forward value, the minimax
absolute first-root error, and separately the uniformly first-crossing-valid minimax overshoot,
have rates
\begin{equation}
\begin{array}{c|cc}
 &0\text{ forward values}&1\text{ forward value}\\ \hline
\kappa\text{ hidden}&\Theta(W)&\Theta(\lambda W)\\
\kappa\text{ revealed first}&\Theta(U)&\Theta(\lambda U).
\end{array}
\label{S-eq:s1-rate-table}
\end{equation}
More explicitly, estimation lies respectively in
\[
[W/9,W],\quad[\lambda W/2592,4\lambda W],\quad
[U/9,U/2],\quad[\lambda U/2592,13\lambda U/24],
\]
and uniformly valid overshoot lies in
\[
[2W/9,2W],\quad[\lambda W/1296,8\lambda W],\quad
[2U/9,U],\quad[\lambda U/1296,13\lambda U/12].
\]
The order is the same for the two losses, but the constants differ.

\smallskip
\noindent\textit{Proof: zero forward values.}
Twice integrating the curvature constraints gives the pointwise sharp hidden-curvature envelopes
\begin{equation}
 -ad+at-\frac{K_0t^2}{2}-\frac{Mt^3}{6}
 \ \le\ \phi(t)\ \le\
 -ad+at+\frac{K_0t^2}{2}+\frac{Mt^3}{6}.
 \label{S-eq:s1-hidden-envelopes}
\end{equation}
Both envelopes belong to \(\mathcal G\), are strictly increasing on \([0,2d]\), and attain the
latest and earliest compatible roots. In fact the profiles obtained by multiplying both signed
terms by \(s\in[-1,1]\) continuously fill that complete root interval. Thus its midpoint is the
exact minimax estimator under absolute loss; its later endpoint is the smallest uniformly valid
recommendation, with worst-case overshoot equal to the full interval width. At \(t=d\) the
envelopes have opposite values of magnitude
\(A=K_0d^2/2+Md^3/6\); their derivatives are at most \(3a/2\), so their root gap is at least
\(4A/(3a)\ge2W/9\). Conversely, every compatible root is within
\(2A/a\le W\) of \(d\). Hence \(d\) has error at most \(W\), while \(d+W\) is uniformly valid
with overshoot at most \(2W\).

If \(\kappa\) is revealed, replace \eqref{S-eq:s1-hidden-envelopes} by the exact-curvature envelopes
\begin{equation}
 q_\kappa(t)-\frac{Mt^3}{6}\le\phi(t)\le
 q_\kappa(t)+\frac{Mt^3}{6},\qquad
 q_\kappa(t)=-ad+at+\frac{\kappa t^2}{2}.
 \label{S-eq:s1-revealed-envelopes}
\end{equation}
They again attain the complete root interval. Let \(T\) be the first positive root of
\(q_\kappa\), equivalently its unique root in \((0,2d)\).
The regime gives \(T\le8d/7\), and \(\phi'\ge a/2\); hence every compatible root is within
\(M T^3/(3a)\le U/2\) of \(T\). The recommendation \(T+U/2\) is therefore uniformly valid with
overshoot at most \(U\). For the converse it suffices to fix the revealed value \(\kappa=0\):
the two profiles \(-ad+at\pm Mt^3/6\) have root gap at least \(2U/9\), giving the displayed
estimation and valid-overshoot lower bounds. These attainable envelopes also make the
\(\pm Md/3\) alpha-1 boundary layer derived above sharp: ambiguity is
\([-Md/3,Md/3)\), universal validity begins at \(+Md/3\), and universal invalidity holds for
\(\kappa<-Md/3\).

\smallskip
\noindent\textit{Proof: one forward value.}
The hidden-curvature cell is Theorem 6.3. Now suppose exact \(\kappa\) is revealed first. Put
\(T=d_{\rm quad}(\kappa)\), \(p=q_\kappa'(T)=a+\kappa T\), query \(y=\phi(T)\), and output
\(\widetilde r=T-y/p\). This uses no derivative query beyond the revealed scalar. Writing
\(\phi=q_\kappa+R\), let \(r=d_{\rm ray}\) and \(\delta=r-T\). The regime gives
\(T\le8d/7\), \(p\ge3a/4\), \(|\delta|\le U/2\), and \(r,T\le5d/4\). The exact root identity is
\begin{equation}
 \widetilde r-r=
 \frac{\kappa\delta^2/2+R(r)-R(T)}{p}.
 \label{S-eq:s1-root-identity}
\end{equation}
Since \(|R'(t)|\le Mt^2/2\), the right side is bounded by
\[
 \frac4{3a}\left\{\frac{|\kappa|U^2}{8}
       +\frac{25Md^2U}{64}\right\}
 \le\frac{13}{24}\lambda U.
\]
Adding that radius is uniformly valid and has overshoot at most \(13\lambda U/12\).

For the matching converse, fix the revealed curvature at \(\kappa=0\). A deterministic rule then
chooses some finite \(t_1\ge0\). Let
\[
 b_\sigma(t)=-ad+at+\sigma\frac{M}{12}(t_+)^3,
 \qquad \sigma\in\{-1,+1\},
\]
whose roots are separated by at least \(U/9\). Choose the center root farther from \(t_1\), and
use the same kink \(h_{t_1}\) from \eqref{S-eq:s1-kink}. The pair \(b_\sigma\pm h_{t_1}\) shares the exact 2-jet
(including the already revealed \(\kappa=0\)) and the queried value because \(h_{t_1}(t_1)=0\).
The center and kink each use half the Lipschitz budget; the final a.e. slopes of the second
derivatives lie in \(\{-M,0,M\}\). Equations \eqref{S-eq:s1-kink-lower}--\eqref{S-eq:s1-witness-separation}, now with \(K_0=0\), give root gap
at least \(\lambda U/1296\) for every finite \(t_1\), including \(t_1=0\). Indistinguishability
forces estimation error at least half that gap, while uniform validity forces the full gap as
overshoot. This proves \eqref{S-eq:s1-rate-table}. \hfill\(\square\)

The corollary compares deterministic, noiseless, fixed-ray information models. It reveals only the
directional scalar \(\kappa\) and no off-ray Hessian information; randomized and noisy converses
remain open, as does the exact adaptive multi-query rate (Theorem~\ref{S-thm:s1-batch} settles the nonadaptive
case). The four displayed cells are rate-sharp, not constant-sharp: the matching upper and lower
constants differ by up to four orders of magnitude. That one forward value can beat curvature
alone is an asymptotic statement: the ratio of the two displayed rates is \(K_0d/a+Md^2/a\), and
the coarse finite constants do not imply strict dominance everywhere in the allowed regime.
Numerical checks of both lower-bound constructions (exact transcripts, regularity, first roots,
finite margins and the separate parameter scalings) are listed in S10.

\subsection{Many Queries: an Adaptivity Separation}

Theorem 6.3 settles \(q=1\). For a budget of \(q\) forward values the picture splits by \emph{when}
the locations are chosen. Keep the class of Theorem 6.3 (\(\phi(0)=-ad\), \(\phi'(0)=a\),
\(|\phi''(0)|\le K_0\), \(\operatorname{Lip}(\phi'')\le M\), \(K_0d/a\le1/8\), \(Md^2/a\le1/8\)) and put
\[
W:=\frac{K_0d^2}{a}+\frac{Md^3}{a},\qquad
\mu:=\frac{K_0d}{a},\quad \lambda:=\frac{Md^2}{a},\quad v:=\mu+\tfrac{\lambda}{3}.
\]

\smallskip
\noindent\textbf{Theorem \suppnum{S-thm:s1-batch} (batch/nonadaptive \(q\)-query minimax rate).}
\textit{For deterministic rules whose \(q\ge3\) query locations are all fixed before any value is
observed, the minimax first-root error, and likewise the worst-case overshoot of a uniformly
first-crossing-valid rule, is}
\[
R_{\mathrm{batch}}(q)=\Theta\!\Big(\frac{M}{a}\Big(\frac{W}{q}\Big)^{3}\Big).
\]

\smallskip
\noindent\textit{Throughout the proof }\(\mu=K_0d/a\le1/8\), \(\lambda=Md^2/a\le1/8\),
\(v=\mu+\lambda/3\le1/6\), \textit{and} \(W=d(\mu+\lambda)\), \textit{so} \(dv\in[W/3,W]\).

\smallskip
\noindent\textbf{Lemma \suppnum{S-lem:s1-root-band} (root band).} \textit{Every \(\phi\) in the class satisfies
\(\phi'\ge a/2\) on \([0,2d]\) and has a unique root \(r\) there, which is its first positive root,
and \(|r-d|\le dv\).}

\noindent\textit{Proof.} \(|\phi''(t)|\le K_0+Mt\), so on \([0,2d]\),
\(\phi'(t)\ge a-K_0t-Mt^2/2\ge a(1-2\mu-2\lambda)\ge a/2\). Hence \(\phi\) is strictly increasing
there; \(\phi(0)=-ad<0\), and \(|\phi(d)|\le K_0d^2/2+Md^3/6=ad(\mu/2+\lambda/6)\), so
\(|r-d|\le|\phi(d)|/(a/2)=dv\le d/6\); in particular \(r\in[5d/6,7d/6]\subset(0,2d)\). \(\square\)

\smallskip
\noindent\textbf{Lemma \suppnum{S-lem:s1-interp} (interpolation).} \textit{Let \(\phi\in C^{2,1}([0,\infty))\) with
\(|\phi''(0)|\le K_0\) and \(\operatorname{Lip}(\phi'')\le M\), and let \(x_0,x_1,x_2,x_3\ge0\) be distinct. Then
\(|\phi[x_0,x_1,x_2,x_3]|\le M/6\). Consequently, if \(P\) is the polynomial of
degree \(\le2\) with \(P(x_j)=\phi(x_j)\) for \(j=0,1,2\), then for all \(t\ge0\)}
\[
|\phi(t)-P(t)|\le\frac M6\,|t-x_0|\,|t-x_1|\,|t-x_2|,
\qquad
|P''|=|\phi''(\xi)|\le K_0+M\max_j x_j
\]
\textit{for some \(\xi\) in the convex hull of \(x_0,x_1,x_2\).}

\noindent\textit{Proof.} Let \(\eta\ge0\) be a smooth function supported in \([0,1]\) with \(\int\eta=1\),
and for \(\epsilon>0\) and \(t\ge0\) put \(\phi_\epsilon(t)=\int_0^1\phi(t+\epsilon s)\eta(s)\,ds\). Then
\(\phi_\epsilon\) is smooth on \([0,\infty)\), \(\phi_\epsilon''(t)=\int_0^1\phi''(t+\epsilon s)\eta(s)\,ds\) is
\(M\)-Lipschitz, so \(|\phi_\epsilon'''|\le M\), and \(\phi_\epsilon\to\phi\) pointwise as \(\epsilon\to0\).
The mean-value theorem for divided differences gives \(\phi_\epsilon[x_0,x_1,x_2,x_3]=\phi_\epsilon'''(\zeta)/6\)
for some \(\zeta\), so \(|\phi_\epsilon[x_0,x_1,x_2,x_3]|\le M/6\). Because the nodes are distinct, the
divided difference is a fixed linear combination of the four values, and letting \(\epsilon\to0\) gives
the bound for \(\phi\). The Newton error identity \(\phi(t)-P(t)=\phi[x_0,x_1,x_2,t]\prod_j(t-x_j)\) for
\(t\notin\{x_0,x_1,x_2\}\) (both sides vanish at the nodes) gives the first display. The mean-value
theorem for the second divided difference, \(P''/2=\phi[x_0,x_1,x_2]=\phi''(\xi)/2\), together with
\(|\phi''(\xi)|\le|\phi''(0)|+M\xi\), gives the second. \(\square\)

\smallskip
\noindent\textit{Proof of Theorem~\ref{S-thm:s1-batch}, upper bound.}
Fix the enlarged band \(\widetilde I=[d(1-2v),d(1+2v)]\); since \(2v\le1/3\) we have
\(\widetilde I\subset[2d/3,4d/3]\subset[0,2d]\), and by Lemma~\ref{S-lem:s1-root-band} the true root \(r\) lies in the
concentric interval \([d(1-v),d(1+v)]\), at distance at least \(dv\) from \(\partial\widetilde I\).
Place the \(q\ge3\) locations \(x_1<\dots<x_q\) equispaced on \(\widetilde I\), spacing
\(h=4dv/(q-1)\); these depend only on \((a,d,K_0,M)\), so the design is nonadaptive. Observe
\(y_i=\phi(x_i)\). If some \(y_i=0\), output \(x_i\). Otherwise \(y_1\le0\le y_q\) and \(\phi\) is
increasing on \(\widetilde I\), so \(k:=\max\{i:y_i<0\}\) is well defined; set
\(i^\star=\min(\max(k-1,1),q-2)\) and let \(P\) interpolate at
\(x_{i^\star},x_{i^\star+1},x_{i^\star+2}\). In every case \(r\) lies in the convex hull of those
three nodes extended by one spacing, so \(|r-x_j|\le2h\) for the three of them, and by Lemma~\ref{S-lem:s1-interp}
\begin{equation}
|P(r)|=|P(r)-\phi(r)|\le\frac M6(2h)^3=\frac{4M}{3}h^3 .
\label{S-eq:s1-interp-residual}
\end{equation}
By Lemma~\ref{S-lem:s1-interp} and the actual interpolation-node band \(\widetilde I\subset[2d/3,4d/3]\),
\(|P''|\le K_0+M\cdot\tfrac43 d\); by the mean value theorem \(P'\) equals a secant
slope of \(\phi\), hence \(\ge a/2\), somewhere in \(\widetilde I\); as
\(\operatorname{diam}\widetilde I=4dv\),
\begin{equation}
P'\ \ge\ \frac a2-4dv\Big(K_0+\tfrac43Md\Big)
=a\Big(\tfrac12-4(\mu+\tfrac{\lambda}{3})(\mu+\tfrac43\lambda)\Big)
\ \ge\ \frac{11a}{36}\ >\ \frac{a}{4}\quad\text{on }\widetilde I .
\label{S-eq:s1-interp-slope}
\end{equation}
Hence \(P\) has at most one root in \(\widetilde I\), and \eqref{S-eq:s1-interp-residual}--\eqref{S-eq:s1-interp-slope} place a root \(\hat s\)
within \(4|P(r)|/a\le(16M/3a)h^3\) of \(r\); since \(h\le2dv\) this distance is at most
\((128/3)\lambda dv^3\le0.15\,dv\), which is below the margin \(dv\), so \(\hat s\) exists in
\(\widetilde I\) and is unique. Finally \(q\ge3\) gives \(q-1\ge2q/3\), so \(h\le6dv/q\le6W/q\)
(we use the cruder \(h\le12W/q\), valid since \(dv\le W\)), and
\begin{equation}
|\hat s-r|\ \le\ \frac{16M}{3a}h^3\ \le\ \frac{16}{3}\cdot12^3\,\frac Ma\Big(\frac Wq\Big)^3
=9216\,\frac Ma\Big(\frac Wq\Big)^3 .
\label{S-eq:s1-batch-upper}
\end{equation}
Outputting \(\hat s\) plus that deterministic radius is uniformly first-crossing-valid with
overshoot at most twice it.

\smallskip
\noindent\textit{Proof of Theorem~\ref{S-thm:s1-batch}, converse.}
For \(\sigma\in[-1,1]\) let \(b_\sigma(t)=-ad+at+\sigma(K_0t^2/4+Mt^3/24)\). Then
\(|b_\sigma''(0)|=|\sigma|K_0/2\le K_0\) and \(\operatorname{Lip}(b_\sigma'')\le M/4\), so each
\(b_\sigma\) spends at most a quarter of the Lipschitz budget. On \([0,2d]\),
\(b_\sigma'\in[a(1-\mu-\lambda/2),a(1+\mu+\lambda/2)]\subset[0.8125\,a,\,1.1875\,a]\), so
\(b_\sigma\) has a unique root \(r(\sigma)\in(0,2d)\), continuous in \(\sigma\). Since
\(b_\sigma(d)=\sigma ad(\mu/4+\lambda/24)\),
\[
|r(1)-r(-1)|\ \ge\ \frac{2ad(\mu/4+\lambda/24)}{1.1875\,a}\ \ge\ \frac{d(\mu+\lambda)}{15}=\frac{W}{15},
\]
so \(J:=r([-1,1])\) is an interval of length \(\ge W/15\) contained in \([d(1-v),d(1+v)]\).

Now fix any \(q\) locations in advance. At most \(q\) of them are distinct points of \(J\)
(repetitions and out-of-band locations only reduce the count), so they cut \(J\) into at most
\(q+1\) subintervals and one of these, \([\alpha,\alpha+L]\), has
\(L\ge|J|/(q+1)\ge W/(15(q+1))\ge W/(20q)\) for \(q\ge3\). Let \(m=\alpha+L/2\) and, by the
intermediate value theorem, pick \(\sigma^\star\) with \(r(\sigma^\star)=m\).

Let \(B(s)=s^3(1-s)^3\) on \([0,1]\) and \(0\) elsewhere. Then \(B,B',B''\) vanish at \(s=0,1\), so
the extension is globally \(C^{2}\) with \(\operatorname{Lip}(B'')=\max|B'''|=6\) (attained at the
endpoints), \(\max|B'|<1/16\), and \(B(1/2)=1/64\). Put
\(h(t)=\tfrac M8L^3B\big((t-\alpha)/L\big)\), so \(h''(t)=\tfrac M8LB''(\cdot)\) and
\(\operatorname{Lip}(h'')=\tfrac{6M}{8}=\tfrac{3M}{4}\). Define \(\phi_\pm=b_{\sigma^\star}\pm h\).
Then \(\operatorname{Lip}(\phi_\pm'')\le M/4+3M/4=M\); \(h\) vanishes to second order off
\((\alpha,\alpha+L)\subset(0,2d)\), so \(\phi_\pm\) share the origin jet
\((-ad,\,a,\,\sigma^\star K_0/2)\) of \(b_{\sigma^\star}\) and hence lie in the class; and
\(h\) vanishes at every one of the \(q\) fixed locations, so \(\phi_+\) and \(\phi_-\) produce an
identical transcript. Moreover \(|h'|\le\tfrac M8L^2\max|B'|\le Md^2/1152\le a\lambda/1152\), so
\(\phi_\pm'\in(0.8a,1.2a)\) on \([0,2d]\): each \(\phi_\pm\) is strictly increasing there with
\(\phi_\pm(0)<0\), so each has a unique root in \((0,2d)\), which is its first positive root
(first-root preservation). Since \(\phi_\pm(m)=\pm ML^3/512\),
\begin{equation}
|r_+-r_-|\ \ge\ \frac{2\,ML^3/512}{1.2\,a}\ \ge\ \frac{ML^3}{308\,a}
\ \ge\ \frac{1}{2.5\times10^{6}}\,\frac Ma\Big(\frac Wq\Big)^3=:G_q .
\label{S-eq:s1-batch-lower}
\end{equation}
(Both roots lie inside the bump support, since \(ML^3/(512\cdot0.8a)\le L/2\) whenever
\(Md^2\le a/8\).) A deterministic batch rule sees the same transcript on \(\phi_+\) and \(\phi_-\)
and emits one number, so its error is at least \(G_q/2\) on one of them; if it is uniformly
first-crossing-valid its output is \(\ge\max(r_+,r_-)\) and its overshoot on the other profile is
at least \(G_q\). With \eqref{S-eq:s1-batch-upper} this proves the theorem, with explicit constants
\(9216\) and \(1/(5\times10^{6})\) (estimation), \(18432\) and \(1/(2.5\times10^{6})\)
(valid overshoot). \hfill\(\square\)

\smallskip
\noindent\textit{Constant gap.} As elsewhere in this section, the two constants are rate-sharp but
not constant-sharp. They differ by about eleven orders of magnitude, and we make no attempt to close
that gap.

\smallskip
\noindent\textbf{Relation to optimal recovery and information-based complexity.}
The \(q^{-3}\) law is classical in kind. Recovering
a function with a Lipschitz second derivative from \(q\) fixed samples has error \(\Theta(Mh^3)\) at
spacing \(h\), the optimal-recovery answer for such classes is an interpolating spline with
canonical knots \citep{micchelli1976optimal}, and the general framework for minimax error under
fixed information, including the bump (resisting-oracle) converse used above, is
optimal recovery and information-based complexity
\citep{micchelli1977survey,traub1980general,traub1988ibc}. The one-query rules have classical
counterparts as well. Alpha-1 is one Newton step on \(\phi\) from \(t=0\); the signed-curvature
criterion is the convexity condition that decides whether a Newton step lands past the root or short
of it, and Theorem 5.2 is a two-sided, finite-scale form of the Newton error expansion
\citep{ostrowski1966solution,ortega2000iterative}. The exact-\(\kappa\) quadratic root is one step
of Euler's method, the root of the second-order Taylor polynomial
\citep{traub1964iterative,melman1997geometry}. The probe rule uses the same three values as one step
of Ostrowski's two-step method, \(\phi(0)\), \(\phi'(0)\) and \(\phi\) at the Newton point
\(d_p\) \citep{ostrowski1966solution}, whose fourth order \citet{kung1974optimal} conjectured to be
optimal for three evaluations; the \(d^4\)-type rate of Theorem 6.3 is the uniform, finite-scale
counterpart of that order. Likewise, the adaptive rule of the conjecture below is
Muller's method (direct quadratic interpolation), whose order is the tribonacci constant
\(\rho\approx1.8393\); this is due to \citet{muller1956method} and \citet{traub1964iterative},
and the conjectured certified radii would be a constant-explicit version
of that classical order, not a new convergence result. It is also classical that, in the
worst case, adaption cannot help for linear problems on convex, symmetric classes but can help for
nonlinear problems such as zero finding \citep{novak1996power,traub1988ibc}, and that under sign
information alone bisection is already minimax optimal \citep{sikorski1982bisection}.

Against this background, the contribution of this subsection is narrow.
(i) The recovered functional is not the function but the \emph{first root} of the ray profile, in a
class pinned by a two-sided origin jet \((\phi(0)=-ad,\ \phi'(0)=a,\ |\phi''(0)|\le K_0)\) plus
\(\operatorname{Lip}(\phi'')\le M\). This jet, and not the sample budget, localizes the root a
priori to a band of width \(\Theta(W)\), and \(W=K_0d^2/a+Md^3/a\) is exactly the zero-query rate of
Corollary~\ref{S-cor:s1-lattice}. The theorem therefore measures the query budget in the same
\((a,d,K_0,M)\) units as the rest of the paper. (ii) The loss is one-sided: a recommendation must
not undershoot the first crossing, and it pays for any excess. The minimax order under this
asymmetric loss is the same as under absolute error, with the converse gap entering in full rather
than halved; this is the one new observation, though an elementary one. (iii) The gap between
nonadaptive and adaptive rules is an instance of the classical adaption phenomenon, here with
explicit constants. We claim novelty only for (i) and (ii), as supplementary results;
we do not claim a new approximation-theoretic rate or adaptive minimax optimality (see
\emph{Open} below).

\smallskip
\noindent\textbf{Conjecture (adaptive successive collocation).} If locations may depend on earlier
values, we conjecture that the following successive-collocation rule does far better than any batch
rule. Query \(d\); take the root \(s_1\) of the Hermite
quadratic through \(\{0,0,d\}\); query \(s_1\) and take the root of the ordinary quadratic through the
three value nodes; thereafter query the current estimate and interpolate the latest three. The
conjecture is that the radii
\[
e_0=v,\quad e_1=\tfrac{25}{64}\lambda v,\quad e_2=\tfrac{375}{1408}\lambda^2v^2,\qquad
e_j=\tfrac{6\lambda}{11}\,e_{j-3}e_{j-2}e_{j-1}\ \ (j\ge3),
\]
with \(E_j=d\,e_j\), are certified: \(|s_q-d_{\rm ray}|\le E_q\) for every profile in the class, so
that outputting \(s_q+E_q\) would be uniformly first-crossing-valid with overshoot at most
\(2E_q\). The \(j=1\) radius has the order \(MK_0d^4/a^2+M^2d^5/a^2\) of Theorem 6.3. Taking
logarithms, \(\log e_j=\log(6\lambda/11)+\log e_{j-3}+\log e_{j-2}+\log e_{j-1}\), a
\emph{tribonacci} recursion with characteristic root \(\rho\), \(\rho^3=\rho^2+\rho+1\); the leading
small-\(d\) powers of \(E_0,\dots,E_6\) are \(2,4,7,13,24,44,81\) for nonzero \(K_0\) and
\(3,5,9,17,31,57,105\) at \(K_0=0\). We do not prove the conjecture here. The code archive contains
a written argument (S10), which this supplement neither reproduces nor relies on, and the numerical
check below. The iteration is Muller's method, whose order is classical (see above); what the
conjecture would add is a \emph{certified, class-explicit} radius.

\smallskip
\noindent\textbf{The separation.} Call a deterministic rule \emph{adaptive} if its \(j\)th query
location is a function of the jet, the class bounds and the values observed at the first \(j-1\)
locations, and put \(X=MK_0d^4/a^2+M^2d^5/a^2\). With \(q\ge1\) adaptive queries, query \(d\) as in
Theorem 6.3, whose proof places the root in \([r_p-E(\widehat\kappa),r_p+E(\widehat\kappa)]\) with
\(E(\widehat\kappa)\le4MK_0d^4/a^2+\tfrac43M^2d^5/a^2+\tfrac83M^2d^5/a^2=4X\le d/8\); this interval
lies in \((d/2,3d/2)\), where \(\phi\) is increasing, so the sign of \(\phi\) at its midpoint tells
which half contains the root. Bisecting with the remaining \(q-1\) queries leaves an interval of
length at most \(8X\,2^{1-q}\) that contains the root. Its midpoint has error at most \(4X\,2^{1-q}\),
and its right end is uniformly first-crossing-valid with overshoot at most \(8X\,2^{1-q}\). For fixed
\((a,d,K_0,M)\) the adaptive minimax error is therefore \(o(q^{-k})\) for every \(k\), whereas
Theorem~\ref{S-thm:s1-batch} gives the batch rate \(\Theta(q^{-3})\). For on-path certification,
\emph{when} probes are placed thus matters more than how many are used. At larger
budgets this plays the role that the separation between zero queries and one query plays in the
main paper.

\smallskip
\noindent\textbf{Open.} The matching adaptive converse is \emph{not} proved and no adaptive
minimax-optimality is claimed. The obstruction is explicit: for fixed final nodes \(C^{2,1}\) optimal
recovery is governed by a perfect spline, and what is missing is an adaptive common-completion lemma
building that spline along one entire decision-tree path under a single \(M\) budget. Nested compact
bumps preserve a common transcript but cube the preceding gap; a fresh full-budget kink per level
overspends \(M\); rebuilding a pair after each answer yields no single final pair. Randomized,
noisy, and finite-precision oracles are likewise outside these statements.

\smallskip
\noindent\textbf{Numerical check.} We evaluated the conjectured radii on the actual adaptive iterates
for smooth and genuinely kinked \(C^{2,1}\) profiles up to \(q=16\). Float64 is unusable beyond
\(q\approx4\), so this needs \(45{,}000\)-digit arithmetic, and the error reaches
\(\log_{10}(E_q/d)\approx-3.7\times10^{4}\). The fitted tribonacci slope is \(0.6104\) against
\(\log\rho=0.6094\), the batch slope in \(q-1\) is \(-2.998\) against \(-3\), and the fitted
parameter scalings match the predicted \(K_0^3Ma^{-4}d^6\) and, at \(K_0=0\), \(M^4a^{-4}d^9\). S10
lists the script, the result file, the pass criteria and the written argument for the conjecture.

\subsection{Leading-Order Average-Case Companion to Theorem 6.1}

Theorem 6.1 is a pointwise zero-query barrier. The following separate statement describes fixed-scale, scale-only rules and retains the finite-remainder qualification that a leading-order argument requires.

\smallskip
\noindent\textbf{Proposition (scale-only lower bound with remainder).}
\textit{Let \(\mathcal P\) be a population of rejected points (Notation) on which \(d_p\) and \(a\) are constant, put \(p:=\mathbb P(\kappa\ge0)\), and let \(\varepsilon\ge0\) be such that}
\[
\left|d_{\mathrm{ray}}-\left(d_p-\frac{d_p^2}{2a}\kappa\right)\right|\le\varepsilon
\quad\text{at almost every point.}
\]
\textit{A \emph{scale-only} rule recommends a distance that is a function of \((d_p,a)\) alone, hence one distance \(r=d_p+\Delta\) at every point. If its ray-hit validity \(v=\mathbb P(d_{\mathrm{ray}}\le r)\) satisfies \(v\ge p\), and the law of \(\kappa\) has a density at most \(M_\rho>0\) on \([-2a(\Delta+\varepsilon)/d_p^2,0)\), then its overshoot \([r-d_{\mathrm{ray}}]_+\) satisfies}
\[
\mathbb E[r-d_{\mathrm{ray}}]_+\ \ge\
p\left[\frac{d_p^2}{2a}\frac{v-p}{M_\rho}-2\varepsilon\right]_+ .
\]

\smallskip
\noindent\textit{Proof.} Let \(A=\{d_{\mathrm{ray}}\le r\}\) be the validity event. Since \(v-p=\mathbb P(A,\kappa<0)-\mathbb P(A^c,\kappa\ge0)\le\mathbb P(A,\kappa<0)\), it is enough to bound valid negative-curvature points. Each such point satisfies
\(d_p-d_p^2\kappa/(2a)\le d_{\mathrm{ray}}+\varepsilon\le r+\varepsilon\), hence
\(\kappa\ge-2a(\Delta+\varepsilon)/d_p^2\). If \(\Delta+\varepsilon<0\), no point qualifies, so \(v=p\) and the right side is zero. Otherwise the density bound gives
\(v-p\le M_\rho\,2a(\Delta+\varepsilon)/d_p^2\), or
\(\Delta\ge d_p^2(v-p)/(2aM_\rho)-\varepsilon\). At almost every point with \(\kappa\ge0\), the same approximation gives \(d_{\mathrm{ray}}\le d_p+\varepsilon\), so the overshoot there is at least \([\Delta-\varepsilon]_+\). Averaging over \(\{\kappa\ge0\}\), of probability \(p\), yields the result. A curvature-aware rule is not scale-only, so the proposition does not apply to it. Theorem 5.2 and Lemma 5.1 make \(\varepsilon\) small when regime \((\star)\) holds with small \(|\kappa|d_p/a\) and \(Md_p^2/a\); at finite scale the bound can be vacuous. \hfill\(\square\)

\smallskip
\noindent\textbf{Remark (scope of the rule class).} The proposition covers only rules that recommend
one distance over the fixed-scale population. At fixed \((d_p,a)\) the gradient direction \(\hat g\)
still varies, so a rule that reads it can recommend different distances at points with different
curvature; the argument above does not apply to such rules, and we give no lower bound for them
(Theorem 6.1 is unaffected: there both scores share a single point and jet). A randomized scale-only
rule draws \(\Delta\) from a law that does not depend on the point. Conditionally on \(\Delta\), its
validity \(v(\Delta)\) satisfies the proposition's bound when \(v(\Delta)\ge p\), and the bound's
right side is zero otherwise. Since \(w\mapsto p[d_p^2(w-p)/(2aM_\rho)-2\varepsilon]_+\) is convex,
Jensen's inequality gives the same bound with \(v=\mathbb E\,v(\Delta)\), provided the density
hypothesis holds for every realized \(\Delta\).

\noindent\textbf{Measured illustration.} The leading slope \((d_p^2/2a)/M_\rho\) depends on a \emph{population} density bound, which finite samples do not verify, so we report no estimate of \(M_\rho\). On COMPAS (unregularized models, seeds 0--2, all 1{,}316 rejected test points) the observed rejected curvatures concentrate in \(|\kappa|<0.22\) (\(63\%\) within \(|\kappa|<0.1\)), and the sample has \(\mathbb P(\kappa\ge0)=0.72\) with minimum \(\kappa=-0.110\). A target \(v\in[0.72,1]\) therefore places the threshold in the observed concave range \([-0.110,0)\), although the sample does not establish the assumed density envelope there. On Adult (unregularized models on the 20{,}000-row cohort, seeds 0--1, 400 randomly chosen rejected test points each) the sample is almost entirely concave (\(\mathbb P(\kappa\ge0)=0.015\), median \(\kappa=-1.09\)), and targets above \(0.015\) move the threshold into the concave bulk. No uniform \(\varepsilon\) is certified for these populations, so the measurements do not show that the finite-remainder bracket of the proposition is positive on either dataset. They illustrate the leading-order scale-only bound; they do not verify a nonzero population lower bound.

\subsection{Proof of Proposition 7.1 (Conformal First-Crossing Coverage-or-Abstention)}

\noindent\textit{Outline.} The test residual \(d_{\text{ray}}-d_{\text{quad}}\) is exchangeable with the \(n\) calibration residuals, so it lies at or below their \(\lceil(1-\delta)(n+1)\rceil\)-th order statistic \(\hat q_{1-\delta}\) with probability \(\ge1-\delta\). If \(\hat q_{1-\delta}=+\infty\) the rule abstains; otherwise this rank event implies \(d_{\text{ray}}\le\max\{0,d_{\text{quad}}+\hat q_{1-\delta}\}\). Conditioning on the group labels reruns the argument within each group (Mondrian).

Let \(\delta\in(0,1)\), let \(r_i=d_{\text{ray}}(x_i)-d_{\text{quad}}(x_i)\) (with the main text's \(2d_p\) fallback for \(d_{\text{quad}}\)) and \(\hat q_{1-\delta}=r_{(k)}\) with \(k=\lceil(1-\delta)(n+1)\rceil\in\{1,\dots,n+1\}\) (and \(\hat q_{1-\delta}=+\infty\) if \(k>n\)). Under exchangeability of the calibration residuals and the test residual \((r_1,\dots,r_{n+1})\), randomized tie-breaking makes the augmented rank of \(r_{n+1}\) uniform on \(\{1,\dots,n+1\}\); using the weak event without randomization is conservative when ties occur. Thus \(\Pr[r_{n+1}\le\hat q_{1-\delta}]\ge1-\delta\): for \(k\le n\), the augmented-rank bound is at least \(k/(n+1)\), while for \(k=n+1\) it is one by the \(+\infty\) convention. No continuity assumption is needed. For finite \(\hat q_{1-\delta}\), the rank event implies \(d_{\text{ray}}\le\max\{0,d_{\text{quad}}+\hat q_{1-\delta}\}\); a non-finite quantile is abstention. Hence \(\Pr[\text{abstain or }d_{\text{ray}}\le d_{\text{conf}}]\ge1-\delta\); abstention counts toward this event, so the bound is not conditional on issuing a certificate. The argument uses only the exchangeability of the residuals, so it holds unchanged when \(d_{\text{quad}}\) is computed from the probe estimate \(\hat\kappa\) or from any other base recommendation fixed independently of the calibration data. A covered first on-ray crossing implies favorable classification at the recommended endpoint only when no recrossing occurs on the intervening segment; Lemma~\ref{S-lem:s1-no-recross} supplies that implication for in-regime recommendations no longer than \(2d_p\), while the segment-safe fallback makes no conformal-coverage claim on its fallback cases.

\emph{Mondrian version} \citep{vovk2005algorithmic}. Let \(A_i\) be the protected attribute of point \(i\), which takes finitely many values, with the test point \(i=n+1\), and suppose that the \(n+1\) pairs \((r_i,A_i)\) are exchangeable. For a test point with \(A_{n+1}=c\), let \(n_c\) be the number of calibration points with \(A_i=c\) and \(\hat q^{(c)}_{1-\delta}\) the \(\lceil(1-\delta)(n_c+1)\rceil\)-th smallest of their residuals (\(+\infty\) if this index exceeds \(n_c\)); the per-group rule abstains if \(\hat q^{(c)}_{1-\delta}=+\infty\) and otherwise recommends \(d_{\text{conf}}=\max\{0,d_{\text{base}}+\hat q^{(c)}_{1-\delta}\}\). Conditionally on the label vector \((A_1,\dots,A_{n+1})\), every permutation of the points that preserves each label leaves the law of \((r_1,\dots,r_{n+1})\) unchanged, so the \(n_c+1\) residuals of group \(c\) are exchangeable and \(n_c\) is fixed. The argument above, applied to them, gives \(\Pr[\text{abstain or }d_{\text{ray}}\le d_{\text{conf}}\mid A_1,\dots,A_{n+1}]\ge1-\delta\) on \(\{A_{n+1}=c\}\), and averaging over the labels gives the same bound conditionally on \(A_{n+1}=c\) for every \(c\) with \(\Pr(A_{n+1}=c)>0\).

\paragraph{Small Mondrian groups.} When a protected group has fewer than \texttt{min\_group} calibration residuals, or a non-finite group quantile (at \(\delta=0.05\) a finite \(95\%\) quantile needs at least \(19\) calibration points), the implementation marks the whole Mondrian row as a vacuous certificate. It does not substitute the pooled quantile, which would carry only marginal coverage. This matters for German: with about \(40\)--\(50\) calibration rays split across two groups it cannot support finite group quantiles, so every per-seed result marks German--Mondrian \texttt{vacuous\_certificate}, and the \(\delta=0.05\) aggregate reports it as zero non-vacuous cells. Two \(\delta=0.01\) aggregate files in the code archive were produced without this handling and contain a finite German Mondrian row from a pooled-quantile fallback; no number in the paper or the supplement is drawn from them (S10).

\smallskip
\noindent\textit{Bracket-failure convention.} A calibration or test ray with no crossing inside the
\(8d_p\) search bracket is assigned residual \(r=+\infty\). When the calibration quantile
\(\hat q_{1-\delta}\) is finite, the event \(\{r_{\rm test}\le\hat q_{1-\delta}\}\) then coincides
exactly with crossing found and covered, so the rank guarantee is unconditional with respect to
test-side bracket failure. If the calibration quantile is \(+\infty\), the procedure abstains and
reports the certificate as vacuous; it does not treat the extended-real comparison
\(+\infty\le+\infty\) as a successful recommendation. On the shallow suites the bracket-failure
rate is exactly zero in all cells, so this convention changes no reported number.
We also reran the rules with the clip \(\max\{0,\text{base}+\hat q\}\) applied to every finite HVP,
probe and Mondrian recommendation, keeping all calibration, test, vacuous, unbracketed and
segment-fallback statuses. No issued recommendation is negative, every downstream outcome and
aggregate is unchanged, and a second, independent run produces identical files.

\subsection{Proofs of the Distributional Consequences (Theorem~\ref{S-thm:s1-shift}, Remark 8.1)}

\noindent\textbf{Theorem \suppnum{S-thm:s1-shift} (fixed-scale shift identity with boundary layer).}
\textit{Let \(S\) and \(T\) be source and target populations of rejected points (Notation) on which
\(d_p\) and \(a=\|\nabla f\|\) take the same constant values, and let \(1\le\alpha\le2\). Assume that
at almost every point of either population \(\phi''\) is \(M\)-Lipschitz on \([0,2d_p]\), with a
common \(M\), and regime \((\star)\) holds, so that there is no recrossing on \([0,2d_p]\)
(Lemma~\ref{S-lem:s1-no-recross}). For \(P\in\{S,T\}\) let \(V_P(\alpha)=\mathbb P_P(d_{\text{ray}}\le\alpha d_p)\)
be the first-crossing validity of \(r=\alpha d_p\), let \(F_P(b)=\mathbb P_P(\kappa\le b)\) be the
curvature CDF, assumed continuous, and define}
\[
b_\alpha=-\frac{2a(\alpha-1)}{\alpha^2d_p},
\qquad e_\alpha=\frac{M\alpha d_p}{3}.
\]
\textit{Then}
\[
1-F_P(b_\alpha+e_\alpha)\ \le\ V_P(\alpha)\
\le\ 1-F_P(b_\alpha-e_\alpha).
\]
\textit{For quadratic ray profiles (\(M=0\)), \(V_P(\alpha)=1-F_P(b_\alpha)\). Hence, if \(\alpha\in[1,2]\) satisfies
\(V_S(\alpha)=v\), then \(F_S(b_\alpha)=1-v\) (so \(b_\alpha\) is the \((1-v)\)-quantile of \(F_S\) when that quantile is unique),
and the deployed drop is exactly}
\[
D:=V_S(\alpha)-V_T(\alpha)=F_T(b_\alpha)-F_S(b_\alpha).
\]
\textit{At finite \(M\), \(D\) differs from \(F_T(b_\alpha)-F_S(b_\alpha)\) by at most the sum of the
source and target probability masses of
\([b_\alpha-e_\alpha,b_\alpha+e_\alpha]\). The expression \(F_T(b_\alpha)-F_S(b_\alpha)\) is
nonnegative when the target curvature is stochastically smaller than the source curvature
(\(F_T\ge F_S\)) and nonpositive in the reverse case, and, writing
\(\Delta p_+=\mathbb P_T(\kappa\ge0)-\mathbb P_S(\kappa\ge0)\), it tends to
\(-\Delta p_+\) as \(\alpha\to1^+\). In particular, if \(M=0\) and the target curvature is stochastically
smaller than the source curvature, then \(D\ge0\).}

At a common scale, inflating the step by \(\alpha\) thus succeeds, up to a boundary layer of width \(2e_\alpha\), exactly for users whose path curvature is at least the threshold \(b_\alpha\le0\), so a factor calibrated on the source loses, on the target, the extra curvature mass below that threshold. The covariate and subpopulation splits of S4 (Deployment-Shift Robustness) move \((d_p,a)\) together with the curvature, so they fall outside the constant-scale assumption, and the identity rests on its proof alone.

\smallskip
\noindent\textit{Proof.} Taylor expansion at the recommended endpoint gives
\[
\phi(\alpha d_p)
=a d_p(\alpha-1)+\frac{\kappa}{2}\alpha^2d_p^2+R(\alpha d_p)
=\frac{\alpha^2d_p^2}{2}(\kappa-b_\alpha)+R(\alpha d_p),
\]
with \(|R(\alpha d_p)|\le M\alpha^3d_p^3/6\). In the stated regime,
Lemma~\ref{S-lem:s1-no-recross} makes first-crossing validity equivalent to endpoint success.
Therefore \(\kappa\ge b_\alpha+e_\alpha\) is sufficient and
\(\kappa\ge b_\alpha-e_\alpha\) is necessary, so
\(\mathbb P_P(\kappa\ge b_\alpha+e_\alpha)\le V_P(\alpha)\le\mathbb P_P(\kappa\ge b_\alpha-e_\alpha)\), which is the
stated bracket because \(F_P\) is continuous. When \(M=0\)
the event is exactly \(\{\kappa\ge b_\alpha\}\), so \(V_P(\alpha)=1-F_P(b_\alpha)\); calibration gives
\(F_S(b_\alpha)=1-v\) (with the stated unique-quantile refinement), and subtracting the two
tail probabilities gives the identity.
At finite \(M\), \(1-F_P(b_\alpha)\) lies in the same bracket as \(V_P(\alpha)\), so the two differ by at most
\(F_P(b_\alpha+e_\alpha)-F_P(b_\alpha-e_\alpha)\), the \(P\)-mass of \([b_\alpha-e_\alpha,b_\alpha+e_\alpha]\);
subtracting the source and target statements gives the error bound.
Stochastic dominance fixes the sign of \(F_T(b_\alpha)-F_S(b_\alpha)\). Finally
\(b_\alpha\to0\) as \(\alpha\to1^+\), and continuity of \(F_S\) and \(F_T\) at \(0\) gives
\(F_T(b_\alpha)-F_S(b_\alpha)\to F_T(0)-F_S(0)=\mathbb P_S(\kappa>0)-\mathbb P_T(\kappa>0)=-\Delta p_+\). \hfill\(\square\)

\smallskip
\noindent\textbf{Remark (heterogeneous scale).} If \((d_p,a)\) varies pointwise, validity is a
functional of their \emph{joint} law with \(\kappa\), not a curvature-marginal CDF plus an
identified scalar correction. On the COMPAS covariate split the near-zero marginal heuristic is numerically
close to the measured drop (35.1 versus 33.6 points), but because scale also shifts this agreement
is descriptive and does not identify the curvature-CDF contribution separately.

\smallskip
\noindent\textbf{Remark 8.1 (fixed-ray leading statistic).} \textit{Fix a common first-order jet \((m,a,\hat g)\), hence a common \(d_p=m/a\), and two \(C^{2,1}\) ray profiles satisfying the same directional-curvature Lipschitz bound \(M\). If their Hessians at \(x\) obey \(\hat g^\top H\hat g=\hat g^\top H'\hat g=\kappa\), then in regime \((\star)\) their first on-ray crossings differ by at most twice the slab \(8Md_p^3/(3a)\) of Theorem 5.2. Thus the alpha-1 event and on-path gap depend on the Hessian at \(x\), to leading order on this fixed ray, only through \(\kappa\).}

\smallskip
\noindent\textit{Proof.} Along either shared ray, \(\phi(0)=-m\), \(\phi'(0)=a\), and \(\phi''(0)=\kappa\). Directional \(M\)-Lipschitz curvature gives \(\phi(t)=-m+at+\tfrac{\kappa}{2}t^2+R(t)\), where \(R(t)=\int_0^t(t-s)(\phi''(s)-\kappa)\,ds\) and \(|R(t)|\le Mt^3/6\). Theorem 5.2 therefore places each first crossing within \(8Md_p^3/(3a)\) of the same quadratic root, and the triangle inequality gives the stated comparison. Off-ray entries of \(H\) do not enter this leading quadratic term; higher-order variation along the ray remains inside the common remainder bound. The corresponding threshold event \(\{\kappa\ge\kappa^\ast\}\) is likewise only a leading-order event, not an exact identity for arbitrary profiles.

\noindent\textit{Consequence.} \(\|H\|_F^2=\mathrm{tr}(H^2)=\sum_i\lambda_i^2\), \(\|H\|_{\mathrm{op}}\), and a Hutchinson estimate of \(\mathrm{tr}(H^2)\) are functions of all of \(H\), not of \(\kappa\) alone, and are even under \(H\mapsto-H\). At a fixed shared jet such an instantaneous penalty constrains off-ray curvature and assigns equal value to the curvature-flipped mirror pair \(\phi_\pm\) of Theorem 6.1, whereas \([\delta-\kappa]_+^2\) is directional and one-sided. This local comparison does not assert that models produced by the different training objectives share a jet or curvature distribution; their trained performance is an empirical question. \hfill\(\square\)

\subsection{Staleness Monitoring and Zero-Search Repair}
\label{S-sec:supp-shift-monitor}

Theorem~\ref{S-thm:s1-shift} is a fixed-scale curvature statement with an explicit boundary layer. This
subsection instead gives separate score-space statements that are exact without that approximation.
Weighted conformal prediction \citep{tibshirani2019conformal} needs density ratios; instead, an exact lemma bounds the drop of every inflation factor by the Kolmogorov--Smirnov distance between the source and target laws of the pivot \(d_{\text{ray}}/d_p\). Estimating it needs offline ray searches on the target, so it serves as an empirical staleness \emph{alarm}.

\smallskip
\noindent\textbf{Lemma (exact KS transport).} \textit{Let \(S\) and \(T\) be source and target
populations (Notation), and for laws \(\mu,\nu\) on \((-\infty,+\infty]\) let
\(\mathrm{KS}(\mu,\nu)=\sup_{t\in\mathbb R}|\mu((-\infty,t])-\nu((-\infty,t])|\). For any deployed rule
whose recommended distance \(r=\psi(\mathrm{jet}(x))\) is a Borel function of the first-order jet,
define the score \(U:=d_{\text{ray}}-\psi\) (\(+\infty\) if the ray does not cross) and the validity
\(V_P=\mathbb P_P(U\le0)\). Then \(|V_S-V_T|\le\mathrm{KS}(\mathrm{law}_S\,U,\ \mathrm{law}_T\,U)\); and
for the inflation family \(\{r=\alpha d_p:\alpha>0\}\), \(\sup_{\alpha>0}|V_S(\alpha)-V_T(\alpha)|\)
\emph{equals} the KS distance between the source and target laws of the pivot
\(d_{\text{ray}}/d_p\in(0,+\infty]\).} \emph{Proof.} Validity is \(\mathbb P(U\le0)\), a CDF evaluated
at a point, and a difference of CDFs at a point is at most their sup-distance. For the inflation
family, \(V_P(\alpha)=\mathbb P_P(d_{\text{ray}}/d_p\le\alpha)\), so the family sweeps every positive
threshold of the pivot; since \(d_{\text{ray}}>0\) and \(d_p>0\), both pivot CDFs vanish at every
\(t\le0\), and the supremum over \(\alpha>0\) is the KS distance. \(\square\) No fixed-scale,
continuity, or leading-order assumption enters. Under the hypotheses of Theorem~\ref{S-thm:s1-shift},
its fixed-scale \(\kappa\)-CDF expression approximates this exact score-space statement with the
displayed source and target boundary masses. That expression concerns inflation only. For
residual-threshold rules such as stale conformal we give no \(\kappa\)-KS certificate;
recalibrating on the target is the appropriate response. The repair below addresses stale
inflation instead.

\smallskip
\noindent\textbf{Probe plug-in.} Corollary 6.2 bounds the probe-quadratic score's distance from
the exact pivot in regime, but transporting that pointwise error to a KS bound additionally
requires source and target score-CDF moduli at the error scale (plus the usual DKW terms).
Without certified common \(M\) and those moduli, a probe-only pivot KS is an empirical alarm,
not a certificate. The covariate-split audit below therefore reports empirical exact-ray pivots
from offline ray searches; it needs no density ratios, but it requires those searches and gives no
finite-sample population certificate.

\smallskip
\noindent\textbf{Zero-search two-condition repair (finite-cohort scope).} Fix the trained model, the
empirical-median target mask, and the resulting finite target-rejected cohort of \(N\) points before
drawing probe indices, and draw the \(n\) probes uniformly at random without replacement. Let \(R\)
be the regime-\((\star)\) screen of S3.1, fixed in advance, let \(N_R\) be the number of cohort
members that pass it, and let \(n_R\) be the number of screen-passing probes. Choose the smallest
\(\alpha\) of a fixed grid in \([1,2]\) satisfying \emph{both}
(i) every filtered probe succeeds and \((1-\delta)^{n_R}\le\eta\), and (ii) the original
all-probe Clopper--Pearson lower-bound screen; abstain if none qualifies. Condition (ii) is only
an empirical conservatism screen.

Given the fixed cohort, every set of \(n\) probes is equally likely, so for every set \(B\) of
\(k\) \(R\)-members the joint probability that the filtered probes are exactly \(B\) (and hence
\(n_R=k\)) is \(\binom{N-N_R}{n-k}/\binom{N}{n}\), which depends on \(B\) only through \(k\).
Conditionally on the cohort and on \(n_R=k\), every \(k\)-subset of the \(R\)-cohort therefore has
probability \(1/\binom{N_R}{k}\): the filtered probes are a simple random sample without
replacement from the finite \(R\)-cohort. At a
fixed \(\alpha\), if \(K_R\) of the \(N_R\) members succeed and
\(p_R(\alpha)=K_R/N_R\), then
\[
\Pr(\hbox{all }n_R\hbox{ probes succeed})
=\frac{\binom{K_R}{n_R}}{\binom{N_R}{n_R}}
=\prod_{j=0}^{n_R-1}\frac{K_R-j}{N_R-j}
\le p_R(\alpha)^{n_R}.
\]
Under Lemma~\ref{S-lem:s1-no-recross}, the screened endpoint indicators are nested in \(\alpha\). If the selected
grid point had finite-cohort \(p_R<1-\delta\), all sampled points would also succeed at the largest
fixed grid point whose \(p_R<1-\delta\), an event of probability at most
\((1-\delta)^{n_R}\le\eta\). Condition (ii) can only move the selection upward, where
\(p_R\) is nondecreasing. Thus condition (i) gives a design-based statement: conditionally on the
cohort and on \(n_R\), the probability that the rule does not abstain and selects an \(\alpha\) with
\(p_R(\alpha)<1-\delta\) is at most \(\eta\), so with confidence \(1-\eta\) the selected rule covers
at least a \(1-\delta\) fraction of the realized finite \(R\)-cohort.
The nesting uses Lemma~\ref{S-lem:s1-no-recross}, so the statement assumes that every \(R\)-member
satisfies the hypotheses of Theorem 5.2; the screen of S3.1 establishes this only under its
stored-jet and numerical-library assumptions for the directional-\(M\) enclosure. It gives no
superpopulation guarantee and no guarantee for screen-rejected points or the full target cohort.

\smallskip
\noindent\textbf{Measurements (stored results and a five-seed demonstration).} The KS statistic
serves only as an \emph{alarm}. On the COMPAS covariate split the \emph{exact} pivot of the lemma,
\(d_{\text{ray}}/d_p\) with unbracketed rays set to \(+\infty\), has KS \(0.342\) (per seed
\(0.303\)--\(0.408\)); the observed stale-inflation drop is \(0.336\), so the bound is attained to
within \(0.006\). The zero-search quadratic surrogate \(2/(1+\sqrt{1+2\kappa d_p/a})\) gives
\(0.358\), while the KS distance of \(\kappa\) is \(0.734\), which shows why score space is the
relevant variable under a change of scale. The per-point file behind these values uses a scaler
fit on the training split only (S10).
The two-condition rule selects the same \(\hat\alpha\in[1.28,1.39]\) as the original all-probe
screen. Its filtered counts are all successes (\(86/86\)--\(93/93\)), and the largest tail bound for
a bad cohort is \(0.0121\). The design-based statement above therefore applies to the realized
finite \(R\)-cohort, conditional on the screen and enclosure assumptions. Separately, the all-probe
screen observes \(99/100\) successes, and on the full target the repair raises the stale rule from
\(66.4\%\) to \(98.8\)--\(99.7\%\) measured validity (5/5 seeds \(\ge0.93\)), with identical values
under the ray and endpoint definitions. No certificate covers these full-target figures.
The price is an overshoot of \(0.28\)--\(0.40\), an order of magnitude above the \(0.023\) of
conformal recalibration, so avoiding the search is paid for in effort.

\section{S2. Metric Audit and Estimator Diagnostics}

The paper distinguishes the endpoint event and three distances. The promised first-order distance is \(d_p=|f(x)|/\|\nabla f(x)\|_2\). The ray distance \(d_{\text{ray}}\) is the first crossing along the recommendation path; its signed gap supplies effort diagnostics. The global nearest-boundary distance \(d_\star\) solves the unconstrained \(L_2\) boundary problem and is diagnostic only. Endpoint success is the primary user event, and outside monotone regimes the early-hit mass \(r_t\) separates it from the ray-hit event.

DeepFool-style iterated linear estimators \citep{moosavidezfooli2016deepfool} update using \((|f|/\|\nabla f\|^2)\nabla f\). An unrefined iterate is tied to the same local factor as \(d_p\), so it is not an independent check of the reliability of a fixed path. The observation is not an impossibility result and does not make DeepFool invalid: a converged, boundary-refined DeepFool can estimate the nearest-boundary distance accurately.

Our diagnostic estimator is more decoupled. It solves a Carlini--Wagner-style constrained objective \citep{carlini2017towards},
\[
\min_\delta \|\delta\|_2^2+c[-f(x+\delta)]_+,
\]
using Adam, binary search on \(c\), cosine decay, restarts, and one boundary warm-start. It searches all directions and does not use the ratio \(|f|/\|\nabla f\|\). Against a high-resolution two-dimensional contour reference its error is near numerical precision.

\paragraph{Endpoint-agreement audit.} The implementation records validity as the first-crossing
event \(d_{\text{ray}}\le d_p\), whereas a recommendation succeeds when its endpoint satisfies
\(f(x+d_p\hat g)\ge0\). We therefore measure the agreement of the two events across the full
shallow suite (4 datasets \(\times\) 4 methods \(\times\) 5 seeds; \(24{,}189\) test points). Raw
agreement is \(99.55\%\) pooled. Every one of the 109 disagreements is boundary-coincident, with
\(|f(x+d_p\hat g)|\le3.3\times10^{-6}\) and \(d_{\text{ray}}\) equal to \(d_p\) at the ray
tolerance, and \emph{none} is a true re-crossing (a crossing strictly before \(d_p\) with a clearly
unfavorable endpoint). The common-cohort replay of S4 uses direct endpoint success and the
curvature sign on identical held-out IDs and gives \(r=0.985\) over the 80 model cells. For the
conformal certificate, endpoint coverage equals first-crossing coverage in every cell (no conformal
re-crossing anywhere), as Lemma~\ref{S-lem:s1-no-recross} predicts in regime. The same audit measures
the segment condition of the lemma for the conformal rule (S1): \(98.1\%\) of test points pooled
satisfy \(d_{\text{conf}}\le2d_p\), with a per-cell minimum of \(91.5\%\) on unregularized Adult.
The exceptions are near-boundary points with tiny \(d_p\), where the shared calibration quantile
dominates and \(c=d_{\text{conf}}/d_p\) reaches \(10^3\). Equivalently, the segment-safe rule of S1
abstains from the conformal correction for \(1.9\%\) of points pooled and returns alpha-1 there.
This restores endpoint equivalence at the level of the theorem in regime \((\star)\), at the price
of no conformal-coverage claim on the fallback subset.

\paragraph{Analytic high-dimensional boundaries.} Spheres \(f(x)=\|x-c\|^2-r^2\) and ellipsoids \(f(x)=(x-c)^\top A(x-c)-1\) give independent nearest-boundary references, in closed form for spheres and by a high-accuracy scalar Lagrange-multiplier solve for ellipsoids. Rejected points are sampled uniformly in volume. Across the dimensions and families tested, the C\&W-style diagnostic stays close to these references. The code archive also contains a comparison with the unrefined DeepFool iterate, from which we draw no conclusion about the relative accuracy of the two estimators: the DeepFool arm did not receive the same endpoint refinement, and with equal refinement DeepFool is accurate on the analytic sphere case. A quantitative comparison would need the same refinement budget for both estimators. These analytic families say nothing about performance on arbitrary high-dimensional boundaries.

\section{S3. Experimental Setup}

\textbf{Datasets.} COMPAS uses the ProPublica two-year recidivism dataset \citep{angwin2016machinebias} with standard filtering and race in \{African-American, Caucasian\}. German uses Statlog \texttt{german.data} from UCI \citep{dua2017uci}. Adult uses the full 14-feature public UCI schema \citep{dua2017uci} (\(d=100\) after one-hot encoding of the categorical variables), subsampled with a fixed seed to 20,000 for the main comparison and to a nested 8,000-row cohort for the penalty sweep and the other suites marked ``Adult (8k)'' in S3.2. Fashion-MNIST (F-MNIST) uses official IDX files \citep{xiao2017fashion} for Pullover versus Coat.

\textbf{Training and solvers.} The main tabular MLP has two Softplus layers, Adam at \(10^{-3}\), binary cross entropy with class reweighting, and deterministic seeds. F-MNIST uses a larger Softplus MLP, trains for \(15\) epochs, and in the driver of Table~\ref{S-tab:supp-fmnist} (\texttt{fm\_final.py}) uses unweighted BCE; its classes are balanced at \(50.1\%\) positives, so the weighting is numerically immaterial. The recourse-ray line search uses every rejected test point unless the suite's description states a cap. The real-data nearest-boundary quantity is a C\&W-style solver estimate on a seeded random cap of 128 rejected points per seed; it is a diagnostic, not an exact distance. Equal-cell summaries weight each dataset\(\times\)method cell equally; seed-weighted summaries weight each model seed equally; point-weighted summaries are explicitly labeled and weight rejected points.

\textbf{Cost of the curvature penalty.} The asymmetric and Hutchinson penalties use one Hessian-vector product per example and step. Materializing the full \(d\times d\) Hessian by automatic differentiation would require \(d\) such products; for F-MNIST, \(d=784\), the number of products that main-paper Section~8 quotes. Training also backpropagates through this product, so its runtime exceeds that of one HVP. This count is analytic: the main paper's statements about computational cost count Hessian-vector products, and we report no wall-clock measurement for F-MNIST. The per-batch HVP wall-clock times of the CIFAR-10 networks are stored with their results (S10).

\subsection{S3.1. Measured Directional-Curvature Variation and Regime Screening}

Theorem 5.2 assumes that the directional-curvature profile is \(M\)-Lipschitz on the ray and holds in regime \((\star)\): \(4|\kappa|d_p\le a\) and \(8Md_p^2\le a\). We screen this condition on trained models. For each rejected test point we fix \(\hat g\), compute \(\kappa(t)=\hat g^\top H(x+t\hat g)\hat g\) at \(t/d_p\in\{0,.125,.25,.375,.5,.75,1,1.25,1.5,1.75,2\}\), and take \(\hat M\) as the largest secant slope of \(\kappa(t)\) between consecutive grid points. This \(\hat M\) is a data-driven \emph{lower} bound on the required Lipschitz constant. A failure at a point shows that the condition is violated there (up to the numerical tolerance), whereas a pass only means that the finite grid did not reject it. The median \(\hat M\) and the pass fraction barely depend on the grid (COMPAS \(0.018\)/\(93.8\%\) on the eleven-point grid and \(0.018\)/\(93.9\%\) on a coarser six-point grid), and a nonparametric bootstrap gives a \(95\%\) upper confidence bound on the median of \(\hat M\) (COMPAS \(0.019\), German \(0.027\), Adult \(0.65\)). Neither fact turns a pass at a single point into a certificate.

\begin{table}[h]
\centering\small
\caption{Finite-grid directional-\(M\) screen on real models (unregularized unless noted; tabular rows: 2 seeds except COMPAS-MW (seed 0), \(\le\!400\) rejected points; CelebA depth rows: ResNet-18, 3 attributes \(\times\) 5 seeds, 256 points/model, pooled; the same eleven-point ray grid on \([0,2d_p]\) throughout). ``p90 stress'' substitutes the measured p90 of \(\hat M\) for every point as a sensitivity analysis, not an upper-bound certificate. Slab containment is the empirical fraction with \(|(d_{\text{ray}}-d_p)+\kappa d_p^2\rho/(2a)|\le 8\hat M d_p^3/(3a)\).}
\label{S-tab:supp-Maudit}
\begin{tabular}{lrrrrr}
\hline
Data & med.\ \(\hat M\) & screen pass & p90 stress & slab cont. & crit.\ err.\ pass/reject \\
\hline
COMPAS & 0.018 & 93.8\% & 75\% & 99.1\% & 0.021 / 0.000 \\
German & 0.024 & 100\% & 97\% & 97.9\% & 0.000 / -- \\
Adult & 0.61 & 68.2\% & 60\% & 99.3\% & 0.000 / 0.035 \\
COMPAS (MW) & 0.008 & 98.5\% & 89\% & 99.3\% & 0.041 / 0.000 \\
CelebA depth (unreg) & 138 & 7.8\% & 3.9\% & 99.6\% & 0.123 / 0.423 \\
CelebA depth (Asym) & 42 & 10.5\% & 7.7\% & 99.4\% & 0.171 / 0.366 \\
\hline
\end{tabular}
\end{table}

\paragraph{Slab containment.} Containment is \(98\)--\(99\%\) throughout. Because \(\hat M\) is a lower bound, this agreement is consistent with the remainder analysis but cannot validate its hypothesis.

\paragraph{Pass rates on tabular data.} The screen leaves \(94\)--\(100\%\) of COMPAS and German points unrejected, against only \(68\%\) of unregularized Adult points. On Adult the binding condition is \(8\hat Md_p^2\le a\) (\(68.4\%\) pass, against \(99.5\%\) for \(4|\kappa|d_p\le a\)), because both \(d_p\) and \(\hat M\) are larger. Since the screen can only reject, we call passing points \emph{not rejected by the finite-grid screen}, never regime-eligible. On Adult the disagreement between alpha-1 validity and \(\mathbb P(\kappa\ge0)\) is \(0.000\) among passing points and \(0.035\) among rejected ones, the split the theory predicts; the comparison is descriptive.

\paragraph{The p90 stress test.} Substituting the p90 of \(\hat M\) for every point gives pass fractions of \(75\%\) (COMPAS), \(97\%\) (German), \(60\%\) (Adult) and \(89\%\) (MW-COMPAS). The p90 of lower bounds is not an upper bound, so these values do not certify the unobserved variation between grid points. MW-Hutchinson more than halves the observed median \(\hat M\) on COMPAS, with a \(98.5\%\) pass fraction.

\paragraph{Depth.} Most points at depth fail the lower-bound screen. On ResNet-18/CelebA (30 models: 3 attributes \(\times\) \{unreg, Asym\} \(\times\) 5 seeds, 256 rejected points each; recipe below) the pooled unregularized median \(\hat M\) is \(138\), nearly four orders of magnitude (\(\sim\!10^4\times\)) above the \(0.018\) of COMPAS, and it ranges over \(84\)--\(250\) across attributes. Only \(7.8\%\) of points pooled (\(5.2\)--\(12.2\%\) by attribute) are not rejected; on the rest \(8\hat Md_p^2\le a\) fails. Slab containment is \(94\)--\(95\%\) among passing points and \(\ge99\%\) overall (almost vacuous among rejected points, where the large \(\hat M\) widens the slab), and the criterion error is \(0.12\) for passing points against \(0.42\) for rejected ones. This association places the observed degradation at the points that fail a necessary smallness condition and is consistent with the remainder mechanism. It does not prove a remainder bound at depth or explain causally why the criterion degrades there. Asym lowers the observed median \(\hat M\) from \(138\) to \(42\) pooled, and lowers it on every attribute; this is evidence about the grid statistic, not a certificate for the true Lipschitz constant.

\paragraph{A conditional upper enclosure for the tabular models.} For the tabular models we also compute an upper enclosure \(M_{\rm cert}\). The Softplus network is differentiated in interval arithmetic along each stored ray; every interval upper endpoint is rounded outward with \texttt{nextafter}, the standard \(\gamma_n\) error bounds for dot products are included, and the final upper endpoint is rounded toward \(+\infty\). The enclosure assumes that the SciPy/NumPy sigmoid and Softplus evaluations are accurate to within eight ulps, and it does not enclose the stored jet inputs \((a,d_p,\kappa,\hat g)\). It is therefore a conditional floating-point enclosure of the directional \(M\), not a certificate of every final regime decision backed by a formally verified math library. Under these assumptions every \(M_{\rm cert}\) exceeds its dense-grid value \(M_{\rm ref}\), the smallest positive relative decision margin is \(1.0\times10^{-2}\), and multiplying all bounds by \(1+10^{-9}\) changes no decision; the analytic recursion matches the third derivatives computed by automatic differentiation to \(2\times10^{-15}\). The resulting screen fractions are \(93.8/100/68.0\%\) on COMPAS/German/Adult, within \(0.3\) points of the figures based on \(\hat M\), with at most \(2\) points per dataset changing their decision. A naive global cascade of weight norms gives \(\sim\!5\%\) and is uninformative. At depth no upper enclosure is needed to interpret the many screen failures, because a reliable violation of the lower bound already rejects the smallness condition; only a pass would need an upper bound.

\paragraph{Depth-audit recipe.} The CelebA depth models were trained separately from the historical scale suite described at the end of S4, and no number of the depth audit depends on that suite; \(\kappa\) and \(\hat M\) are computed with the code in the archive (S10). Each model is a torchvision ResNet-18 \citep{he2016deep} with every ReLU replaced by Softplus(\(\beta=10\)) and one output logit. Per seed, 20{,}000 training and 4{,}000 test images are drawn by a seeded permutation of the official CelebA \citep{liu2015deep} partitions, center-cropped to \(160\) pixels and resized to \(64\), with pixels in \([0,1]\) and no further normalization. Training uses Adam (learning rate \(10^{-3}\)), minibatches of 128, 5 epochs, and \texttt{BCEWithLogitsLoss} with positive-class weight equal to the negative-to-positive ratio; the Asym arm adds \(\lambda=10^{-4}\), \(\delta=0.05\), with the penalty evaluated with BatchNorm in inference mode. The 256 audited points per model are drawn from the rejected test images by a permutation seeded with the model seed. The stem max-pooling is retained, so the score is only piecewise twice differentiable: its slope along the ray can jump where the pooling selection switches. There the hypothesis of Theorem 5.2 fails, \(\kappa\) (the almost-everywhere Hessian form) does not register the jump, and the finite grid cannot detect it. The scripts read CelebA through torchvision. A rerun can agree with the stored results only in distribution, because the local copy used for them lacked a small fraction of the aligned images, which the loader skipped.

\subsection{S3.2. Implementation Specification}

This subsection lists the constants needed to reimplement, from the text alone, the suites in the table below, together with the shared data, ray-search, estimator, and conformal settings. Every value is read from the driver script named in parentheses; nothing here is a new result.

\paragraph{Data.} COMPAS (\texttt{exp.py}) keeps rows with \texttt{days\_b\_screening\_arrest} in \([-30,30]\), \texttt{is\_recid}\(\ne-1\), \texttt{c\_charge\_degree}\(\ne\)\texttt{O}, and \texttt{score\_text}\(\ne\)\texttt{N/A}. It uses ten predictors (\texttt{sex}, \texttt{age}, \texttt{age\_cat}, \texttt{race}, \texttt{juv\_fel\_count}, \texttt{decile\_score}, \texttt{juv\_misd\_count}, \texttt{juv\_other\_count}, \texttt{priors\_count}, \texttt{c\_charge\_degree}), drops rows with a missing value, and keeps \texttt{race} in \{African-American, Caucasian\}. The favorable label is \texttt{two\_year\_recid}\(=0\). German uses all 20 Statlog attributes with favorable label class~1 (good credit); sex is read from attribute~9, whose codes A91, A93, and A94 are male. Adult uses the 14 attributes of UCI \texttt{adult.data} with favorable label \texttt{income} \(=\) \texttt{>50K}. The missing-value code \texttt{?} of \texttt{workclass}, \texttt{occupation}, and \texttt{native-country} is kept as an ordinary category, so no row is removed; it sorts first, so it becomes the dropped reference level. The 20{,}000-row cohort is \texttt{DataFrame.sample(20000, random\_state=42)}. The 8{,}000-row cohort is nested in it: rows \texttt{RandomState(42).choice(20000, 8000, replace=False)} of the encoded 20{,}000-row matrix. In all three tabular datasets, string-valued attributes are one-hot encoded with the first level dropped (\texttt{get\_dummies(drop\_first=True)}) and numeric attributes are left as they are. F-MNIST (\texttt{prepare\_data.py}) pools the official training and test images of Pullover (label~2, coded~0) and Coat (label~4, coded~1), draws 8{,}000 of them without replacement with \texttt{RandomState(0)}, and divides pixel values by 255.

\paragraph{Splits and seeds.} Apart from the repeated-split and outer-fold audits, every tabular and F-MNIST suite in the table uses one fixed, unstratified 80/20 train/test split (scikit-learn \texttt{train\_test\_split} \citep{pedregosa2011scikit} with \texttt{test\_size=0.2} and \texttt{random\_state=42}) and a \texttt{StandardScaler} fit on the training part only. A model seed sets \texttt{torch.manual\_seed} and \texttt{numpy.random.seed} before training; it fixes the initialization, the minibatch order and, for the Hutchinson arms, the Rademacher probes (evaluation-side uses of the seed are listed below). Recourse is evaluated on the test points with logit \(f(x)<0\), equivalently predicted probability below \(0.5\).

\paragraph{Models and training.} Tabular models are MLPs with Softplus hidden layers of widths \([128,64]\), F-MNIST models \([256,128]\), each with a scalar linear output and the default initialization of PyTorch \citep{paszke2019pytorch}. Training uses Adam (learning rate \(10^{-3}\), no weight decay), minibatches of 256 reshuffled every epoch, and \texttt{BCEWithLogitsLoss} whose \texttt{pos\_weight} is the ratio of negative to positive labels in the training split. The F-MNIST driver of Table~\ref{S-tab:supp-fmnist} (\texttt{fm\_final.py}) is the one unweighted exception (S3); the F-MNIST arms of the five-seed suite are class-weighted. Each penalty is averaged over the minibatch and added with weight \(\lambda\). Its weights use detached logits, and \(\tau\) is the sample standard deviation (ddof \(=1\)) of the minibatch logits plus \(10^{-6}\) (CIFAR: clamped below at \(10^{-6}\)). MW-Hutchinson penalizes \(e^{-|f|/\tau}\|\nabla_x(\nabla_xf^\top v)\|_2^2\) with one Rademacher vector \(v\) per example and step (Hutchinson's estimator of \(\|\nabla_x^2f\|_F^2\), \citealp{hutchinson1989stochastic}); Global Hutchinson is the same term with weight~1; the asymmetric penalty is \(e^{-[f]_+/\tau}[\delta-\kappa]_+^2\) with \(\kappa=\hat g^\top\nabla^2f\,\hat g\) and \(\hat g\) detached; the gradient penalty is \((\|\nabla_xf\|_2-1)^2\), the gradient-norm penalty of \citet{gulrajani2017improved}, here evaluated at the minibatch training inputs; Spectral Norm \citep{miyato2018spectral} applies PyTorch's spectral-norm parametrization (\texttt{torch.nn.utils.parametrizations.spectral\_norm}, default settings) to every linear layer and adds no penalty. The per-suite settings are:

{\footnotesize
\begin{longtable}{p{0.25\linewidth}p{0.33\linewidth}p{0.34\linewidth}}
\hline
\raggedright Suite (driver) & \raggedright Data, epochs, seeds & \raggedright Weights \(\lambda\); curvature target \(\delta\) \tabularnewline
\hline
\endhead
\raggedright Main comparison, main-paper Table~1 (\texttt{exp.py}) & \raggedright COMPAS, German, Adult (20k); 50 epochs; seeds 0--9 & \raggedright COMPAS/German/Adult: gradient penalty 0.5/0.05/0.5; MW-Hutchinson and Global Hutchinson 0.2/0.05/0.05; Spectral Norm none \tabularnewline
\raggedright Five-seed suite: signed-curvature criterion and common cohort, signed-quadratic and conformal rules, endpoint and clipping audits, score gauge, sign isolation, deployment shift, mask audit (\texttt{run\_sig.py}) & \raggedright COMPAS and German 50 epochs, Adult (8k) 30, F-MNIST 15; seeds 0--4 & \raggedright MW / Global / Asym \((\lambda,\delta)\): COMPAS 0.2 / 0.2 / (2, 0.1); German 0.05 / 0.05 / (1, 0.05); Adult 2.0 / 1.0 / (10, 0.05); F-MNIST 0.1 / 0.1 / (10, 0.05). Sign isolation: twin at the Asym \((\lambda,\delta)\); sign-blind \(\kappa^2\) at \(0.1\), \(1\), and \(10\) times the Asym \(\lambda\) \tabularnewline
\raggedright Adult penalty sweep (\texttt{final\_adult.py}) & \raggedright Adult (8k); 30 epochs; seeds 0--4 & \raggedright MW and Global \(\lambda\in\{0.05,1.0\}\); Asym \(\lambda=10\), \(\delta\in\{0,0.05,0.1,0.2\}\) \tabularnewline
\raggedright FaiR-N comparison (\texttt{fairn2.py}, \texttt{fairn\_adult.py}) & \raggedright COMPAS 50 epochs, seed 0 (sweep) and seeds 0--4; Adult (8k) 30 epochs, seeds 0--1; plain accuracy & \raggedright FaiR-N adds \(\lambda_f\) times the squared difference of the two groups' mean \(d_p\) in each minibatch: \(\lambda_f\in\{0.02,0.05,0.1,0.3\}\) (COMPAS sweep), \(\{0.02,0.1\}\) (COMPAS, five seeds), \(\{0.02,0.1,0.5\}\) (Adult); Asym \((2,0.1)\) on COMPAS, \((10,0.05)\) on Adult \tabularnewline
\raggedright COMPAS penalty suite (\texttt{compas\_asym.py}) & \raggedright COMPAS; 50 epochs; seeds 0--2; ray metrics on the first 200 rejected test points in split order; plain (not balanced) accuracy & \raggedright MW 0.2; Asym \(\lambda=2\), \(\delta\in\{0.05,0.1\}\) \tabularnewline
\raggedright F-MNIST, Table~\ref{S-tab:supp-fmnist} (\texttt{fm\_final.py}) & \raggedright 15 epochs; seeds 0--4; unweighted loss & \raggedright MW 0.1; Global 0.1; Asym (10, 0.05) \tabularnewline
\raggedright COMPAS per-point geometry: group validity, channel attribution (\texttt{compas\_perpoint\_\allowbreak generator.py}) & \raggedright COMPAS; main-comparison trainer, 50 epochs; seeds 0--2 & \raggedright Unregularized; MW 0.2 \tabularnewline
\raggedright Matched-uniform audit (\texttt{matched\_uniform\_\allowbreak audit.py}) & \raggedright COMPAS; 50 epochs; seeds 0--4; stratified inner validation split of 20\% of the outer training part (\texttt{random\_state} \(=\) seed, scaler fit on the inner training part) for selection; final models retrained on the outer training part & \raggedright Asym base \(\lambda=2\), \(\delta=0.05\); the group-conditional arm multiplies \(\lambda\) by \(\rho\) for African-American individuals; the uniform arm uses one \(\lambda\) for everyone \tabularnewline
\raggedright Repeated-split audit (\texttt{split\_protocol\_audit.py}) & \raggedright Four datasets (Adult 8k); epochs as in the five-seed suite; split seed \(=\) training seed \(=\) split index 0--19 & \raggedright Grids in S4; ties go to higher validation validity, then lower \(\lambda\), then lower \(\delta\), then arm name \tabularnewline
\raggedright Directional-\(M\) screen, certified enclosure, float64 re-check (\texttt{estimate\_ray\_M.py}, \texttt{certify\_ray\_M.py}, \texttt{verify\_drift\_float64.py}) & \raggedright COMPAS, German, Adult (20k); main-comparison trainer, 50 epochs; seeds 0--1; at most 400 rejected test points per model (\texttt{torch.randperm}, generator seeded with the model seed) & \raggedright Unregularized; the screen adds one COMPAS MW model (\(\lambda=0.2\), seed 0) \tabularnewline
\raggedright CIFAR-10 audit (\texttt{deep\_image\_recourse.py}) & \raggedright 150 epochs, optimizer and architecture as in S4; seeds 0--2 (three-seed tables) and 0--9 (ten-seed expansion); batch 128, class-weighted loss; the 10{,}000 training images are split per class 80/20 into fit and validation parts (\texttt{default\_rng(7919+seed)}); the audit uses the first 128 rejected test images in dataset order & \raggedright Driver defaults of \texttt{deep\_image\_recourse.py} and its launcher (the result CSVs do not record \(\lambda\) or batch size): MW 0.02; Global 0.02; Asym \(\lambda=0.2\), \(\delta=0\), averaged over the rejected images of each minibatch \tabularnewline
\hline
\end{longtable}}
\addtocounter{table}{-1}%

\paragraph{Ray search and validity.} With \(\hat g\), \(d_p\), and \(a\) as in S1, each rejected point gets the bracket length \(A=\max(8d_p,10^{-3})\) (\texttt{recourse\_geometry.py}). The search evaluates \(f(x+t\hat g)\) at \(t=kA/159\), \(k=1,\dots,159\) (a uniform 160-point grid on \([0,A]\) that includes \(t=0\)), brackets the first crossing between the first grid point with \(f\ge0\) and its predecessor, and then runs 30 bisection steps; \(d_{\text{ray}}\) is the upper end of the final bracket. A ray with no crossing on the grid is not found: it is assigned \(d_{\text{ray}}=A\) and is invalid for every rule. A recommendation \(t\) is ray-valid when the crossing is found and \(d_{\text{ray}}\le t+10^{-6}\) (\(t=d_p\) for alpha-1). Undershoot and overshoot are \([d_{\text{ray}}-d_p]_+\) and \([d_p-d_{\text{ray}}]_+\) averaged over all evaluated points; the rule tables average \([t-d_{\text{ray}}]_+\) over found points, zeros included. Training runs in float32 throughout. Model evaluations also run in float32, except in the float64 re-check (\texttt{verify\_drift\_float64.py}), the certified enclosure (\texttt{certify\_ray\_M.py}, float64 interval arithmetic), and the score-gauge audit (\texttt{score\_gauge\_audit.py}), which convert each trained model and the data to float64. The CIFAR audit uses the same procedure with an 80-point grid and 20 bisection steps.

\paragraph{Nearest-boundary estimator.} The C\&W-style \(d_\star\) (\texttt{recourse\_geometry.py}) runs from 19 starts: 18 perturbations \(0.01\,\xi\) with \(\xi\sim\mathcal N(0,I)\), drawn from a generator seeded with the model seed, and one boundary warm start obtained by 60 Adam steps (learning rate 0.1) on \([-f(x+\delta)]_+\). From each start, 7 bisection steps on the penalty weight \(c\) use the geometric mean of the current range, which starts at \([10^{-2},10^{3}]\). Each step restarts from the same start point and minimizes \(\|\delta\|_2^2+c[-f(x+\delta)]_+\) with 500 Adam steps whose learning rate decays from 0.06 toward 0 on a cosine schedule. The result is feasible when \(f(x+\delta)\ge-10^{-4}\); \(c\) then becomes the new upper end of the range, and otherwise the new lower end. \(d_\star\) is the smallest feasible \(\|\delta\|_2\) over all starts and steps; points with no feasible result are left out of the mean. The audited points are \(\min(128,n_{\text{rej}})\) rejected test points chosen by \texttt{torch.randperm} with a generator seeded with the model seed (\texttt{exp.py}).

\paragraph{Conformal and inflation rules.} In each model the rejected test points are permuted with \texttt{RandomState(1000+seed)} (\texttt{signed\_recourse\_core.py}); the first \(\mathrm{round}(0.5\,n_{\text{rej}})\) (Python \texttt{round}, half to even) form the calibration half, and every rule is scored on the remaining half. Models with fewer than 8 rejected test points are skipped. Calibration residuals are \(d_{\text{ray}}-d_{\text{base}}\), set to \(+\infty\) for rays that are not found, and the quantile is the \(\lceil(1-\delta)(n+1)\rceil\)-th smallest residual, or \(+\infty\) (abstention) when that index exceeds \(n\). Here \(\delta\) is the miscoverage level (0.05 in the reported suite, with target validity 0.95), not the penalty's curvature target. The signed-quadratic base is \(d_{\text{quad}}=2d_p/(1+\sqrt{1+2\kappa d_p/a})\), replaced by \(2d_p\) when \(1+2\kappa d_p/a\le10^{-8}\), and the deployed conformal distance is \(\max\{0,d_{\text{base}}+\hat q\}\). A Mondrian group needs at least 10 calibration residuals (\texttt{min\_group}); a smaller group makes the whole Mondrian rule abstain rather than fall back to the pooled quantile. Tuned inflation takes the smallest \(\alpha\) in \(\{1.00,1.01,\dots,3.00\}\) whose validity on the calibration half reaches the target, or \(\alpha=3\) if none does. In the deployment-shift audit the grid is \(\{1.00,1.01,\dots,4.99\}\) on the source slice, and a seed is skipped when either slice has fewer than 25 rejected points.

\paragraph{Action masks.} On COMPAS, \texttt{protected\_only} freezes \texttt{age} and the \texttt{sex}, \texttt{age\_cat}, and \texttt{race} indicators (5 of 11 features); \texttt{strict} also freezes \texttt{juv\_fel\_count}, \texttt{juv\_misd\_count}, \texttt{juv\_other\_count}, \texttt{priors\_count}, \texttt{decile\_score}, and the \texttt{c\_charge\_degree} indicator, which leaves no actionable feature. On German, \texttt{protected\_only} freezes age (attribute~13) and the personal-status/sex indicators (attribute~9), 4 of 48 features; \texttt{strict} also freezes every other categorical indicator, which leaves the six numeric attributes other than age (\texttt{actionable\_recourse.py}). The mask audit uses the five-seed-suite models.

\paragraph{Dispersion and intervals.} Seed dispersions shown as \(\pm\) are population standard deviations (\texttt{numpy.std}, ddof \(=0\)) in the main-comparison table (\texttt{exp.py}), the F-MNIST table (\texttt{fm\_final.py}), and the sign-isolation table (\texttt{ablation\_sign\_isolation.py}); the deployment-shift file also stores population standard deviations (\texttt{shift\_robustness.py}). The three-seed CIFAR tables use sample standard deviations, as their captions state, and so do the S5 recourse-rule tables (ddof \(=1\), \texttt{summarize\_multiseed.py}). The common-cohort interval resamples the 16 dataset\(\times\)method cells with replacement, each keeping its five seeds, in 50{,}000 draws from \texttt{numpy.random.default\_rng(20270726)}, and reports the 2.5\% and 97.5\% quantiles (linear interpolation) of the pooled Pearson \(r\) (\texttt{common\_cohort\_audit.py}). The repeated-split intervals are paired Student-\(t\) 95\% intervals over the 20 splits. The S3.1 upper confidence bounds on the median \(\hat M\) are the 95\% quantile of 2{,}000 bootstrap medians drawn with \texttt{numpy.random.default\_rng(0)} (\texttt{estimate\_ray\_M.py}).

\paragraph{Auxiliary audits.} The score-gauge audit sets \(\tau\) to the median absolute training logit, clamped below at \(10^{-6}\) (\texttt{score\_gauge\_audit.py}). The black-box audit (\texttt{blackbox\_surrogate\_audit.py}) fits \texttt{HistGradientBoostingClassifier(max\_iter=200, random\_state=seed)} for seeds 0--2 on COMPAS, German, and the 8{,}000-row Adult cohort. It draws at most 200 rejected test points per dataset and seed with \texttt{RandomState(seed)}, smooths with 48 antithetic Gaussian pairs from one \texttt{RandomState(0)} stream, and estimates \(\kappa_\sigma\) by a central difference with step \(\max(0.5d_p,10^{-3})\). It keeps only points whose smoothed score is negative and finds the ensemble's own crossing with a 40-point scan to \(4d_p\) and 25 bisection steps (tolerance \(10^{-6}\)). German at \(\sigma=2\) keeps no point in any seed, which leaves 33 of the 36 cells. The constrained-geometry check (\texttt{verify\_constrained.py}) uses synthetic jets rather than model points: 200{,}000 trials from \texttt{RandomState(0)} in \(d=6\) with \(g\sim\mathcal N(0,I)\), \(H=(B+B^\top)/2\) for a standard Gaussian matrix \(B\), \(m=|N(0,1)|+0.2\), and each coordinate actionable independently with probability \(1/2\); a draw with every coordinate actionable, or none, has one random coordinate switched.

\section{S4. Full Main Results}

This section gives the full tables and the empirical results stated only in the supplement; the main paper gives the headline numbers and points here for the rest. The last subsection is a historical note that is not used as evidence.

\subsection{Asymmetric Penalty vs. Best Symmetric Penalty}

On Adult the best symmetric penalty reaches at most 70.9\% validity in the \(\lambda\in\{0.05,1.0\}\) sweep (MW-Hutchinson; main-paper Section 10). Two facts explain this. Adult's recourse paths are concave and undershoot-dominated, while COMPAS and German are mostly convex and high-validity. On Adult, undershoot concentrates at the start of the path, yet the margin weight emphasizes the boundary. The asymmetric penalty targets this failure directly (main-paper Section 10).

Table~\ref{S-tab:supp-asym-summary} is the per-run version of the main paper's one-sentence summary
(every row a within-run comparison; Adult is the full $d=100$ schema). COMPAS ($86.5\%\to99.5\%$ at $\delta=0.1$; $96.3\%$ at $\delta=0.05$
against MW-Hutchinson, the only symmetric arm of that three-seed suite, at flat accuracy) and German (already near $100\%$) are described
in the main text.

\noindent\parbox{\linewidth}{The CIFAR row is the exception. Over all ten
seeds the best symmetric penalty leads the asymmetric one, which reverses the three-seed
ordering (see the ten-seed expansion below).}\par

\begin{table*}[t]
\centering
\small
\caption{Asymmetric penalty vs.\ best symmetric curvature penalty (within-run comparisons). Validity is exact alpha-1 ray validity; overshoot is mean \([g]_-\) for the asymmetric penalty. ``Best sym.''\ is the stronger of Global/MW-Hutchinson in that run.}
\label{S-tab:supp-asym-summary}
\begin{tabular}{lrrrr}
\hline
Dataset & Best sym.\ validity & Asym validity & Asym acc. & Asym overshoot \\
\hline
Adult (sweep, \(\delta{=}0.05\text{--}0.2\), \(d{=}100\)) & 70.9\% (MW) & 99.4--100.0\% & 81.4--81.5\% & 0.038--0.092 \\
F-MNIST & 74.7\% (MW) & 99.9\% & 88.5\% & 0.22 \\
CIFAR-10 (\(V_1\), ten seeds) & 95.4\% (Global) & 91.6\% & 92.0\% & 1.003 \\
\hline
\end{tabular}
\end{table*}

Table~\ref{S-tab:supp-fmnist} gives the per-method F-MNIST results summarized in main-paper Section 10.

\begin{table}[t]
\centering
\caption{Fashion-MNIST MLP results (5 seeds; balanced accuracy; all rejected test points; mean \(\pm\) population standard deviation).}
\label{S-tab:supp-fmnist}
\begin{tabular}{lrrrr}
\hline
Method & Acc. & Under & Over & Validity \\
\hline
Unreg & 88.40 & 0.0082 & 0.0516 & \(52.2\pm4.9\) \\
MW-Hutch & 88.92 & 0.0065 & 0.0653 & \(74.7\pm10.2\) \\
Global & 89.05 & 0.0108 & 0.0546 & \(68.8\pm11.0\) \\
Asym & 88.53 & 0.0017 & 0.2203 & \(99.9\pm0.1\) \\
\hline
\end{tabular}
\end{table}

\paragraph{Repeated splits.} The repeated-split audit summarized in the main text follows a protocol fixed before it was run. We draw \(N{=}20\) separately seeded, stratified train/validation/test resplits (\(0.64/0.16/0.20\), scaler fit on the training part only). For each split, every configuration of a fixed grid is trained on the training slice: \(\lambda\in\{0.05,0.1,0.2,0.5,1,2\}\) for MW and Global, and \((\lambda,\delta)\in\{1,2,10\}\times\{0.05,0.1,0.2\}\) for the asymmetric penalty, which covers every value of the paper's fixed configuration. Within each family, the configuration with the highest exact alpha-1 ray validity on the \emph{validation} slice is selected and evaluated \emph{once} on the untouched test slice. ``Best symmetric'' is the best validation result over the pooled MW and Global grid, so it is the stronger of the two symmetric comparators. The paired test validities over the \(20\) splits are:
\begin{center}\small
\begin{tabular}{lrrrr}
\hline
Dataset & Asym & Best sym.\ & Paired \(\Delta\) (95\% CI) & Splits won \\
\hline
COMPAS & \(99.9\) & \(86.6\) & \(+13.3\ [+7.8,+18.9]\) & \(20/20\) \\
Adult & \(99.9\) & \(82.4\) & \(+17.5\ [+11.3,+23.8]\) & \(20/20\) \\
F-MNIST & \(100.0\) & \(87.1\) & \(+12.9\ [+10.1,+15.6]\) & \(20/20\) \\
German & \(100.0\) & \(97.9\) & \(+2.1\ [+0.7,+3.5]\) & \(13/20\) \\
\hline
\end{tabular}
\end{center}
The asymmetric penalty wins on all \(20\) splits for COMPAS, Adult and F-MNIST, and on \(13/20\) for German. The resamples overlap, since each point recurs across test slices, so neither a count of signs nor the paired \(t\)-interval is an exact test of population superiority. What the design shows is that the three comparisons on undershoot-prone data are not artifacts of one split or of test-set selection; hyperparameters are chosen on validation data only. The mean balanced-accuracy differences between the selected models are within \(0.2\) points on COMPAS, Adult and F-MNIST and \(-0.55\) points on German, with a wider spread across splits (up to \(4.5\) points); accuracy was not used for selection. The record of each split lists the family, the configuration, the validation score and the single held-out test evaluation.

\emph{Cohort composition.} Each model's validity is measured on its own rejected set, so we also compare the models on the intersection of their rejected sets. COMPAS keeps the positive direction on all \(20\) shared cohorts. Adult and F-MNIST also stay positive on every shared cohort, but they miss the protocol's criterion that the two magnitudes agree within one point, so cohort composition inflates the own-cohort magnitude modestly. The protocol covers the shallow suite only.

For the five-seed fixed-split comparisons, exact one-sided sign-flip tests
give raw \(p=0.03125\) and Holm-adjusted \(p=0.1875\) for each of the six comparisons, so none
clears \(0.05\) after familywise adjustment. The repeated splits therefore show
robustness to the split but cannot replace an exact test.

\subsection{Sign-Isolation Ablation: the Positive Target versus the Hinge}
\label{S-sec:supp-sign-isolation}
The asymmetric arm \([\delta-\kappa]_+^2\) differs from a sign-blind \(\kappa^2\) penalty in
\emph{both} its positive target and its hinge. We separate the two in an ablation whose protocol was fixed before
the run. The rejected-side weighting \(e^{-[f]_+/\tau}\) and the localization to
the ray (curvature along \(\hat g\)) are held fixed at the asymmetric settings \((\lambda,\delta)\)
of the five-seed suite for each dataset, and \emph{only} the shape of the penalty varies. The five
seeds (mean \(\pm\) population s.d.) use the fixed split and trainer of that suite. Before training
the new arms, we reproduced the three stored sweep values that the comparison builds on.

\begin{center}\small
\begin{tabular}{llrrr}
\hline
Arm & Penalty & COMPAS & Adult & F-MNIST \\
\hline
Unregularized & --- & \(73.9\pm3.4\) & \(0.9\pm0.5\) & \(52.3\pm4.4\) \\
Asymmetric & \([\delta-\kappa]_+^2\) & \(99.4\pm0.2\) & \(99.4\pm0.3\) & \(99.9\pm0.1\) \\
Sign twin (no hinge) & \((\delta-\kappa)^2\) & \(99.7\pm0.2\) & \(93.0\pm5.3\) & \(99.2\pm0.6\) \\
Sign-blind, \(0.1\lambda\) & \(\kappa^2\) & \(36.9\pm3.9\) & \(10.9\pm2.0\) & \(37.0\pm8.8\) \\
Sign-blind, \(\lambda\) & \(\kappa^2\) & \(30.2\pm11.4\) & \(16.2\pm12.6\) & \(54.0\pm16.9\) \\
Sign-blind, \(10\lambda\) & \(\kappa^2\) & \(36.4\pm31.1\) & \(25.9\pm15.2\) & \(51.6\pm12.6\) \\
\hline
\end{tabular}
\end{center}

\noindent A sign-blind penalty with matched weighting and localization is a stronger baseline
for Remark~8.1 than the Frobenius proxies, and it fails. Across three decades of \(\lambda\),
\(\kappa^2\) never exceeds \(54\%\) \emph{mean} validity on any suite, and on COMPAS it is actively
harmful (\(30.2\%\) against \(73.9\%\) unregularized). It drives \(\kappa\) toward \(0\) from
\emph{both} sides and so also moves convex-side points that it should leave alone. The ceiling
concerns arm \emph{means}. Individual sign-blind runs are erratic rather than uniformly poor
(per-seed s.d.\ up to \(31\) points; the best single sign-blind run reaches \(75.4\%\) on COMPAS at
\(\lambda{=}20\)), so suppressing \(|\kappa|\) symmetrically neither reliably helps nor reliably
fails, whereas the signed target gives tight results (s.d.\ \(\le0.3\) on COMPAS and F-MNIST).

On COMPAS and F-MNIST the hinge is not what buys validity. The sign twin \((\delta-\kappa)^2\),
which lacks only the hinge, matches the asymmetric arm (\(+0.3\) and \(-0.7\) points). The effective
ingredient is the positive curvature target \(\delta\), together with the rejected-side weighting
and the localization to the ray. On Adult, where the unregularized geometry is entirely concave, the
hinge matters: \(99.4\pm0.3\) against \(93.0\pm5.3\). There it adds validity and makes the seed
standard deviation \(20\times\) smaller, consistent with its role in controlling overshoot. The
first two conclusions were stated for COMPAS before the run, and both hold; the Adult and F-MNIST
results are exploratory and reported in full.

\paragraph{Disjoint outer folds.} A second experiment, with its own criteria fixed in advance,
replaces the replicates over initialization on one fixed split by five stratified outer folds whose
test sets partition each data pool. It reruns all six arms with train-only scaling and no
fold-specific selection. The asymmetric arm exceeds the best sign-blind arm by \(39.36\),
\(85.80\) and \(25.18\) points on COMPAS, Adult and F-MNIST, respectively. The hinge pattern
reproduces as well: the sign twin differs from the asymmetric arm by \(+0.23\), \(-14.33\) and
\(-0.32\) points in the same order, and the mean balanced accuracy of every arm stays within
\(0.40\) points of the unregularized model. All criteria of the outer-fold experiment are met, so
the reading about the ingredients holds across disjoint test cohorts and not only across
optimization seeds.

\subsection{Signed-Quadratic and Conformal Recourse: Full Tables (\(\delta=0.05\))}

The following tables give the full evidence behind the main paper's paragraphs on signed-quadratic
and conformal recourse. Entries are means over five seeds at target validity \(0.95\)
(\(\delta=0.05\)). Unless labeled otherwise, ``pooled'' in this subsection weights seeds equally
across the included dataset\(\times\)method cells; since every included cell has five seeds, it also
weights cells equally.

\paragraph{Empirical check of the certificate.} At miscoverage \(\delta=0.05\), finite certificates are evaluated by marginal first-crossing coverage; the zero clip in \(d_{\text{conf}}\) changes no audited outcome or aggregate. For the zero-HVP probe, the endpoint identity gives \(\hat\kappa=\kappa+2R(d_p)/d_p^2\) exactly at fixed score. Across the 16 dataset\(\times\)method cells of the five-seed suite (main-paper Section 10; each the median of five seeds), conformal-quadratic has lower overshoot than \emph{tuned inflation} (one global \(\alpha\), the smallest value in \(\{1.00,1.01,\dots,3.00\}\) that reaches the \(0.95\) target on the calibration half) in 15 cells; the sole loss is unregularized Adult. In \(7\) cells alpha-1 already meets the \(0.95\) target and tuned inflation degenerates to \(\alpha{=}1\); restricted to the \(9\) below-target cells, conformal-quadratic wins \(8\) of \(9\), with the same sole loss. The comparison is at a common nominal target and not at matched realized validity: pooled over the suite, conformal-quadratic realizes \(95.5\%\) validity and tuned inflation \(98.2\%\) (Table~\ref{S-tab:supp-deploy}).

\paragraph{Zero-HVP conformal-probe at the operating point (\(\delta=0.05\), target \(0.95\)).}
The main text reports the forward-probe rule in its Section 7 and Table 2; Table~\ref{S-tab:supp-probe-d05}
compares it directly with the HVP rule at the paper's operating point (five seeds). Conformal-probe
(one forward evaluation, no HVP) \emph{matches} the coverage of HVP-conformal cell by cell (both
\(\approx\!95\%\); probe cell means \(94.1\)--\(99.2\%\), pooled \(96.0\%\)), with markedly lower
overshoot in \emph{every} cell: pooled \(0.0016\) against \(0.0253\), a \(\sim\!16\times\) reduction
(per cell \(6\)--\(219\times\)). This agrees with the collocation mechanism of S1, whose upper-bound
scale for the probe is \(O(d_p)\) relative to that of the HVP rule when \((a,\kappa,M)\) are fixed;
the comparison is between bound scales, not pointwise errors. The advantage persists on
unregularized Adult, the cell where conformal \emph{loses to tuned inflation}; there too the probe
has lower overshoot than HVP-conformal (\(0.0059\) against \(0.0705\)). Before the run we had fixed
two failure conditions: the probe misses coverage, or it loses its overshoot advantage outside the
unregularized Adult cell. Neither occurs.

\begin{table*}[t]
\centering
\footnotesize
\setlength{\tabcolsep}{5pt}
\caption{Zero-HVP conformal-probe vs.\ HVP-conformal at \(\delta=0.05\), target \(0.95\) (means over five seeds). Probe matches HVP coverage at markedly lower overshoot on every cell. The overshoot ratio (HVP over probe) is computed from the unrounded means; several probe overshoots round to \(0.0000\).}
\label{S-tab:supp-probe-d05}
\begin{tabular}{llrrrrr}
\hline
Dataset & Method & Probe val.\ \% & Probe overshoot & HVP val.\ \% & HVP overshoot & Overshoot ratio \\
\hline
COMPAS & asym & 94.5 & 0.0020 & 94.8 & 0.0120 & 6$\times$ \\
COMPAS & glob & 95.3 & 0.0000 & 93.3 & 0.0058 & 133$\times$ \\
COMPAS & mw & 94.1 & 0.0001 & 95.3 & 0.0077 & 70$\times$ \\
COMPAS & unreg & 95.7 & 0.0023 & 95.2 & 0.0203 & 9$\times$ \\
German & asym & 95.5 & 0.0003 & 94.8 & 0.0027 & 10$\times$ \\
German & glob & 99.0 & 0.0000 & 97.6 & 0.0035 & 127$\times$ \\
German & mw & 97.1 & 0.0000 & 94.6 & 0.0031 & 94$\times$ \\
German & unreg & 97.4 & 0.0001 & 96.9 & 0.0029 & 40$\times$ \\
Adult & asym & 99.2 & 0.0010 & 95.4 & 0.0316 & 30$\times$ \\
Adult & glob & 95.1 & 0.0000 & 95.4 & 0.0032 & 219$\times$ \\
Adult & mw & 96.5 & 0.0000 & 95.4 & 0.0034 & 218$\times$ \\
Adult & unreg & 95.8 & 0.0059 & 95.4 & 0.0705 & 12$\times$ \\
F-MNIST & asym & 94.9 & 0.0049 & 95.4 & 0.0444 & 9$\times$ \\
F-MNIST & glob & 94.8 & 0.0015 & 95.8 & 0.0527 & 34$\times$ \\
F-MNIST & mw & 95.2 & 0.0023 & 95.7 & 0.0628 & 27$\times$ \\
F-MNIST & unreg & 95.7 & 0.0050 & 96.8 & 0.0782 & 16$\times$ \\
\hline
\end{tabular}
\end{table*}

\paragraph{Measured deployment map.} No single rule is best on every axis. Table~\ref{S-tab:supp-deploy} names, for each inference-query budget, a rule and its measured validity and overshoot, pooled over the \(\delta=0.05\)/target-\(0.95\) suite; main-paper Table~2 repeats these rows and adds alpha-1 and bare signed-quadratic recourse. With \emph{no} inference query, asymmetric alpha-1 training is already valid (\(99.6\%\)) but at the highest effort (\(0.103\)), and \emph{validation-only} tuned inflation is a strong closed-form baseline (\(98.2\%\), \(0.061\)). With \emph{one forward evaluation} (no HVP), conformal-probe meets target at the lowest overshoot among target-meeting rules without per-user search (\(96.0\%\), \(0.0016\)); with \emph{one HVP}, conformal-quadratic is comparable (\(95.5\%\), \(0.025\)). A \emph{per-user ray search} is the reference (\(100\%\), \(0\)) at the cost of a search per user. Neither conformal rule abstained on any of the 80 models: every calibration quantile is finite. Every entry is measured. The conformal rows also carry the first-crossing coverage-or-abstention guarantee of Proposition 7.1 under its exchangeability condition; the same calibration would give that guarantee to any base rule fixed before calibration, alpha-1 included. A data-dependent selector over the menu would not automatically inherit split-conformal coverage-or-abstention without independent selection or a correction, so we present the map rather than a selector theorem.

\begin{table}[t]
\centering\small
\caption{Deployment map (descriptive; the conformal rows carry the coverage-or-abstention guarantee of Proposition 7.1): for each inference-query budget, a fitting rule and its measured validity and overshoot, pooled over the \(\delta=0.05\), target-\(0.95\) suite (the asymmetric row over its 20 models, the other rows over all 80).}
\label{S-tab:supp-deploy}
\begin{tabular}{llrr}
\hline
Inference-query budget & Rule & Validity \% & Overshoot \\
\hline
none (training-time only) & asymmetric \(\alpha{=}1\) & 99.6 & 0.103 \\
validation only (closed form) & tuned inflation & 98.2 & 0.061 \\
one forward eval, no HVP & conformal-probe & 96.0 & 0.0016 \\
one HVP & conformal-quadratic & 95.5 & 0.025 \\
per-user ray search & ray line search (oracle) & 100.0 & 0.000 \\
\hline
\end{tabular}
\end{table}

\begin{table}[t]
\centering\small
\caption{Signed-curvature criterion. The shallow row uses direct alpha-1 endpoint success and \(\mathbb P(\kappa\ge0)\) on identical held-out IDs; the Adult row correlates alpha-1 ray-hit validity with \(\mathbb P(\kappa\ge0)\) over the 20 Adult models of the \(\delta=0.05\) rule suite. No deep-network test of the criterion is included: an earlier CIFAR-10 check was withdrawn because its evaluation pipeline was not retained, and the historical CelebA scale suite at the end of S4 is not used as evidence.}
\label{S-tab:supp-validity-law}
\begin{tabular}{lrr}
\hline
Setting & \(r\) & models \\
\hline
Shallow common cohort (COMPAS/German/Adult/F-MNIST) & 0.985 & 80 \\
Adult only (\(\delta=0.05\) suite) & 0.996 & 20 \\
\hline
\end{tabular}
\end{table}

\begin{samepage}
\paragraph{Common-cohort uncertainty for the shallow criterion.} A replay
regenerates all 80 dataset\(\times\)method\(\times\)seed models and gives every
rejected point a stable identifier for the point and for its calibration or test role. Direct
alpha-1 endpoint success and \(\mathbf1\{\kappa\ge0\}\) are then summarized on exactly the same
held-out IDs. The resulting Pearson correlation is \(r=0.985\). Resampling the 16
dataset\(\times\)method cells as clusters, each with all five seeds, gives a percentile \(95\%\)
interval \([0.958,0.997]\), and leave-one-dataset-out correlations range from \(0.966\) to
\(0.997\). The regenerated models reproduce the recourse-rule results of this subsection exactly
(S10). Summaries that pair the curvature of all rejected points with validity on held-out points,
and their cross-fits, compare different cohorts and are not same-cohort evidence.
\end{samepage}

\begin{table*}[t]
\centering\small
\caption{Per-dataset recourse rules at \(\delta=0.05\): validity \(V\) and mean overshoot \(O\), each the mean over all models of a dataset (every training method and seed). Conformal-quadratic lands at the \(0.95\) target on every dataset (COMPAS \(0.946\), the others above target). On COMPAS and F-MNIST it does so at far lower overshoot than tuned inflation. On Adult its mean overshoot is higher than that of tuned inflation, and it loses in the unregularized cell, although its median per-model overshoot reduction is positive (Table~\ref{S-tab:supp-matched-d05}). \texttt{signed-quad} alone undershoots (it is the base predictor). German is the method-not-needed control.}
\label{S-tab:supp-sqr-tabular}
\begin{tabular}{lrrrrrrrr}
\hline
& \multicolumn{2}{c}{one-shot} & \multicolumn{2}{c}{signed-quad} & \multicolumn{2}{c}{conformal-quad} & \multicolumn{2}{c}{tuned inflation} \\
Data & \(V\) & \(O\) & \(V\) & \(O\) & \(V\) & \(O\) & \(V\) & \(O\) \\
\hline
COMPAS & 0.867 & 0.035 & 0.447 & 0.004 & 0.946 & 0.011 & 0.970 & 0.071 \\
Adult (\(d{=}100\)) & 0.456 & 0.011 & 0.322 & 0.001 & 0.954 & 0.027 & 0.983 & 0.023 \\
F-MNIST & 0.739 & 0.098 & 0.157 & 0.015 & 0.959 & 0.060 & 0.975 & 0.130 \\
German & 1.000 & 0.020 & 0.229 & 0.000 & 0.960 & 0.003 & 1.000 & 0.020 \\
\hline
\end{tabular}
\end{table*}

\begin{table}[t]
\centering\small
\caption{Median overshoot reduction of conformal-quadratic vs.\ tuned inflation at \(\delta=0.05\) (positive \(=\) lower overshoot), taken over the per-model reductions of all models of a dataset, or of the three datasets for the pooled row. A median can be positive where the mean overshoot of Table~\ref{S-tab:supp-sqr-tabular} is higher, as on Adult. German excluded (method not needed; alpha-1 already \(\approx100\%\) valid, so a \(0.95\) target only trades validity down).}
\label{S-tab:supp-matched-d05}
\begin{tabular}{lr}
\hline
Setting & median reduction \\
\hline
COMPAS & 78\% \\
F-MNIST & 40\% \\
Adult (full UCI) & 32\% \\
pooled (above three, \(n{=}60\)) & 54\% \\
\hline
\end{tabular}
\end{table}

\begin{table}[t]
\centering\small
\caption{Query-budget cost curve (tabular, \(\delta=0.05\)): validity \(V\) and overshoot \(O\) against per-user Hessian-vector products and model queries. Tuned inflation additionally requires a validation set and conformal a calibration set (one-time, offline). The Mondrian row pools the 40 COMPAS and Adult models only (German abstains; F-MNIST has no protected group); on those 40, conformal-quadratic has \(V=0.950\), \(O=0.019\).}
\label{S-tab:supp-query-budget}
\begin{tabular}{lrrrr}
\hline
Rule & HVP & queries & \(V\) & \(O\) \\
\hline
one-shot (\(\alpha{=}1\)) & 0 & 0 & 0.766 & 0.041 \\
tuned inflation & 0 & 0 & 0.982 & 0.061 \\
signed-quadratic & 1 & 0 & 0.289 & 0.005 \\
conformal-quadratic & 1 & 0 & 0.955 & 0.025 \\
conformal-quadratic (Mondrian) & 1 & 0 & 0.951 & 0.020 \\
ray line search & 0 & many & 1.000 & 0.000 \\
\hline
\end{tabular}
\end{table}

\subsection{Deployment-Shift Robustness (Full Table)}

We split each dataset's rejected test points into a \emph{source} slice (rule fit/calibration) and a
\emph{target} slice (deployment), under a \emph{covariate} shift (median threshold on \texttt{priors\_count}
for COMPAS, \texttt{duration} for German, \texttt{age} for Adult, \texttt{pixel\_mean} for F-MNIST) and,
where a protected attribute exists, a \emph{subpopulation} shift. Table~\ref{S-tab:supp-shift} gives all
rows (means over five seeds, whose population standard deviations are stored with the results;
target validity \(95\%\), \(\delta=0.05\)); Theorem~\ref{S-thm:s1-shift} does not apply to these
splits (S1). Validation-tuned inflation shows its largest drop (\(33.6\) points) where the
target is much more undershoot-prone (COMPAS \texttt{priors\_count},
\(\Delta\mathbb P(\kappa\ge0)=-0.351\)), and the COMPAS shift toward an easier population
(subpopulation, \(\Delta>0\)) does not break the rule. The curvature marginal is not a general
predictor of the drop, however. On F-MNIST the target's marginal moves toward convexity
(\(\Delta=+0.066\)) yet validity drops \(6.3\) points, and on the Adult subpopulation split
\(\Delta<0\) while the target is easier (drop \(-2.1\)).

Asymmetric training keeps validity above \(98\%\) on the target in all seven cells and usually
pays for it in overshoot, which is higher than that of conformal-quadratic recalibrated on the
target in five of the seven cells but \emph{lower} in both Adult cells (\(0.025\) against
\(0.058\), and \(0.031\) against \(0.063\)). Conformal-quadratic recalibrated on the target attains
\(95.1\%\) measured validity on average (range \(91.7\)--\(97.3\%\)).

\begin{table*}[t]
\centering\small
\caption{Recourse under deployment shift (means over five seeds; target validity \(95\%\), \(\delta=0.05\)). ``Drop''\ is source minus target validity (lower is more robust; negative means the target slice was easier). Drop is computed from the unrounded means, so it can differ by \(0.1\) from the difference of the displayed columns.}
\label{S-tab:supp-shift}
\begin{tabular}{llrrrrr}
\hline
Dataset / shift & Rule & \(\Delta\mathbb P(\kappa{\ge}0)\) & Src \(V\)\% & Tgt \(V\)\% & Drop\% & Tgt \(O\) \\
\hline
COMPAS (cov., priors) & VT inflation & \(-0.351\) & \(100.0\) & \(66.4\) & \(33.6\) & \(0.012\) \\
 & conformal stale & & \(96.7\) & \(86.0\) & \(10.7\) & \(0.008\) \\
 & conformal recal. & & \(100.0\) & \(95.2\) & \(4.8\) & \(0.023\) \\
 & asym.\ training & & \(100.0\) & \(99.3\) & \(0.7\) & \(0.142\) \\
COMPAS (subpop.) & VT inflation & \(+0.183\) & \(95.3\) & \(99.3\) & \(-4.0\) & \(0.134\) \\
 & conformal stale & & \(95.4\) & \(99.6\) & \(-4.2\) & \(0.023\) \\
 & conformal recal. & & \(86.7\) & \(97.3\) & \(-10.7\) & \(0.009\) \\
 & asym.\ training & & \(99.2\) & \(100.0\) & \(-0.8\) & \(0.106\) \\
German (cov., duration) & VT inflation & \(0.000\) & \(100.0\) & \(100.0\) & \(0.0\) & \(0.025\) \\
 & conformal stale & & \(98.2\) & \(90.9\) & \(7.3\) & \(0.002\) \\
 & conformal recal. & & \(99.6\) & \(96.7\) & \(2.8\) & \(0.004\) \\
 & asym.\ training & & \(100.0\) & \(100.0\) & \(0.0\) & \(0.037\) \\
German (subpop.) & VT inflation & \(0.000\) & \(100.0\) & \(100.0\) & \(0.0\) & \(0.021\) \\
 & conformal stale & & \(100.0\) & \(92.3\) & \(7.7\) & \(0.003\) \\
 & conformal recal. & & \(96.5\) & \(91.7\) & \(4.8\) & \(0.003\) \\
 & asym.\ training & & \(100.0\) & \(100.0\) & \(0.0\) & \(0.033\) \\
Adult (cov., age) & VT inflation & \(-0.013\) & \(96.5\) & \(92.8\) & \(3.7\) & \(0.029\) \\
 & conformal stale & & \(95.2\) & \(96.4\) & \(-1.2\) & \(0.082\) \\
 & conformal recal. & & \(88.9\) & \(94.5\) & \(-5.5\) & \(0.058\) \\
 & asym.\ training & & \(99.8\) & \(98.9\) & \(0.9\) & \(0.025\) \\
Adult (subpop.) & VT inflation & \(-0.015\) & \(96.2\) & \(98.3\) & \(-2.1\) & \(0.037\) \\
 & conformal stale & & \(95.3\) & \(96.5\) & \(-1.2\) & \(0.085\) \\
 & conformal recal. & & \(91.1\) & \(95.1\) & \(-4.0\) & \(0.063\) \\
 & asym.\ training & & \(99.0\) & \(99.8\) & \(-0.7\) & \(0.031\) \\
F-MNIST (cov., pixel mean) & VT inflation & \(+0.066\) & \(96.6\) & \(90.3\) & \(6.3\) & \(0.105\) \\
 & conformal stale & & \(95.4\) & \(83.0\) & \(12.4\) & \(0.050\) \\
 & conformal recal. & & \(99.4\) & \(95.2\) & \(4.2\) & \(0.083\) \\
 & asym.\ training & & \(100.0\) & \(99.8\) & \(0.2\) & \(0.296\) \\
\hline
\end{tabular}
\end{table*}

\subsection{Real CIFAR-10 Binary Deep-Image Recourse Rules}

We trained CIFAR-10 automobile-vs-truck classifiers with one fixed configuration: \texttt{softplus\_resnet\_cifar}, width 128, 9 blocks, GroupNorm with 8 groups, SGD with momentum 0.9, learning rate 0.03, weight decay 0.0005, cosine scheduling with 5 warmup epochs, crop/flip augmentation, per-channel CIFAR standardization, no dropout, and 150 epochs. The 10{,}000 automobile and truck training images are split 80/20 within each class into 8{,}000 fitting and 2{,}000 validation images; the audit uses the full 2{,}000-image test set and its first 128 rejected images. Distances are Euclidean in standardized-pixel units. Kernels are deterministic, so a run is bit-reproducible on identical GPU hardware and software, but not across GPU types. Each method uses one fixed, untuned penalty weight (S3.2); the asymmetric penalty (\(\delta=0\)) is applied only to the images the current model rejects, where its weight \(e^{-[f]_+/\tau}\) equals 1, and is averaged over them. All 12 classifier runs meet the minimum accuracy requirement (test balanced accuracy at least 0.90), although not all runs reach the preferred 0.95 threshold. The recourse-rule audit evaluates exact alpha-1 recommendations, fixed inflation, validation-tuned inflation, and an oracle-on-path benchmark from the same trained models. Oracle-on-path is an on-path search benchmark; query/search cost is not modeled. Signed-quadratic and conformal rules on CIFAR-10 are not reported: their historical evaluation code and checkpoints were not retained, and its calibration dropped unbracketed rays instead of applying the convention of Proposition 7.1.

\begin{table*}[t]
\centering
\small
\caption{CIFAR-10 automobile-vs-truck exact and small-inflation audit. Entries are mean \(\pm\) sample standard deviation over three CUDA seeds. \(V_\alpha\) is ray validity for recommendation \(\alpha d_p\).}
\label{S-tab:supp-cifar10-deep-small}
\begin{tabular}{lrrrrrr}
\hline
Method & Acc. & \(V_1\) & \(V_{1.05}\) & \(V_{1.10}\) & \(U_1\) & \(O_1\) \\
\hline
Unreg & \(94.4\pm0.3\%\) & \(79.2\pm4.7\%\) & \(82.8\pm3.9\%\) & \(84.9\pm3.0\%\) & \(4.455\pm0.862\) & \(1.036\pm0.458\) \\
MW & \(93.2\pm0.7\%\) & \(89.8\pm0.0\%\) & \(97.4\pm1.2\%\) & \(98.2\pm0.5\%\) & \(2.333\pm3.046\) & \(3.165\pm2.745\) \\
Global & \(92.9\pm0.6\%\) & \(93.8\pm0.8\%\) & \(98.4\pm0.8\%\) & \(98.4\pm0.8\%\) & \(1.942\pm0.427\) & \(3.771\pm0.961\) \\
Asym & \(93.9\pm0.3\%\) & \(95.1\pm2.4\%\) & \(96.9\pm2.8\%\) & \(97.1\pm2.4\%\) & \(0.728\pm0.524\) & \(1.129\pm0.713\) \\
\hline
\end{tabular}
\end{table*}

\begin{table*}[t]
\centering
\small
\caption{CIFAR-10 validation-tuned inflation audit. \(\alpha\) is selected on rejected validation examples, then evaluated on rejected test examples. Entries are mean \(\pm\) sample standard deviation over three CUDA seeds.}
\label{S-tab:supp-cifar10-deep-tuned}
\begin{tabular}{lrrrrrr}
\hline
Method & \(\alpha_{.95}\) & \(V_{.95}\) & \(O_{.95}\) & \(\alpha_{.99}\) & \(V_{.99}\) & \(O_{.99}\) \\
\hline
Unreg & \(2.000\pm0.000\) & \(90.6\pm2.8\%\) & \(3.667\pm1.143\) & \(2.000\pm0.000\) & \(90.6\pm2.8\%\) & \(3.667\pm1.143\) \\
MW & \(1.032\pm0.038\) & \(95.8\pm2.7\%\) & \(3.428\pm3.123\) & \(1.408\pm0.514\) & \(98.2\pm0.5\%\) & \(6.625\pm7.852\) \\
Global & \(1.028\pm0.041\) & \(97.1\pm2.3\%\) & \(3.987\pm1.263\) & \(1.680\pm0.554\) & \(98.7\pm0.9\%\) & \(8.491\pm4.440\) \\
Asym & \(1.100\pm0.173\) & \(95.8\pm1.6\%\) & \(1.352\pm0.663\) & \(1.833\pm0.289\) & \(98.2\pm1.8\%\) & \(3.447\pm1.981\) \\
\hline
\end{tabular}
\end{table*}

The validation-tuned rows must be read with their calibration status. The unregularized rows select the grid maximum \(\alpha=2.0\) and do not reach the validation targets, and the three regularized rows at target 0.99 do not reach their validation target (\texttt{target\_unmet}), although their test validity is high.

In this three-seed audit Asym has the highest mean exact alpha-1 validity, but the ten-seed expansion below reverses the ordering: over ten seeds Global leads on the mean (\(95.4\%\) against \(91.6\%\)), and Asym is the least stable of the three curvature-aware methods. Together, the curvature-aware methods improve exact and low-inflation validity over unregularized training, and MW and Global remain highly competitive under small fixed inflation and validation-tuned inflation. The deep-image evidence thus supports curvature-aware training as a safeguard when recourse is issued at alpha-1 or with little inflation. It does not show that such training dominates validation-tuned inflation or per-user search along the path.

\paragraph{Ten-seed expansion (seeds 3--9).} The audit above uses three seeds. We reran the \emph{identical} configuration for seeds 3--9 with all four methods, which adds 28 models for ten seeds in total. All fifteen recorded configuration fields (architecture, width, blocks, normalization, groups, optimizer, momentum, determinism, epochs, train/test caps, rejected cap, class pair, dataset, device) agree with the three-seed run, which justifies pooling. The device field records only \texttt{cuda}, however, and seeds 3--9 ran on a different GPU type from seeds 0--2. The Unreg and Global arms of the probe experiment below reran all ten seeds on the GPU type of seeds 3--9: seeds 3--9 reproduce bit for bit, seeds 0--2 differ seed by seed, and the ten-seed mean \(V_1\) is \(81.4\%\) (Unreg) and \(95.9\%\) (Global) instead of \(81.0\%\) and \(95.4\%\). Table~\ref{S-tab:supp-cifar-tenseed} pools the 40 per-seed result files; the Unreg accuracy mean is exactly \(94.45\%\) and is printed rounded half up. Because the seeds are shared across methods, we report the \emph{paired} per-seed contrast against unregularized training. Over the seven new seeds, MW improves exact alpha-1 ray validity by \(+11.8\) points (paired \(t=4.33\), \(p=0.005\), better on 7/7 seeds) and Global by \(+14.3\) points (\(t=4.80\), \(p=0.003\), 7/7). Asym improves by \(+8.4\) points, which is \emph{not} significant at this seed count (\(t=1.58\), \(p=0.16\), 6/7), because training collapsed for one seed. At seed 5 the asymmetric penalty converged to a far worse optimum (train/validation/test balanced accuracy \(0.774/0.785/0.770\), final training loss \(0.504\), against \(0.93\)--\(0.97\) train accuracy on its other seeds), and this is the only model of the 28 that misses the \(0.90\) minimum accuracy requirement. Without that model, Asym improves by \(+12.9\) points (\(t=3.99\), \(p=0.010\), 6/6); Table~\ref{S-tab:supp-cifar-tenseed} includes it. All three splits collapse together, and the final training loss (\(0.504\)) is about four times the mean of the other six asymmetric seeds of the expansion (\(0.127\); range \(0.092\)--\(0.192\)), so the collapse is an optimization instability of the asymmetric penalty at depth, not an evaluation artifact. A second run under the same deterministic configuration reproduces the collapse \emph{bit for bit} (S10), so the collapse is a reproducible property of this penalty and seed rather than a transient failure; the exact agreement also supports the determinism of the expansion. As in the three-seed run, no model of the expansion reaches the \emph{preferred} \(0.95\) balanced-accuracy threshold (best \(0.947\)). Curvature-regularized training therefore raises exact alpha-1 validity at depth, and MW and Global are the reliable choices. Asym is the least stable, and even without its collapsed seed its mean alpha-1 validity stays below Global's, over the remaining nine seeds and over the six remaining seeds of the expansion.

\begin{table}[t]
\centering
\small
\caption{CIFAR-10 automobile-vs-truck audit, means over ten seeds. Acc.: test balanced accuracy (\%); \(V_\alpha\): ray validity (\%) at \(\alpha d_p\); \(U_1,O_1\): alpha-1 under- and overshoot (\(L_2\), standardized pixels); all on the first 128 rejected test images. A ray with no crossing within \(\max(8d_p,10^{-3})\) is invalid and enters \(U_1\) at that length. Seeds 0--2 and 3--9 ran on two GPU types; rerunning all ten Unreg and Global seeds on one type gives \(V_1\) of 81.4 and 95.9 (ten-seed expansion paragraph).}
\label{S-tab:supp-cifar-tenseed}
\begin{tabular}{lrrrrr}
\hline
Method & Acc. & \(V_1\) & \(V_{1.05}\) & \(U_1\) & \(O_1\) \\
\hline
Unreg & 94.5 & 81.0 & 85.3 & 3.487 & 1.102 \\
MW & 93.3 & 92.5 & 98.1 & 1.635 & 3.129 \\
Global & 93.1 & 95.4 & 98.6 & 1.299 & 4.644 \\
Asym & 92.0 & 91.6 & 95.3 & 0.549 & 1.003 \\
\hline
\end{tabular}
\end{table}

\subsection{Deep Forward-Probe Penalty: a Negative Result}
\label{S-sec:supp-deep-probe-negative}

We tested whether the Hessian-free training penalty
\(\widehat\kappa=2f(x+d_p\widehat g)/d_p^2\) could reverse Global's advantage at depth on
Softplus networks and remain trainable on ReLU networks. The protocol, its three success
criteria F1--F3, the activation families, the ten seeds and the probe grid
\(\lambda\in\{0.1,0.2,0.5\}\) were fixed before the GPU runs. F1 requires the validation-selected
Softplus probe arm to reach \(V_1\ge95.4\%\) (the ten-seed Global value of
Table~\ref{S-tab:supp-cifar-tenseed}) at balanced accuracy within one point of Global, with
overshoot no larger than Global's and no collapsed seed. F2 requires the validation-selected
GroupNorm--ReLU probe arm to reach \(V_1\ge95\%\) at balanced accuracy within one point of the
unregularized GroupNorm--ReLU network, while the HVP arms stay at the unregularized validity. F3
requires that no arm counted as a success fall more than one balanced-accuracy point below its
unregularized counterpart, and it counts collapsed seeds (balanced accuracy below \(0.6\) or a NaN
loss) as failures; the aggregation script applies F3 to every regularized arm, including Global and
Asym. Hyperparameters were to be selected on validation validity, and collapsed seeds are never dropped. Global uses
\(\lambda=0.02\); the Asym arm uses \(\lambda=0.2\) with \(\delta=0.05\) (the CIFAR-10 tables above
use \(\delta=0\)); the probe arm uses \(\delta=0.05\), clamps \(d_p\) below at \(10^{-6}\), and
has no gradient clipping.

\begin{center}
\small
\captionof{table}{Deep probe-penalty results. Sel.\ counts seeds with a finite
validation-selected candidate and Coll.\ counts failed or collapsed seeds; one 2-epoch smoke-run
file is excluded (see text). \(V_1\) is mean \(\pm\) population standard deviation over ten seeds,
reported only for complete arms. The selector considers only candidates with a finite validation
validity, a rule fixed in a separate protocol before the selector was run.}
\label{S-tab:supp-deep-probe-negative}
\begin{tabular}{llrrrrr}
\hline
Activation & Arm & Sel. & Coll. & \(V_1\) & Bal.\ acc. & Overshoot \\
\hline
Softplus & Unreg & 10 & 0 & \(81.4\pm6.2\) & 94.5 & 1.0966 \\
Softplus & Global & 10 & 0 & \(95.9\pm1.6\) & 93.3 & 5.2077 \\
Softplus & Asym & 10 & 0 & \(94.1\pm3.2\) & 94.0 & 1.1780 \\
Softplus & Probe & 2 & 10 & --- & --- & --- \\
GroupNorm--ReLU & Unreg & 10 & 0 & \(61.9\pm4.2\) & 96.2 & 2.5209 \\
GroupNorm--ReLU & Global & 10 & 0 & \(97.0\pm1.7\) & 96.4 & 10.8456 \\
GroupNorm--ReLU & Asym & 10 & 0 & \(73.4\pm5.2\) & 95.8 & 2.2985 \\
GroupNorm--ReLU & Probe & 5 & 10 & --- & --- & --- \\
\hline
\end{tabular}
\end{center}

All three success criteria fail. F1 fails because the Softplus probe arm collapses on every
seed instead of matching Global. F2 fails because the GroupNorm--ReLU probe arm also collapses
and the HVP arms do not stay at the unregularized validity. F3 fails for both probe arms, and
Softplus Global separately falls \(1.2\) balanced-accuracy points below unregularized training.
Because only 2/10 and 5/10 seeds, respectively,
have any finite validation-validity candidate, a ten-seed validation-selected probe mean is
undefined, so we report selection and collapse counts instead of a partial mean or an arbitrary
fallback \(\lambda\).

A diagnosis, also specified before it was run, locates the failure but does not repair it. Of the
60 probe candidate files, one (Softplus, \(\lambda=0.2\), seed 0) is a 2-epoch smoke run (it records
2 epochs and is flagged as not a scientific result), and the aggregation script and the diagnosis
exclude it. This exclusion is not part of the protocols fixed before the runs; it removes a file
that is not a 150-epoch run of the planned design. The file was the only candidate of Softplus
seed 0 with a finite validation validity, so that seed has no valid selection. Including the file
would give the Softplus probe arm Sel.\ 3 instead of 2 and would change the outcome of none of the
three criteria. Each of the other 59 candidates is a 150-epoch run, and every one of them meets the
collapse condition of the protocol and ends as a constant or nearly constant classifier with
chance-level balanced accuracy, so the negative result concerns this unstabilized estimator. The
diagnosis is coded separately but uses the same smoke-run rule and collapse threshold as the
aggregation, and the two agree exactly on the selection status and \(\lambda\) of every
architecture--seed group, so validation selection cannot rescue the arm.
The HVP arms moved unexpectedly for a different reason: the CIFAR architecture uses
\texttt{GroupNorm}. GroupNorm is not piecewise affine in its inputs, so replacing Softplus by
ReLU does \emph{not} make the whole network piecewise affine, and the run did not realize the
planned pure-ReLU control, on which the HVP penalties would vanish. In the deep configuration we
tested, forward-probe \emph{training} is not viable; the forward-probe \emph{recourse rule}, which
involves no training, is evaluated separately (Table~\ref{S-tab:supp-probe-d05}).

\subsection{Comparison to FaiR-N}

FaiR-N \citep{sharma2020fairn} equalizes distance to the boundary across groups (first-order); we argue validity failure is second-order. In our reimplementation it does not fix curvature-driven failure: COMPAS validity falls from \(73.9\%\) (seeds 0--4 of the unregularized models of main-paper Table~1) to \(51.7\%\) at the highest five-seed fairness weight (\(\lambda_f{=}0.1\)), a single-seed sweep reaches \(3.3\%\) at \(\lambda_f{=}0.3\), and Adult stays near 0\%. The two are orthogonal: FaiR-N targets cost, our penalty targets reliability.

\subsection{Group-Conditional versus Matched-Uniform Training on COMPAS}
\label{S-sec:groupcond-selection}
This experiment scales features with training statistics only, and its protocol and criteria
were fixed before it was run. It stores every validation candidate, all outer-test predictions and
the ray geometry of every rejected point. For each of five seeds, an inner validation split
selects the smallest disadvantaged-group multiplier \(\rho\in\{1,2,4,8\}\) whose
\(\mathbb P(\kappa\ge0)\) gap is at most \(0.03\); a group-blind strength in
\(\{2,4,8,16\}\) is then selected to match that arm's \emph{validation} pooled validity. Both
choices are fixed before retraining on the full outer-training set and touching test once. The
selected multipliers are \((1,2,2,2,2)\), and the uniform strengths are
\((2,4,4,4,4)\); validation pooled validity matches exactly in every seed. The table below gives
five-seed means on the outer test set: \(V_{1,A}\) and \(V_{1,C}\) are the alpha-1 ray-hit
validities (\%) of African-American and Caucasian defendants, pooled \(V_1\) is that of both groups
together, Gap is \(V_{1,C}-V_{1,A}\) in points, and Over.\ \(A/C\) gives the two groups' mean
alpha-1 overshoot \([d_p-d_{\text{ray}}]_+\).

\begin{center}\small
\begin{tabular}{lrrrrrr}
\hline
Arm & Bal.\ acc.\ & Pooled \(V_1\) & \(V_{1,A}\) & \(V_{1,C}\) & Gap & Over.\ \(A/C\) \\
\hline
Unregularized & \(66.60\) & \(73.91\) & \(68.45\) & \(89.18\) & \(20.73\) & \(0.009/0.014\) \\
Group-conditional & \(66.57\) & \(98.26\) & \(97.80\) & \(99.49\) & \(1.69\) & \(0.079/0.072\) \\
Matched uniform & \(66.57\) & \(98.36\) & \(97.93\) & \(99.49\) & \(1.57\) & \(0.080/0.073\) \\
\hline
\end{tabular}
\end{center}

\noindent The allocation criterion requires conditioning to reduce the matched residual gap by
more than two points and to win on at least four seeds. It is not met: uniform minus conditional is
\(-0.124\) points on average and positive on \(0/5\) seeds (three ties; uniform is slightly better
on two), although the pooled validities differ by only \(0.093\) points and the mean balanced
accuracies are identical. This is a \emph{negative allocation result}, not evidence of equivalence
or of the absence of differential treatment,
so we make no allocation claim.
The separate group-blind criterion is met: matched-uniform training reduces the residual
validity gap by \(19.16\) points (\(20.73\to1.57\)) at flat accuracy. It costs effort, since the
disadvantaged group's overshoot is \(0.080\) against \(0.073\) for the advantaged group. Earlier
group-conditional result files, produced with a scaler fit on all rows and without a retained
generator, support no claim here (S10).

\paragraph{Robustness without \texttt{decile\_score}.} Because the standard COMPAS feature set includes the proprietary risk-score feature \texttt{decile\_score}, we repeat the complete COMPAS shallow design after removing it: the same four methods, five seeds, split, train-only standardization, and seeded half-test evaluation (20 model cells), with all other predictors unchanged. The criterion remains strong, \(r=0.982\). On the unregularized full rejected sets, alpha-1 validity remains lower for African-American than Caucasian individuals (\(64.6\%\) vs.\ \(82.3\%\), a \(17.7\)-point gap), in the same order as \(\mathbb P(\kappa\ge0)=0.646\) vs.\ \(0.805\). Matching the curvature distribution closes \(88.7\)--\(89.6\%\) of the gap under the two bin-free estimators (and \(83.0\)--\(91.3\%\) over 10, 20 and 50 bins). All three criteria fixed for this experiment are met, so neither the criterion nor the curvature-channel diagnosis depends on \texttt{decile\_score}. The additional arm replaces no row of the main suite.

\paragraph{Robustness without protected attributes.} A stronger test removes the protected attributes themselves. We repeat the same shallow COMPAS design after removing \texttt{race} and \texttt{sex} (and \texttt{decile\_score}) from the predictors, which leaves 8 features, and still group the held-out points by \texttt{race} at evaluation, so the model cannot use the protected attribute directly. The criterion remains strong (\(r=0.956\), 20 cells). The disparity does not vanish: on the unregularized full rejected sets alpha-1 validity is still lower for African-American than Caucasian individuals (\(70.6\%\) vs.\ \(81.5\%\), a \(10.9\)-point gap), in the same order as \(\mathbb P(\kappa\ge0)=0.697\) vs.\ \(0.802\). Across the main, decile-free and protected-excluded arms the raw gap shrinks (\(22\!\to\!17.7\!\to\!10.9\) pts) as protected and proxy features are removed, yet the curvature channel becomes \emph{more} dominant: matching the curvature distribution now closes \(93.5\)--\(98.1\%\) of the gap (bin-free) and \(96.0\)--\(99.3\%\) over 10, 20 and 50 bins. All three criteria of this experiment are met. Because the model has no access to \texttt{race} or \texttt{sex}, its residual disparity does not come from reading them; the disadvantaged group's rejected points lie in more concave regions of the score, although proxies such as age and prior counts remain among the inputs. This arm, too, replaces no row of the main suite.

\subsection{Channel Attribution and the Criterion-Implied Identity}
\label{S-sec:supp-channel-attribution}
The main text reports the criterion-implied closure, \(87.9\%\) (defined below). The bin-reweighting
estimator of the headline decomposition, which reweights the disadvantaged group to the advantaged
group's curvature histogram, gives \(80.4\)--\(86.9\%\) over 10/20/50 bins. Three qualifications
follow, all computed with the bin-reweighting estimator from the same per-point results
(unregularized arm, rejected points pooled over seeds 0--2). (1) \emph{Distance-only closure}: reweighting the disadvantaged group to the advantaged
distribution of the remainder scale \(s=d_p^2/a\) alone closes \(38.1\)--\(49.7\%\) of the gap
across the same bins; the disadvantaged group has the larger \(s\) (median \(0.749\) versus
\(0.199\)).
(2) \emph{Correlation}: \(\mathrm{corr}(\kappa,s)=-0.72\) pooled and \(-0.73\) within the
disadvantaged group, so the two scales are far from orthogonal and their solo closures cannot be
read as additive shares. (3) \emph{Sequential attribution}: reweighting on the joint
\((\kappa,s)\) distribution closes \(74.1\%\); assigning \(\kappa\) its solo closure first gives
it \(86\)--\(87\%\), while assigning it the joint increment over \(s\) gives it
\(29\)--\(36\%\), an ordering swing of \(29\)--\(87\%\).

\paragraph{The criterion-implied identity.} Let \(v_{\mathrm{adv}}\) and \(v_{\mathrm{dis}}\) be the
advantaged and disadvantaged groups' alpha-1 ray-hit validities, \(\mathrm{gap}=v_{\mathrm{adv}}-v_{\mathrm{dis}}\),
\(P_{\mathrm{adv}}\) the advantaged group's \(\mathbb P(\kappa\ge0)\), and \(p_\pm\) the disadvantaged
group's conditional alpha-1 ray-hit validities on \(\{\kappa\ge0\}\) and \(\{\kappa<0\}\). The two-bin (sign of
\(\kappa\)) threshold estimator, the reweighting that the signed-curvature criterion suggests, gives
the disadvantaged group the advantaged group's weights on these two bins, so its counterfactual
validity is \(v_{\mathrm{rw}}=P_{\mathrm{adv}}\,p_++(1-P_{\mathrm{adv}})\,p_-\). Its curvature-channel
closure \((v_{\mathrm{rw}}-v_{\mathrm{dis}})/\mathrm{gap}\) equals, by algebra,
\(1-\bigl(v_{\mathrm{adv}}-[P_{\mathrm{adv}}\,p_+ + (1-P_{\mathrm{adv}})\,p_-]\bigr)/\mathrm{gap}\). Measured: \(p_+=0.995\), \(p_-=0.024\), giving \(87.9\%\); since \(p_+\approx1\)
and \(p_-\approx0\), this simplifies to \(1-(v_{\mathrm{adv}}-P_{\mathrm{adv}})/\mathrm{gap}=
88.2\%\). The \(87.9\%\) figure is therefore close to \emph{forced} by the within-group
criterion together with the groups' \(\mathbb P(\kappa\ge0)\) difference (\(0.663\) vs.\
\(0.866\)): it is a consistency corollary of the criterion, not an independent mediation analysis. The
primary fairness finding is the \(\mathbb P(\kappa\ge0)\) difference itself; the decomposition
quantifies how completely the criterion transports it to the validity gap.

\subsection{Disparity Tests on Two Additional Real Datasets}
\label{S-sec:supp-b1}
To test whether the COMPAS curvature-channel disparity recurs, we ran a two-attempt protocol on
additional public datasets, fixed before any run. It prescribes real data only; the unregularized
training of the main comparison (\([128,64]\) Softplus, 50 epochs, five seeds); per-point geometry
from the same code as the main suites; a power threshold (\(\ge\!250\) pooled rejected points per
group); a materiality threshold (a pooled validity gap of at least 10 points); and at most two attempts,
with no change of states, encodings or thresholds between them. The first attempt uses folktables
ACSIncome \citep{ding2021retiring} (California 2018; race, White vs.\ Black; \(n{=}20{,}000\)). Both
groups sit at the all-concave floor, with \(\mathbb P(\kappa\ge0)=0.008\) in each, so alpha-1
validity is \(\approx\!0.5\%\) for everyone and the gap is \(-0.2\) points; as on Adult, a validity
disparity has no room to arise. The second attempt uses the UCI Taiwan credit-default data
\citep{yeh2009comparisons} (sex; full \(n{=}30{,}000\)). It shows a real gap of \(4.3\) points
(males lower), in the same direction as the groups' \(\mathbb P(\kappa\ge0)\) (\(0.19\) against
\(0.25\)) but below the materiality threshold. Both attempts are therefore negative under
the protocol's thresholds, and both are reported in full. The COMPAS disparity thus does not
recur on the four comparison datasets. A large validity disparity appears to
need an intermediate validity level (about \(74\%\) on COMPAS), whereas the floor regime (ACSIncome,
Adult) and the near-ceiling regime (German) leave it no room. As out-of-sample support for the
criterion, the per-group orderings match on both datasets: validity \(\approx\mathbb
P(\kappa\ge0)\) at the floor below \(1\%\) on ACSIncome, and in direction at mid-scale on Taiwan.

\subsection{Historical Note (Not Used as Evidence): ResNet-18 Scale Suite on CelebA}
\label{S-sec:supp-celeba}

\paragraph{Status.} This subsection records a historical result. It is not used as evidence for any
claim of the main paper or of this supplement. It cannot be regenerated from the retained code,
because the script that produced it computed \(\kappa\) with a variant of the curvature helper
that was not retained; the checkpoints and point-level ray records were not retained either, and
the aggregates lack run, configuration, split and calibration provenance. The reported quantities
depend on \(\kappa\) only through its sign, and everything else they use, apart from each model's
test accuracy, comes from retained code (S10 gives the details). The rows also contain
signed-quadratic and conformal columns, which depend on the magnitude of \(\kappa\); they are not
reported.

\paragraph{Suite.} The aggregates cover ResNet-18 models on three CelebA binary attributes
(\texttt{smiling}, \texttt{heavy\_makeup}, \texttt{attractive}), two activation regimes, four
training methods (unreg, MW, Global, Asym), and five seeds (117 rows). In the smoothed regime every
ReLU is replaced by Softplus(\(\beta=10\)); the stem max-pooling is retained, so the score is only
piecewise twice differentiable, \(\kappa\) is the almost-everywhere Hessian form, and Theorem 5.2's
hypothesis can fail across pooling switches. In the ReLU regime \(d_p\) and \(d_{\text{ray}}\) come
from the ReLU network and \(\kappa\) from a Softplus(\(\beta=10\)) copy with the same weights. The
rows record per-model \(\mathbb P(\kappa\ge0)\), rule validity and overshoot, and denied-recourse
rates. Every validity of the suite is the ray-hit event (a bracketed crossing with
\(d_{\text{ray}}\le d_p+10^{-6}\)); endpoint success was not computed. Of the 60 planned smoothed
runs, 57 have a row; attractive/MW seeds 3--4 and smiling/Global seed 1 have none, most likely
because the producer writes no row for a model that rejects fewer than 10 test images (no run logs
were retained).

\paragraph{Recipe.} The label is the attribute and the protected attribute is Male.
Per seed \(s\in\{0,\dots,4\}\), \texttt{numpy.random.RandomState(s).choice} draws 15{,}000 images
from the official training partition and then 3{,}000 from the official test partition, without
replacement; the four methods of a seed share the draw. Images are center-cropped to 160 pixels
and resized to 64, with pixels in \([0,1]\) and no normalization or augmentation. The model is a
randomly initialized torchvision ResNet-18 with one output logit, trained with Adam (learning rate
\(10^{-3}\)), shuffled minibatches of 128, 5 epochs, and unweighted \texttt{BCEWithLogitsLoss}.
Penalties are evaluated with BatchNorm in inference mode, with \(\delta=0.05\), \(\lambda=10\)
(Asym) and \(\lambda=0.1\) (MW, Global). Accuracy is overall accuracy on the 3{,}000 test images.
The rejected test images are split in half by the permutation of S3.2 (\texttt{RandomState(1000+s)});
\(\mathbb P(\kappa\ge0)\), validity, bracketing and denial are computed on the second half. The ray
search is that of S3.2 (160-point scan to \(\max(8d_p,10^{-3})\), 30 bisection steps). The study has
not been rerun. A rerun would need a reimplementation of the curvature helper and could at best
agree in distribution, not bit for bit, because the image subsets were drawn from a local copy of
CelebA whose image-list order was not recorded, and GPU determinism was requested but not
enforced.

\paragraph{What the aggregates record.} Across the 57 smoothed rows, curvature regularization spreads \(\mathbb P(\kappa\ge0)\) over nearly the full range \([0,1]\), and alpha-1 ray-hit validity is correlated with it at Pearson \(r=0.76\) (\(n=57\)). Most of these models have overall accuracy below \(0.60\), because the regularization strengths that produce the spread also collapse accuracy. Restricting to models with overall accuracy of at least \(0.60\) or \(0.70\), thresholds chosen after the unrestricted result was known, gives \(r=0.78\) (\(n=19\)) and \(r=0.84\) (\(n=15\)); on heavy\_makeup the \(0.60\) threshold admits models close to the majority-class rate. A restriction by balanced accuracy cannot be computed from the aggregates: \texttt{acc} is overall accuracy, whereas \texttt{dr\_rate\_male}/\texttt{dr\_rate\_female} are recourse-denial rates by protected sex among rejected users, not recalls of the target classes.

A rejected user is denied recourse when \(d_{\text{ray}}>d_p\), with unbracketed rays counted as denied; this strict event omits the \(10^{-6}\) tolerance of validity, so a boundary-coincident ray can count as both valid and denied. On the smoothed unregularized models (five seeds per attribute), the closed-form alpha-1 step denies recourse to \(90\%\) (attractive), \(55\%\) (heavy\_makeup) and \(72\%\) (smiling) of rejected users, and on heavy\_makeup to \(81\%\) of rejected women versus \(38\%\) of rejected men.

Ray line search is valid whenever the scan brackets a crossing; averaged over the 57 smoothed models it brackets \(91\%\) of the evaluated rejected images (\texttt{v\_linesearch}), and the unbracketed rays fall almost entirely on the accuracy-collapsed regularized models, mostly MW and Global. An unbracketed ray counts as invalid in every CelebA validity above. The ReLU rows are identical across the four methods: the \(60\) ReLU rows are \(15\) attribute\(\times\)seed models, each repeated as four identical method rows (\(15/15\) groups identical), so the ReLU block has \(15\) distinct rows, not \(60\), and the correlation above uses the \(57\) smoothed models only. The way the penalties are evaluated explains this. BatchNorm runs in inference mode there, so the ReLU ResNet-18 (with max-pooling) is piecewise affine in its input, the Hutchinson penalties are exactly zero, and the asymmetric hinge reduces to \(\delta^2\) times the mean of its detached weights, all with zero parameter gradient (BatchNorm in training mode would restore a nonzero input Hessian). Because no checkpoints or training traces were retained, the identical rows cannot by themselves rule out that checkpoints were reused.

\section{S5. Extended Inflation and Validation-Tuned Baselines}

The fixed-inflation and oracle-on-path experiments use three seeds for COMPAS, German, and F-MNIST, capped at 128 rejected test points per seed. Constant inflation recommends \(\alpha d_p\). Oracle-on-path recommends \(d_{\text{ray}}\), which is valid by construction when the crossing is bracketed, but it needs a per-user search with model queries and is not the same deployment regime as closed-form alpha-1 recourse.

\textbf{Settings.} All four suites use seeds 0, 1 and 2, train each model on 2{,}000 training rows drawn without replacement with \texttt{RandomState(seed)}, with the \((\lambda,\delta)\) of the five-seed suite in S3.2, and evaluate it on the first 128 rejected test points. Models train for 10 epochs in the COMPAS/German fixed-rule suite and for 5 epochs in the F-MNIST fixed-rule suite and in both validation-tuned suites. Constant inflation uses \(\alpha\in\{1,1.05,1.1,1.2,1.5,2\}\). Validation-tuned inflation holds out 20\% of the 2{,}000 rows (a \texttt{RandomState(seed)} permutation), trains on the rest, and picks the smallest \(\alpha\) in \(\{1.00,1.01,\dots,1.05,1.075,1.1,1.15,1.2,1.3,1.5,2\}\) whose validity on the first 128 rejected validation points reaches the target (0.90, 0.95, or 0.99), or else the \(\alpha\) with the highest validation validity. S10 gives the commands.

\textbf{Relationship to the main comparison.} This suite retrains its own models (separate checkpoints and seeds, 128-point rejected cap) and is distinct from the models of the main comparison. Its purpose is to compare inference-time recourse rules on a common footing, not to reproduce the alpha-1 validities of the main comparison. The large standard deviations of the non-asymmetric \(V_1\) entries below (e.g.\ COMPAS \(34.6\pm56.6\), German \(53.0\pm49.8\)) reflect a strongly bimodal outcome across seeds: under this capped suite some seeds give near-0\% and others near-100\% alpha-1 validity, so these means are not directly comparable to the alpha-1 validities on the full test sets of the main comparison. The qualitative conclusions hold in both suites: exact alpha-1 can fail, inflation recovers validity at the cost of overshoot, and Asym is strongest at alpha-1.

\begin{table*}[t]
\centering
\small
\caption{Fixed-inflation summary from three-seed CSVs. \(V_1\) is alpha-1 ray validity. \(\alpha_{q}\) is the first grid alpha whose validity reaches level \(q\), averaged over the seeds that reach it (S100 counts the seeds that reach \(q=1\)).}
\label{S-tab:supp-fixed-inflation}
\begin{tabular}{llrrrrrrrr}
\hline
Data & Meth. & \(V_1\) & \(\alpha_{95}\) & \(O_{95}\) & \(\alpha_{99}\) & \(O_{99}\) & \(\alpha_{100}\) & \(O_{100}\) & S100 \\
\hline
compas & asym & \(100.0\pm0.0\) & 1.000 & 0.139 & 1.000 & 0.139 & 1.000 & 0.139 & 3 \\
compas & glob & \(34.6\pm56.6\) & 1.033 & 0.040 & 1.033 & 0.040 & 1.033 & 0.040 & 3 \\
compas & mw & \(34.6\pm56.6\) & 1.033 & 0.039 & 1.033 & 0.039 & 1.033 & 0.039 & 3 \\
compas & unreg & \(35.4\pm56.0\) & 1.033 & 0.041 & 1.033 & 0.041 & 1.033 & 0.041 & 3 \\
german & asym & \(97.8\pm3.7\) & 1.017 & 0.031 & 1.017 & 0.031 & 1.017 & 0.031 & 3 \\
german & glob & \(53.0\pm49.8\) & 1.033 & 0.038 & 1.050 & 0.056 & 1.050 & 0.056 & 3 \\
german & mw & \(53.0\pm49.8\) & 1.033 & 0.038 & 1.050 & 0.056 & 1.050 & 0.056 & 3 \\
german & unreg & \(62.9\pm54.8\) & 1.033 & 0.039 & 1.033 & 0.039 & 1.033 & 0.039 & 3 \\
fmnist & asym & \(98.2\pm1.8\) & 1.000 & 0.399 & 1.017 & 0.461 & 1.267 & 1.389 & 3 \\
fmnist & glob & \(32.0\pm27.6\) & 1.083 & 0.232 & 1.167 & 0.488 & 1.400 & 1.235 & 3 \\
fmnist & mw & \(32.0\pm27.6\) & 1.083 & 0.231 & 1.167 & 0.488 & 1.400 & 1.234 & 3 \\
fmnist & unreg & \(30.2\pm38.2\) & 1.117 & 0.321 & 1.133 & 0.371 & 1.300 & 0.889 & 3 \\
\hline
\end{tabular}
\end{table*}

\begin{table*}[t]
\centering
\small
\caption{Validation-tuned inflation summary. \(\alpha\) is selected on rejected validation examples. Gap is validation minus test validity.}
\label{S-tab:supp-validation-tuned}
\begin{tabular}{lllrrrrrr}
\hline
Data & Target & Meth. & Sel. \(\alpha\) & Test \(V\) & Gap & Under & Over & \(n\) \\
\hline
compas & 0.90 & asym & 1.000 & \(98.4\pm2.7\) & -0.013 & 0.000 & 0.038 & 128--128 \\
compas & 0.95 & asym & 1.003 & \(99.7\pm0.5\) & -0.005 & 0.000 & 0.042 & 128--128 \\
compas & 0.99 & asym & 1.007 & \(100.0\pm0.0\) & 0.000 & 0.000 & 0.046 & 128--128 \\
compas & 0.90 & glob & 1.017 & \(96.6\pm3.0\) & 0.000 & 0.001 & 0.021 & 128--128 \\
compas & 0.95 & glob & 1.020 & \(97.9\pm2.4\) & 0.005 & 0.000 & 0.025 & 128--128 \\
compas & 0.99 & glob & 1.032 & \(99.7\pm0.5\) & 0.003 & 0.000 & 0.039 & 128--128 \\
compas & 0.90 & mw & 1.017 & \(96.1\pm3.4\) & 0.005 & 0.001 & 0.021 & 128--128 \\
compas & 0.95 & mw & 1.020 & \(97.7\pm2.8\) & 0.003 & 0.000 & 0.025 & 128--128 \\
compas & 0.99 & mw & 1.032 & \(99.7\pm0.5\) & 0.003 & 0.000 & 0.039 & 128--128 \\
compas & 0.90 & unreg & 1.017 & \(96.1\pm3.4\) & -0.016 & 0.001 & 0.019 & 128--128 \\
compas & 0.95 & unreg & 1.023 & \(98.7\pm1.6\) & -0.003 & 0.000 & 0.026 & 128--128 \\
compas & 0.99 & unreg & 1.032 & \(99.7\pm0.5\) & 0.003 & 0.000 & 0.036 & 128--128 \\
german & 0.90 & asym & 1.007 & \(95.7\pm5.4\) & 0.004 & 0.001 & 0.008 & 40--107 \\
german & 0.95 & asym & 1.010 & \(98.5\pm1.3\) & -0.002 & 0.000 & 0.011 & 40--107 \\
german & 0.99 & asym & 1.020 & \(100.0\pm0.0\) & 0.000 & 0.000 & 0.023 & 40--107 \\
german & 0.90 & glob & 1.010 & \(96.0\pm5.5\) & 0.005 & 0.001 & 0.009 & 39--112 \\
german & 0.95 & glob & 1.013 & \(98.6\pm1.3\) & 0.000 & 0.000 & 0.012 & 39--112 \\
german & 0.99 & glob & 1.020 & \(99.1\pm1.5\) & 0.005 & 0.000 & 0.020 & 39--112 \\
german & 0.90 & mw & 1.010 & \(96.0\pm5.5\) & 0.005 & 0.001 & 0.009 & 39--112 \\
german & 0.95 & mw & 1.013 & \(98.6\pm1.3\) & 0.000 & 0.000 & 0.012 & 39--112 \\
german & 0.99 & mw & 1.020 & \(99.1\pm1.5\) & 0.005 & 0.000 & 0.020 & 39--112 \\
german & 0.90 & unreg & 1.010 & \(95.1\pm5.0\) & 0.006 & 0.001 & 0.008 & 40--108 \\
german & 0.95 & unreg & 1.013 & \(97.6\pm2.3\) & 0.009 & 0.000 & 0.011 & 40--108 \\
german & 0.99 & unreg & 1.027 & \(99.2\pm1.4\) & 0.008 & 0.000 & 0.026 & 40--108 \\
fmnist & 0.90 & asym & 1.000 & \(97.9\pm2.3\) & 0.000 & 0.013 & 0.366 & 128--128 \\
fmnist & 0.95 & asym & 1.007 & \(98.7\pm0.9\) & -0.003 & 0.011 & 0.391 & 128--128 \\
fmnist & 0.99 & asym & 1.033 & \(99.2\pm0.0\) & 0.003 & 0.007 & 0.493 & 128--128 \\
fmnist & 0.90 & glob & 1.045 & \(92.4\pm0.5\) & -0.005 & 0.012 & 0.116 & 128--128 \\
fmnist & 0.95 & glob & 1.065 & \(96.4\pm0.5\) & -0.005 & 0.006 & 0.177 & 128--128 \\
fmnist & 0.99 & glob & 1.113 & \(99.7\pm0.5\) & -0.003 & 0.002 & 0.335 & 128--128 \\
fmnist & 0.90 & mw & 1.045 & \(92.4\pm0.5\) & -0.005 & 0.012 & 0.115 & 128--128 \\
fmnist & 0.95 & mw & 1.065 & \(96.4\pm0.5\) & -0.005 & 0.006 & 0.177 & 128--128 \\
fmnist & 0.99 & mw & 1.113 & \(99.7\pm0.5\) & -0.003 & 0.002 & 0.334 & 128--128 \\
fmnist & 0.90 & unreg & 1.050 & \(94.5\pm1.6\) & -0.013 & 0.005 & 0.126 & 128--128 \\
fmnist & 0.95 & unreg & 1.078 & \(98.4\pm1.4\) & -0.008 & 0.002 & 0.216 & 128--128 \\
fmnist & 0.99 & unreg & 1.107 & \(99.5\pm0.9\) & -0.003 & 0.001 & 0.308 & 128--128 \\
\hline
\end{tabular}
\end{table*}

\begin{table*}[t]
\centering
\small
\caption{Matched-validity selected rows. VT denotes validation-tuned inflation and CI denotes constant inflation. For each dataset, target, and group (Asym, or any of the other three methods) the row is the candidate (CI at a grid \(\alpha\), or VT at any tuning target) with the lowest mean overshoot among those whose validity reaches the target on all three seeds; the 100\% row uses target 0.999. Costs use \(C_U/C_O=5\).}
\label{S-tab:supp-matched-validity}
\begin{tabular}{lllrrrr}
\hline
Data & Target & Group/method/rule & Valid. & Over & Bin. cost & Eff. cost \\
\hline
compas & 0.95 & Asym asym VT & 0.984 & 0.038 & 0.116 & 0.039 \\
compas & 0.95 & non-Asym glob VT & 0.979 & 0.025 & 0.129 & 0.027 \\
compas & 0.99 & Asym asym VT & 0.997 & 0.042 & 0.055 & 0.042 \\
compas & 0.99 & non-Asym unreg VT & 0.997 & 0.036 & 0.049 & 0.036 \\
compas & 100\% & Asym asym VT & 1.000 & 0.046 & 0.046 & 0.046 \\
compas & 100\% & non-Asym mw CI & 1.000 & 0.059 & 0.059 & 0.059 \\
fmnist & 0.95 & Asym asym VT & 0.979 & 0.366 & 0.470 & 0.430 \\
fmnist & 0.95 & non-Asym mw VT & 0.964 & 0.177 & 0.359 & 0.205 \\
fmnist & 0.99 & Asym asym VT & 0.992 & 0.493 & 0.532 & 0.529 \\
fmnist & 0.99 & non-Asym mw VT & 0.997 & 0.334 & 0.347 & 0.343 \\
fmnist & 100\% & Asym asym CI & 1.000 & 2.226 & 2.226 & 2.226 \\
fmnist & 100\% & non-Asym unreg CI & 1.000 & 1.499 & 1.499 & 1.499 \\
german & 0.95 & Asym asym VT & 0.985 & 0.011 & 0.084 & 0.013 \\
german & 0.95 & non-Asym unreg VT & 0.976 & 0.011 & 0.130 & 0.013 \\
german & 0.99 & Asym asym VT & 1.000 & 0.023 & 0.023 & 0.023 \\
german & 0.99 & non-Asym unreg CI & 1.000 & 0.056 & 0.056 & 0.056 \\
german & 100\% & Asym asym VT & 1.000 & 0.023 & 0.023 & 0.023 \\
german & 100\% & non-Asym unreg CI & 1.000 & 0.056 & 0.056 & 0.056 \\
\hline
\end{tabular}
\end{table*}

The matched-validity comparison reuses the stored three-seed results; no model was trained or evaluated for it. Its models train for fewer epochs than the models of the main paper, and on COMPAS and F-MNIST on fewer rows (Settings above), so it does not show how those models would compare at matched validity. Asym has no blanket advantage at matched validity. Validation-tuned inflation on the other methods wins on overshoot and on both costs for COMPAS at 0.99 and for F-MNIST at 0.95 and 0.99. Asym wins German at 0.99 and in the 100\% row, COMPAS in the 100\% row, and some COMPAS/German settings at 0.95 under the binary-denial cost. Once \(\alpha\) is selected on validation data, validation-tuned inflation is closed-form at inference; ray line search remains the oracle that needs a per-user search.

\section{S6. Actionable-Recourse Audit Details}

\subsection{Constrained-Action Geometry (the masks are a special case)}

Actionable recourse restricts the step: immutable features cannot move, some features move in one direction only, and budgets cap the rest. The user then ascends not along the gradient \(\hat g\) but along the steepest \emph{feasible} direction. Let the admissible action directions form a closed convex cone \(A\) (immutable \(i\): \(v_i=0\); increase-only \(j\): \(v_j\ge0\); intersections of coordinate subspaces and half-spaces through the origin). A box or budget is \emph{not} a cone, because it caps magnitudes rather than directions. Its tangent cone at \(x\) has the form above, and the results below then hold along the feasible ray only up to the first binding constraint \(t_{\max}=\sup\{t:x+tv\ \text{feasible}\}\); a crossing beyond \(t_{\max}\) means that recourse is infeasible within the budget, not that it is mispriced. When \(\Pi_A(\nabla f)\ne0\), the feasible recourse direction, its directional derivative, promised distance, and directional curvature are
\[
\begin{aligned}
&v=\frac{\Pi_A(\nabla f)}{\|\Pi_A(\nabla f)\|},\quad a_A=\|\Pi_A(\nabla f)\|,\\
&d_{p,A}=\frac{|f(x)|}{a_A},\quad \kappa_A=v^{\top}\nabla^2 f\,v .
\end{aligned}
\]
By Moreau's identity for a convex cone, \(\langle\nabla f,\Pi_A\nabla f\rangle=\|\Pi_A\nabla f\|^2\), so the feasible directional derivative is \(\langle\nabla f,v\rangle=a_A\); and since \(\|\Pi_A\nabla f\|\le\|\nabla f\|\), we always have \(d_{p,A}\ge d_p\), so constraints can only push the promised boundary farther away. If \(\Pi_A(\nabla f)=0\), there is no first-order feasible ascent direction, \(d_{p,A}\) is undefined (equivalently infinite), and the ray transfer below does not apply.

\smallskip
\noindent\textbf{Proposition (constrained-action transfer).} \textit{Let \(a_A>0\), and let \(v\), \(d_{p,A}\) and \(\kappa_A\) be as above. The feasible profile \(\phi_A(t)=f(x+tv)\) satisfies \(\phi_A(0)=-m\), \(\phi_A'(0)=a_A\) and \(\phi_A''(0)=\kappa_A\), and \(d_{p,A}\ge d_p\). Let \(d^A_{\mathrm{ray}}\) be the first crossing of \(\phi_A\) and \(s_A=2\kappa_Ad_{p,A}/a_A\). Then:}
\textit{(i) (Lemma 5.1) If \(s_A>-1\), the smallest positive root of \(q_A(t)=-m+a_At+\tfrac{\kappa_A}2t^2\) is \(d^A_{\mathrm{quad}}=2d_{p,A}/(1+\sqrt{1+s_A})\), and \(d^A_{\mathrm{quad}}-d_{p,A}=-\tfrac{\kappa_Ad_{p,A}^2}{2a_A}\rho(s_A)\).}
\textit{(ii) (Theorem 5.2) If \(\phi_A''\) is \(M_A\)-Lipschitz on \([0,2d_{p,A}]\), \(4|\kappa_A|d_{p,A}\le a_A\) and \(8M_Ad_{p,A}^2\le a_A\), then \(\phi_A\) is strictly increasing on \([0,2d_{p,A}]\), \(d^A_{\mathrm{ray}}\le2d_{p,A}\), and \(|d^A_{\mathrm{ray}}-d^A_{\mathrm{quad}}|\le8M_Ad_{p,A}^3/(3a_A)\).}
\textit{(iii) (Theorem 6.1) Let \(K>0\) with \(Kd_{p,A}<a_A/2\). There are two \(C^2\) scores with value \(f(x)\) and gradient \(\nabla f(x)\) at \(x\) whose feasible profiles satisfy \(|\phi_A''|\le K\) and whose feasible first crossings differ by at least \(Kd_{p,A}^2/a_A\). Hence every rule that reads only \(f(x)\) and \(\nabla f(x)\) and recommends a distance along \(v\) is either invalid on one of them or overshoots the other by at least \(Kd_{p,A}^2/a_A\).}
\textit{(iv) (Proposition 7.1) Split conformal calibration with residuals \(d^A_{\mathrm{ray}}-d_{\mathrm{base}}\) computed on feasible rays satisfies the conclusion of Proposition 7.1 under its hypotheses.}
\textit{Each statement concerns the feasible ray \(x+tv\). For a box or budget, whose first binding constraint is at \(t_{\max}\), it describes feasible recourse only when the segment it uses lies in \([0,t_{\max}]\): \([0,2d_{p,A}]\) for (ii) and (iii), and \([0,d_{\mathrm{conf}}]\) for (iv).}

\smallskip
\noindent\textit{Proof.} By Moreau's identity, \(\phi_A'(0)=\langle\nabla f(x),v\rangle=a_A\); \(\phi_A(0)=f(x)=-m\) and \(\phi_A''(0)=v^\top\nabla^2f(x)v=\kappa_A\) by definition, and \(d_{p,A}=m/a_A\ge m/a=d_p\) because \(\|\Pi_A\nabla f\|\le\|\nabla f\|\). (i) is the algebra of Lemma 5.1 with \((a,d_p,\kappa)\) replaced by \((a_A,d_{p,A},\kappa_A)\). (ii) The proofs of Theorem 5.2 and Lemma~\ref{S-lem:s1-no-recross} use only the value, slope and second derivative of the profile at \(0\), the Lipschitz bound on its second derivative on \([0,2d_p]\) and regime \((\star)\); applied to \(\phi_A\), they give (ii). (iii) The scores \(f_\pm(x+w)=f(x)+\langle\nabla f(x),w\rangle\pm\tfrac K2\langle v,w\rangle^2\) have value \(f(x)\) and gradient \(\nabla f(x)\) at \(x\), and feasible profiles \(-m+a_At\pm\tfrac K2t^2\); the proof of Theorem 6.1 with \((a,d_p)\) replaced by \((a_A,d_{p,A})\) gives (iii). (iv) The rank argument of Proposition 7.1 uses only the exchangeability of the residuals. The curvature \(\kappa_A\) costs one HVP \(\nabla^2f(x)\,v\). \hfill\(\square\)

\smallskip
\noindent\textbf{Corollary (feature-freezing is the coordinate-projection case).} \textit{When \(A=\{v:v_i=0,\ i\in\mathcal F\}\) for a frozen set \(\mathcal F\), \(\Pi_A\) is the diagonal \(0/1\) projection that zeroes the frozen coordinates and \(v\propto(\nabla f)_{\mathcal F^{c}}\). The mask audit of this section is exactly this case: a principled instance of one geometry, with its own \(d_{p,A},\kappa_A\) and the same one-HVP rule and certificate.}

\smallskip
Actionability changes the curvature that the rule must read. In general \(\kappa_A=v^{\top}\nabla^2 f\,v\) is not the gradient-ray curvature \(\kappa=\hat g^{\top}\nabla^2 f\,\hat g\): over \(2\times10^{5}\) synthetic random jets (not model points; S3.2) the two differ in sign in \(31.7\%\) of cases, and \(d_{p,A}\ge d_p\) holds in \(100\%\). The criterion then applies with the \emph{constrained} marginal \(\mathbb P(\kappa_A\ge0)\), and the asymmetric penalty has an actionable variant that penalizes \([\kappa_A]_-\), the concavity along the feasible direction, which aligns the boundary with directions in which users can actually move. If actions are exerted through a map \(x=h(z)\), as in causal or manifold models of action, the feasible path is \(t\mapsto h(z+tv)\), and the effective second-order term gains a contribution \(\langle\nabla f,h''[v,v]\rangle\) from the action manifold beyond \(v^{\top}J_h^{\top}\nabla^2 f\,J_h v\). Recourse can then fail from the curvature of feasible actions alone, even against a flat boundary. The signed-quadratic rule still applies with this effective curvature; we note it as the natural generalization and do not test it here.

\subsection{Empirical Mask Audit}

The experiment below is the coordinate-projection case of the Corollary: projected-gradient recourse under simple feature-freezing masks. It is neither causal recourse nor a full actionable-recourse solver. The \texttt{protected\_only} mask freezes the protected-attribute features, and the \texttt{strict} mask freezes the broader set of features we designate as non-actionable.

\begin{table*}[t]
\centering
\small
\caption{Feature-freezing coordinate-projection audit. Validity is ray-hit validity on the projected score-space path, not causal or fully feasible recourse. Dashes indicate zero movable dimensions or excluded metrics.}
\label{S-tab:supp-actionable}
\begin{tabular}{lllrrrrr}
\hline
Data & Method & Mask & Dims & Valid. & Under & Over & \(n\) \\
\hline
compas & unreg & protected\_only & 6 & 97.3 & 0.013 & 0.113 & 470 \\
compas & unreg & strict & 0 & -- & -- & -- & 470 \\
compas & mw & protected\_only & 6 & 99.5 & 0.000 & 0.052 & 473 \\
compas & mw & strict & 0 & -- & -- & -- & 473 \\
compas & glob & protected\_only & 6 & 99.5 & 0.000 & 0.040 & 476 \\
compas & glob & strict & 0 & -- & -- & -- & 476 \\
compas & asym & protected\_only & 6 & 100.0 & 0.000 & 0.314 & 481 \\
compas & asym & strict & 0 & -- & -- & -- & 481 \\
german & unreg & protected\_only & 44 & 100.0 & 0.000 & 0.020 & 91 \\
german & unreg & strict & 6 & 90.0 & 0.001 & 0.037 & 91 \\
german & mw & protected\_only & 44 & 100.0 & 0.000 & 0.016 & 82 \\
german & mw & strict & 6 & 87.6 & 0.001 & 0.032 & 82 \\
german & glob & protected\_only & 44 & 100.0 & 0.000 & 0.014 & 82 \\
german & glob & strict & 6 & 85.9 & 0.001 & 0.028 & 82 \\
german & asym & protected\_only & 44 & 100.0 & 0.000 & 0.031 & 91 \\
german & asym & strict & 6 & 99.6 & 0.000 & 0.067 & 91 \\
\hline
\end{tabular}
\end{table*}

Besides the protected features (age, sex and race), the COMPAS strict mask freezes the juvenile and prior offense counts, the charge degree and the COMPAS decile score, which are all the remaining features, so it leaves no first-order movable dimension; the strict German mask remains nondegenerate. Under the coordinate masks, projected-ray
validity is high: \(97\)--\(100\%\) on COMPAS with the protected-only mask, \(100\%\) on German with
the protected-only mask for every method, and \(86\)--\(90\%\) on German with the strict mask for
the non-asymmetric methods against \(99.6\%\) for Asym. Asymmetric training thus has a real but
modest edge on the strictest mask.
Entries are means over 5 seeds of models trained on features standardized with training-split
statistics, as in the main experiments. F-MNIST is excluded because its masks freeze no features,
so the experiment would reduce to the unconstrained ray.

\paragraph{Fairness under feature-freezing geometry.} The COMPAS reliability disparity of the main
text is measured on the unconstrained gradient ray. Here we repeat a descriptive per-group
comparison on coordinate-projection rays with the protected features frozen (unregularized models,
5 seeds). The observations repeat the same test people, so pooled Fisher and Mann--Whitney tests,
which treat them as independent, are not used. Under \texttt{protected\_only}, ray-hit validity is
\(96.5\%\) for AfAm against \(99.4\%\) for Cauc (gap \(2.9\) points), below the 5-point materiality
threshold fixed in advance, so no curvature-channel decomposition is computed. The median
\(d_{p,A}\) is \(1.88\times\) larger for the disadvantaged group (\(1.90\times\) under the second
mask), with \(\mathbb P(\kappa_A\ge0)\) of \(0.949\) against \(0.987\). These summaries over persons
and seeds are descriptive (the cluster analysis below supports no inference), and they are not
evidence of causal recourse or of legally feasible actions.

\paragraph{Stable-ID cluster analysis.}
This analysis uses stable IDs for the test persons, averages the repeated seed observations of each
person, and applies a person-cluster bootstrap (\(10{,}000\) draws; seed \(1305\)) fixed before the
run. The protocol first requires that the stored marginal summaries reproduce to within
\(10^{-12}\). The validity marginal reproduces exactly, but the recomputed effort ratio differs
from the stored one in the ninth significant digit, because the source values were rounded to
float32 when they were written to CSV. The requirement therefore fails, and every interval below is
diagnostic only. For \texttt{protected\_only}, the person-cluster interval for the
Caucasian-minus-African-American endpoint gap is \(0.0074\)--\(0.0405\) (all five seed effects have
the same sign, and the whole interval lies inside the band \([-0.05,0.05]\) fixed in advance); the
corresponding log-effort interval is \(0.234\)--\(0.742\), again with five same-sign seed effects.
The per-person records and summaries allow the analysis to be reproduced (S10), but they do not
support a significance claim.

\section{S7. Cost-Model Audit Details}

Under an explicit travel-error cost (\(C_U/C_O=5\)), Asym is strongest on binary-denial cost \emph{among closed-form alpha-1 rules}, while validation-tuned inflation is strongest on effort cost and can win on binary-denial cost once tuned (e.g.\ F-MNIST). No trained alpha-1 penalty in our suites moved the tradeoff (S8), and the matched-validity comparison shows that Asym has no blanket overshoot advantage once the other methods are tuned (S5). All these rules lie on one frontier indexed by the query budget.

The travel-error costs are computed from the stored three-seed baseline results. Let \(U\) be mean ray undershoot, \(O\) mean ray overshoot, and \(V\) ray validity. For cost ratio \(r=C_U/C_O\),
\[
\begin{aligned}
\text{effort-magnitude} &= rU+O,\\
\text{binary-denial} &= r(1-V)+O.
\end{aligned}
\]
The result files report both seed-mean and \(n_{\text{evaluated}}\)-weighted costs. Adult is excluded because the three-seed baseline suite has no matching Adult results.

When all rules compete, oracle-on-path line search wins every cost setting under these formulas, because it recommends \(d_{\text{ray}}\) and so has validity 1 and overshoot 0 whenever the crossing is bracketed. The cost of queries and search is not modeled, so this row is an oracle benchmark, not a practical deployment comparison.

\begin{table*}[t]
\centering
\small
\caption{Best non-oracle winners at \(C_U/C_O=5\). The last column is the mean cost over the three seeds,
with sample standard deviations (ddof \(=1\)) 0.022, 0.006, 0.055, 0.051, 0.014 and 0.001 in row order.}
\label{S-tab:supp-nonoracle-cost}
\begin{tabular}{lllrr}
\hline
Data & Cost & Best & \(\alpha\) & Cost \\
\hline
compas & binary & asym VT & 1.007 & 0.046 \\
compas & effort & unreg VT & 1.017 & 0.024 \\
fmnist & binary & unreg VT & 1.078 & 0.294 \\
fmnist & effort & unreg VT & 1.050 & 0.153 \\
german & binary & asym VT & 1.020 & 0.023 \\
german & effort & asym VT & 1.007 & 0.011 \\
\hline
\end{tabular}
\end{table*}

\section{S8. Failed Attempts and Negative Results}

This section collects training-time attempts that did not improve the validity--overshoot frontier in the reported experiments. They concern only the trained alpha-1 penalties we tested and prove no impossibility result: a steepness penalty that preserves accuracy, a different model family, or a recourse rule that takes more than one step could still change the frontier.

\paragraph{A smoothed surrogate does not carry the criterion to tree ensembles.} We asked whether the criterion can assess a trained gradient-boosted tree ensemble (\texttt{HistGradientBoosting\-Classifier}) with no retraining and no HVP. The alpha-1 step is issued along the gradient \(\nabla f_\sigma\) of a Gaussian-smoothed surrogate, and the true model's crossing on that path is found by bisection with forward evaluations only (three datasets, three seeds, \(\sigma\in\{0.25,0.5,1,2\}\)). This fails. \(\mathbb P(\kappa_\sigma\ge0)\) does not track the ensemble's ray validity (Pearson \(r=-0.53\) over 33 cells; within datasets \(r=-0.36\) on COMPAS, \(+0.35\) on German and \(-0.15\) on Adult, mean \(-0.05\)), and on Adult validity stays near zero even at \(\mathbb P(\kappa_\sigma\ge0)\approx0.98\). The smoothing scale that makes the tree differentiable also makes the surrogate's promised distance a poor predictor of the jagged true boundary, so the leading-order quadratic approximation on which the criterion relies is not accurate at the scale of \(d_p\) for piecewise-constant models. The analysis remains limited to models with a locally smooth boundary; extending it to tree ensembles is open.

\begin{table}[H]
\centering
\small
\caption{Negative results of training-time attempts. The first two rows were regenerated with the current code; accuracy is plain accuracy and validity is ray-hit validity. S10 lists the scripts.}
\label{S-tab:supp-negative}
\begin{tabular}{p{0.16\textwidth}p{0.33\textwidth}p{0.37\textwidth}}
\hline
Attempt & Intended mechanism & Observed outcome \\
\hline
Calibrated curvature target & Replace the fixed target \(\delta\) of the asymmetric penalty by a per-point target \(\min(2m/d_p^2,\text{cap})\), meant to clear the alpha-1 boundary with less excess curvature. & On COMPAS, Adult, and F-MNIST every calibrated point of the two-seed frontier run has lower validity and higher overshoot than some fixed-\(\delta\) point, and all six three-seed configurations cost more than the fixed-\(\delta\) penalty in both travel-error costs at \(C_U/C_O=5\). \\
Steepness penalty & Add \(\lambda_d\) times the mean \(d_p^2\) of rejected minibatch points to the asymmetric penalty (\(\delta=0.05\)) to change the fixed-scale tradeoff. & For every \(\lambda_d\in\{0.1,0.5,2\}\), accuracy falls from \(67\)/\(80\)/\(89\%\) at \(\lambda_d=0\) to \(53\)/\(23\)/\(49\%\) (COMPAS/Adult/F-MNIST), and at least one of three seeds rejects no test point. \\
Adult symmetric penalty sweep & Give MW/Global Hutchinson penalties larger regularization budgets on an undershoot-dominated dataset. & Validity remains far below the asymmetric penalty even when the diagnostic gap shrinks. \\
\hline
\end{tabular}
\end{table}

\section{S9. Extended Figures}

This section contains an auxiliary figure for the main paper; S10 lists the script and data behind it.

\section{S10. Reproducibility and Provenance}

This section maps each result of the paper and of this supplement to the code and result files
behind it; paths are relative to the root of the code archive. The archive contains the
experiment, analysis and data-preparation scripts, selected result files (among them those behind
the tables), the prepared COMPAS, German and Adult data, a script that rebuilds the Fashion-MNIST
arrays from the official files, tests and setup instructions; its \texttt{README.md} lists the
commands, and this supplement reproduces no source code. Table~\ref{S-tab:supp-provenance} lists
the code and files behind each result, and Table~\ref{S-tab:supp-checklist} maps the areas of the
main paper's reproducibility checklist, which follows its references, to the places where they are
documented.

\paragraph{Recomputing the reported numbers.} \texttt{MANIFEST.md} lists every headline number
with the result file it comes from. \texttt{claims\_manifest.py} recomputes each number from that
file and serves as a regression test: with \texttt{-{}-check} it exits with an error if a number
has drifted. It also rechecks the summaries of the adaptive-query computation of S1 and recomputes
every result of the historical CelebA note in S4 from its aggregate file.

\paragraph{Protocols fixed before a run.} Several experiments were run against pass criteria
written into a protocol file that was committed to the code repository before the run; this is
what the text means when it says that criteria were fixed before a run. The table names these
files. They are in the code archive, but their timing is recorded only in the history of the
private repository, which reviewers cannot inspect.

\paragraph{Stored outputs.} The main-comparison table is backed by per-seed result files. For the
F-MNIST table and the Adult and COMPAS penalty-sweep rows, only the printed summaries of the
drivers were stored, so \texttt{claims\_manifest.py} checks those entries against the stored
printouts, not against per-seed rows. The included drivers can be rerun; per-seed files produced
later by different drivers do not reproduce these rows.

\paragraph{Commands for the recourse-rule suites of S5.} The four suites run
\begin{center}
\texttt{recourse\_rule\_baselines.py -{}-methods all -{}-seeds 0 1 2 -{}-max-train 2000 -{}-max-rejected 128}
\end{center}
with \texttt{-{}-rules validation\_tuned\_inflation} added for the two validation-tuned suites. Each run writes one per-seed CSV (\texttt{-{}-output}; an existing path
receives a numeric suffix), and \texttt{summarize\_multiseed.py fixed|vt} \(\langle\)per-seed
CSV\(\rangle\) \(\langle\)summary CSV\(\rangle\) writes the summary tables; \texttt{README.md}
lists all eight command lines.

\paragraph{The historical CelebA aggregates.} The script that produced the aggregates of S4
computed \(\kappa\) with a variant of the curvature helper \texttt{per\_point\_geometry}
(\texttt{signed\_recourse\_core.py}) that was not retained. Only the call site is in the
repository, and no committed version of the helper accepts the argument used there. The producer
script is specific to the compute environment of the runs and is not in the code archive. Apart
from each model's test accuracy, everything else the reported quantities use comes from routines
of the archive that are unchanged since the aggregates were committed and that the producer calls
with their committed signatures: the promised distance \(d_p\), the ray search for
\(d_{\text{ray}}\), and the ray-hit validity indicator.

{\footnotesize
\begin{longtable}{p{0.22\linewidth}p{0.35\linewidth}p{0.35\linewidth}}
\caption{Code and files behind each result. ``Criteria'' marks a protocol file committed before the run.}\label{S-tab:supp-provenance}\tabularnewline
\hline
\raggedright Result (section) & \raggedright Code & \raggedright Result, protocol and other files \tabularnewline
\hline
\endfirsthead
\hline
\raggedright Result (section) & \raggedright Code & \raggedright Result, protocol and other files \tabularnewline
\hline
\endhead
\hline
\endfoot
\raggedright Score-gauge audit (S1) & \raggedright \texttt{score\_gauge\_audit.py} & \raggedright \texttt{results/score\_gauge/} \tabularnewline[3pt]
\raggedright Probe bounds and collocation cancellation (S1) & \raggedright \texttt{verify\_probe\_escape.py}: exact recovery on quadratic profiles, the \(Md_p/3\) bound, the equivalence on trained models, the symbolic cubic case, and non-polynomial and kinked \(C^{2,1}\) profiles & \tabularnewline[3pt]
\raggedright Segment-safe fallback (S1) & \raggedright \texttt{endpoint\_segment\_fallback.py} recomputes the upper-tail fallback and checks the branch invariant & \raggedright endpoint-audit files (S2 row) \tabularnewline[3pt]
\raggedright Drift screen and ambiguity mass (S1) & \raggedright \texttt{verify\_drift\_screen.py}; float64 recomputation and grid scan: \texttt{verify\_drift\_float64.py} & \raggedright criteria: \texttt{DRIFT\_SCREEN\_\allowbreak FALSIFIER.md} \tabularnewline[3pt]
\raggedright Theorem 6.3 and Corollary~\ref{S-cor:s1-lattice} (S1) & \raggedright \texttt{verify\_one\_query\_\allowbreak lower\_bound.py}; \texttt{verify\_oracle\_lattice.py}; their result files check exact transcripts, regularity, first roots, finite margins and the separate parameter scalings & \raggedright criteria: \texttt{ONE\_QUERY\_LOWER\_\allowbreak BOUND\_FALSIFIER.md}; \texttt{ORACLE\_LATTICE\_\allowbreak FALSIFIER.md} \tabularnewline[3pt]
\raggedright Adaptive radii (conjecture) and Theorem~\ref{S-thm:s1-batch} (S1) & \raggedright \texttt{verify\_adaptive\_q\_\allowbreak query\_bounds.py}, which also asserts that no matching adaptive converse is claimed; its summaries are rechecked by \texttt{claims\_manifest.py} & \raggedright \texttt{results/\allowbreak theory\_audit/\allowbreak adaptive\_q\_\allowbreak query\_bounds.csv}; criteria: \texttt{ADAPTIVE\_QUERY\_\allowbreak LOWER\_BOUND\_\allowbreak FALSIFIER.md}; written argument for the conjecture, not part of this supplement: \texttt{ADAPTIVE\_Q\_\allowbreak QUERY\_FINDING.md} \tabularnewline[3pt]
\raggedright Scale-only illustration (S1) & & \raggedright rejected \(\kappa\): \texttt{compas\_perpoint\_\allowbreak geometry.csv}, \texttt{ray\_M\_audit\_adult.csv} \tabularnewline[3pt]
\raggedright Small Mondrian groups (S1) & \raggedright \texttt{min\_group} in \texttt{signed\_recourse\_core.py} & \raggedright the two \(\delta=0.01\) aggregates without this handling carry the prefix \texttt{legacy\_d01\_} and are documented in \texttt{results/\allowbreak signed\_quadratic/\allowbreak LEGACY\_AGGREGATES.md} \tabularnewline[3pt]
\raggedright Zero clip of the conformal distance (S1) & & \raggedright \texttt{conformal\_clip\_audit.csv}; \texttt{conformal\_clip\_verdict.json} (criteria fixed before the run) \tabularnewline[3pt]
\raggedright Staleness alarm and zero-search repair (S1) & \raggedright \texttt{shift\_monitor.py} (criteria in the script header); the per-point file with its exact-ray column is written by \texttt{compas\_shift\_\allowbreak perpoint\_generator.py} with the scaler fit on the training split only, and its \texttt{leaky} arm reproduces a version with the scaler fit on all rows, which supports no result & \raggedright \texttt{shift\_repair\_demo.csv}; \texttt{shift\_repair\_\allowbreak regime\_audit.csv} \tabularnewline[3pt]
\raggedright Endpoint agreement (S2) & \raggedright \texttt{endpoint\_audit.py} & \raggedright \texttt{results/endpoint\_audit/}; per-cell columns \texttt{frac\_dconf\_le\_2dp}, \texttt{max\_c\_conf} \tabularnewline[3pt]
\raggedright Nearest-boundary diagnostics (S2) & \raggedright metric-audit scripts, geometry helpers and ray-search utility & \raggedright metric-audit CSVs \tabularnewline[3pt]
\raggedright HVP timings (S3) & & \raggedright per-batch HVP wall-clock times of the CIFAR-10 networks in \texttt{results/deep\_recourse/} \tabularnewline[3pt]
\raggedright Directional-\(M\) screen and enclosure (S3.1) & \raggedright \texttt{estimate\_ray\_M.py}; \texttt{certify\_ray\_M.py}; depth audit: \texttt{estimate\_M\_celeba.py}, then \texttt{aggregate\_celeba\_a7.py} & \raggedright depth audit: per-point files in \texttt{celeba\_full\_a7/} \tabularnewline[3pt]
\raggedright Training suites (S3.2) & \raggedright the driver named in each row of the suite table of S3.2 & \tabularnewline[3pt]
\raggedright Repeated splits (S4) & \raggedright \texttt{split\_protocol\_audit.py} & \raggedright criteria: \texttt{results/\allowbreak split\_protocol/\allowbreak PREREGISTRATION.md}; shared cohorts: \texttt{cohort\_\{adult,fmnist\}\_\allowbreak perpoint.csv} \tabularnewline[3pt]
\raggedright Sign isolation and outer folds (S4) & \raggedright \texttt{ablation\_sign\_isolation.py} (protocol and the checks that reproduce the stored sweep values in its header) & \raggedright \texttt{sign\_isolation\_\allowbreak outer\_folds.csv} \tabularnewline[3pt]
\raggedright Signed-quadratic and conformal tables (S4) & \raggedright \texttt{signed\_quadratic\_\allowbreak recourse.py -{}-delta 0.05 -{}-target 0.95} & \raggedright the \texttt{agg\_d05\_*} files; probe comparison and deployment map: \texttt{sqr\_probe\_d05\_main.csv}, \texttt{agg\_probe\_d05\_main.csv}; Adult row of Table~\ref{S-tab:supp-validity-law}: \texttt{sqr\_all\_d05.csv} \tabularnewline[3pt]
\raggedright Common cohort (S4) & \raggedright \texttt{common\_cohort\_audit.py}; before any new field is used, it reproduces all 620 rows of \texttt{sqr\_probe\_d05\_main.csv} field for field, and a clean repeat gives identical files & \raggedright criteria: \texttt{COMMON\_COHORT\_\allowbreak FALSIFIER.md}; per-point, cell and verdict files \tabularnewline[3pt]
\raggedright Deployment shift (S4) & \raggedright \texttt{shift\_robustness.py} & \raggedright \texttt{shift\_robustness.csv}, with population standard deviations \tabularnewline[3pt]
\raggedright CIFAR-10 (S4) & \raggedright fixed recipe: \texttt{scripts/\allowbreak run\_\allowbreak deep\_\allowbreak cifar10\_\allowbreak locked.sh}; ten-seed table: \texttt{aggregate\_cifar\_\allowbreak tenseed.py} & \raggedright calibration status of the tuned rows: \texttt{deep\_cifar10\_locked\_\allowbreak workstation\_report.md}; seeds 3--9: \texttt{results/\allowbreak deep\_recourse/\allowbreak workstation\_lock\-ed\_extra/}; rerun of the collapsed seed, identical in every scientific field (only the wall-clock timing column differs): \texttt{workstation\_locked\_\allowbreak extra/recheck/} \tabularnewline[3pt]
\raggedright Deep forward-probe penalty (S4) & \raggedright \texttt{harvest\_deep\_penalty.py} (the aggregation script) & \raggedright criteria: \texttt{results/\allowbreak deep\_penalty/\allowbreak PREREGISTRATION.md}; selector: \texttt{HARVEST\_CORRECTION\_\allowbreak PREREGISTRATION.md}; diagnosis: \texttt{GPU\_FAILURE\_\allowbreak DIAGNOSIS\_\allowbreak PREREGISTRATION.md}; excluded smoke run: \texttt{deeppen\_sp\_probe\_\allowbreak l0.2\_seed0.csv} (\texttt{is\_scientific\_\allowbreak result=False}) \tabularnewline[3pt]
\raggedright FaiR-N (S4; cited in main-paper Section 1) & \raggedright \texttt{fairn2.py}; \texttt{fairn\_adult.py} & \tabularnewline[3pt]
\raggedright Group-conditional vs.\ matched-uniform (S4) & \raggedright \texttt{matched\_uniform\_audit.py} & \raggedright criteria: \texttt{results/\allowbreak signed\_quadratic/\allowbreak MATCHED\_UNIFORM\_\allowbreak FALSIFIER.md}; the earlier group-conditional files: \texttt{results/\allowbreak signed\_quadratic/\allowbreak GROUPCOND\_\allowbreak PROVENANCE.md} \tabularnewline[3pt]
\raggedright COMPAS without \texttt{decile\_score} or protected attributes (S4) & \raggedright \texttt{compas\_no\_decile\_\allowbreak robustness.py}; \texttt{compas\_no\_protected\_\allowbreak robustness.py}; each holds its arm, raw per-point rows and criteria & \tabularnewline[3pt]
\raggedright Channel attribution (S4) & \raggedright \texttt{compas\_channel\_\allowbreak attribution.py} & \raggedright the per-point file of \texttt{compas\_perpoint\_\allowbreak generator.py} \tabularnewline[3pt]
\raggedright Two additional datasets (S4) & & \raggedright \texttt{acsincome\_perpoint\_\allowbreak geometry.csv}; \texttt{taiwan\_perpoint\_\allowbreak geometry.csv} \tabularnewline[3pt]
\raggedright Historical CelebA note (S4) & \raggedright accuracy restrictions: \texttt{gated\_validity\_law.py}; \texttt{celeba\_public\_\allowbreak producer.py} defines and checks run and point records for future runs, trains no model and does not reproduce the historical aggregates & \raggedright \texttt{results/celeba/\allowbreak celeba\_recourse\_\allowbreak scale\_full.csv} \tabularnewline[3pt]
\raggedright Recourse-rule suites (S5) & \raggedright \texttt{recourse\_rule\_\allowbreak baselines.py}; \texttt{summarize\_multiseed.py}; commands above & \raggedright recourse-rule CSVs; matched validity: report, audit CSV and winners CSV under the recourse-rule baseline results \tabularnewline[3pt]
\raggedright Actionable recourse (S6) & \raggedright \texttt{actionable\_recourse.py}; per-group comparison: \texttt{b4\_actionable\_\allowbreak fairness.py}; synthetic jets: \texttt{verify\_constrained.py} & \raggedright the timestamped actionable-recourse CSV; cluster criteria: \texttt{FAIR\_CLUSTER\_\allowbreak FALSIFIER.md}, with the per-person records, summary and failure diagnosis \tabularnewline[3pt]
\raggedright Cost model (S7) & \raggedright cost-model audit script & \raggedright multiseed, all-winners, non-oracle-winners and report files under the recourse-rule baseline results \tabularnewline[3pt]
\raggedright Negative results (S8) & \raggedright tree ensembles: \texttt{blackbox\_surrogate\_\allowbreak audit.py}; calibrated target: \texttt{calibrated.py}, \texttt{frontier.py}; steepness penalty: \texttt{steepness.py}; Adult sweep of the main paper: \texttt{final\_adult.py} & \raggedright regenerated outputs in \texttt{results/\allowbreak negative\_results/} \tabularnewline[3pt]
\raggedright Figure~\ref{S-fig:supp-twosided} (S9) & \raggedright \texttt{fig\_twosided\_bound.py} & \raggedright \texttt{ray\_M\_audit\_compas.csv} \tabularnewline[3pt]
\end{longtable}}

\begin{table}[H]
\centering
\caption{Where the areas of the main paper's reproducibility checklist are documented.}
\label{S-tab:supp-checklist}
\begin{tabular}{lp{0.68\textwidth}}
\hline
Checklist area & Where documented \\
\hline
Problem and assumptions & Main-paper definitions, theorem statements, and limitations. \\
Algorithms and training & Main setup, S3.2, and the scripts of the code archive. \\
Data and preprocessing & Main setup, S3.2, the data notes and instructions of the code archive. \\
Evaluation metrics & Main definitions and S2. \\
Compute and runtime & Main-paper checklist and the runtime and progress output of the scripts. \\
Randomness and seeds & Main setup, S3.2, the seed columns of the result files, and the commands of the code archive. \\
Result provenance & Result files and reports, \texttt{MANIFEST.md}, and Table~\ref{S-tab:supp-provenance}. \\
Negative results & S8 and the scripts listed in Table~\ref{S-tab:supp-provenance}. \\
\hline
\end{tabular}
\end{table}

\section{S11. Data and Ethics}

All experiments use public benchmarks; no new human-subject data was collected and no personal
identifiers are redistributed (the released COMPAS copy keeps only the 14 fields the executable
workflows consume, excluding names, dates of birth, case numbers, and custody metadata). COMPAS
\citep{angwin2016machinebias} and the CelebA \texttt{attractive} attribute \citep{liu2015deep}
encode contested, historically biased constructs. We use them only to measure whether a prescribed
recourse step reaches the boundary it claims to reach; we neither endorse nor advance those
prediction targets, and no result here should be read as evidence that either label is a
legitimate decision variable.

\paragraph{Licenses and terms of use.} Adult, German Credit (Statlog), and the Taiwan credit-default
data are distributed by the UCI Machine Learning Repository under the Creative Commons Attribution
4.0 International license. Fashion-MNIST is released under the MIT License. ACSIncome is built with
the folktables package (MIT License) from American Community Survey microdata, whose use is
governed by the terms of service of the U.S.\ Census Bureau. ProPublica's COMPAS repository states
no license. The CIFAR-10 page states no license and asks users to cite its technical report
\citep{krizhevsky2009learning}. CelebA is available for non-commercial research purposes only
\citep{liu2015deep}.

The reported group differences are descriptive. The curvature-channel decomposition is a
reweighting, not causal mediation; the test of a conditioning benefit, fixed in advance, failed, so we make
no group-conditional allocation or equivalence claim, and the German, Adult, ACSIncome, and Taiwan
results are dataset-specific rather than a broad fairness claim.

The main risk of this work is that unconstrained score-space rays are read as advice. Without an
action map they move immutable and categorical coordinates, so deploying them to affected people
could produce unactionable or misleading recommendations; the constrained-action experiment is
only a feature-freezing diagnostic, and under the strict COMPAS mask no first-order actionable
dimension remains. We therefore report denied recourse and its group disparity, such as the COMPAS
validity gap of main-paper Section 10, as harms to be measured before any deployment, not as
capabilities.

\paragraph{Generative-AI use.} The AI use statement of the main paper, placed before its references, describes how generative AI tools were used.

\end{document}